\documentclass[
    fontsize=11pt,
    paper=A4,
    pagesize,
    twoside=true,
    BCOR=12mm,
    DIV=13,
    parskip=never,
    fleqn,
    draft=false,
    captions=bottombeside,
    listof=flat,
    toc=flat,
    footheight=138.60004pt,
    listof=nochaptergap
    ]{scrbook}

\usepackage{scrhack}

\usepackage[utf8]{inputenc}
\usepackage[T1]{fontenc}
\usepackage[greek.polutoniko,british]{babel}
\usepackage{fvextra}
\usepackage[style=british]{csquotes}
\usepackage{mathtools}
\usepackage{amssymb}

\usepackage[
    tracking=true,
    letterspace=50,
    babel=true,
    kerning=true,
    protrusion=true,
    ]{microtype}

\usepackage{FiraMono}

\usepackage{lmodern}

\usepackage{anyfontsize}
\usepackage{multicol}

\DeclareMathAlphabet{\mathbfit}{T1}{\rmdefault}{\bfdefault}{\sldefault}

\usepackage[singlespacing, nodisplayskipstretch]{setspace}
\usepackage[headsepline, automark, footwidth=foot:20pt]{scrlayer-scrpage}

\AtBeginDocument{%
  \abovedisplayskip=6pt plus 0pt minus 2pt
  \abovedisplayshortskip=6pt plus 0pt minus 2pt
  \belowdisplayskip=6pt plus 0pt minus 2pt
  \belowdisplayshortskip=6pt plus 0pt minus 2pt
}

\usepackage[low-sup]{subdepth}
\usepackage[]{graphicx}
\usepackage[export]{adjustbox}
\usepackage{xcolor}

\usepackage{overpic}

\usepackage[
    singlelinecheck=false,
    labelfont={bf,sf,color=draculagray},
    indention=20pt,
    skip=15pt,
    font={footnotesize},
    labelsep=quad,
    labelformat=simple,
    format=plain,
    hypcap=false
    ]{caption}

\usepackage[
    labelformat=brace,
    singlelinecheck=false,
    justification=raggedright,
    font={footnotesize}
    ]{subcaption}
\usepackage{listings}

\makeatletter
\newcommand{\currentfontsize}{\fontsize{\f@size}{\f@baselineskip}\selectfont}
\makeatother

\newcommand{\cminline}[1]{%
    {\fontsize{9}{9}\selectfont\lstinline{#1}}%
}
\newcommand{\tminline}[1]{%
    {\fontsize{10}{10}\selectfont\lstinline{#1}}%
}

\makeatletter
\let\FV@ListProcessLineOrig\FV@ListProcessLine
\def\FV@ListProcessLine#1{%
  \ifx\FV@Line\empty
    \hbox{}\vspace{-7pt}%
  \else
    \FV@ListProcessLineOrig{#1}%
  \fi}
\makeatother

\usepackage{booktabs}
\usepackage{tabularx}
\usepackage{multirow}
\usepackage{pbox}
\usepackage{makecell}

\usepackage{soulutf8}
\usepackage{siunitx}

\usepackage{lettrine}
\definecolor{defaulttextcolor}{HTML}{020122} 
\definecolor{draculablack}{HTML}{282a36}
\definecolor{draculagray}{HTML}{585C74}  
\definecolor{draculawhite}{HTML}{f8f8f2}
\definecolor{draculablue}{HTML}{6272a4}
\definecolor{draculacyan}{HTML}{8be9fd}
\definecolor{draculagreen}{HTML}{50fa7b}
\definecolor{draculaorange}{HTML}{ffb86c}
\definecolor{draculapink}{HTML}{ff79c6}
\definecolor{draculapurple}{HTML}{bd93f9}
\definecolor{draculared}{HTML}{ff5555}
\definecolor{draculayellow}{HTML}{f1fa8c}
\newcommand{\chaptersubtitle}[1]{%
    {\LARGE\textit{\textcolor{draculagray}{#1}}}
}

\usepackage{normalcolor}
\setnormalcolor{defaulttextcolor}

\newcommand{\captionurl}[1]{%
    {\fontsize{7.7}{7.7}\selectfont\url{#1}}%
    }

\usepackage[
    backref=none,
    colorlinks=true,
    linkcolor=draculagray,
    citecolor=draculagray,
    urlcolor=draculagray,
    draft=false
    ]{hyperref}
\usepackage[
    backend=biber,
    style=chem-angew,
    backref=false,
    backrefstyle=all+,
    chaptertitle=true
    ]{biblatex}
\makeatletter
\DeclareCiteCommand{\fullcite}
  {\defcounter{maxnames}{\blx@maxbibnames}%
    \usebibmacro{prenote}}
  {\usedriver
     {\DeclareNameAlias{sortname}{default}}
     {\thefield{entrytype}}}
  {\multicitedelim}
  {\usebibmacro{postnote}}
\DeclareCiteCommand{\footfullcite}[\mkbibfootnote]
  {\defcounter{maxnames}{\blx@maxbibnames}%
    \usebibmacro{prenote}}
  {\usedriver
     {\DeclareNameAlias{sortname}{default}}
     {\thefield{entrytype}}}
  {\multicitedelim}
  {\usebibmacro{postnote}}
\makeatother

\usepackage{bookmark}
\usepackage{marginnote}
\usepackage[record,automake,toc,section=section]{glossaries-extra}
\usepackage{glossary-mcols}

\GlsXtrLoadResources[src={abbrev}, sort={letter-case}]
\setabbreviationstyle{long-short}  

\setglossarystyle{long}

{\begin{longtable}[l]{rp{0.7\textwidth}}}%
    {\end{longtable}}

\RedeclareSectionCommands[
    pagestyle=empty,
]{part}

\RedeclareSectionCommand[
    tocbeforeskip=0pt plus .2pt,
    toconstarthigherlevel=\addvspace{12pt}
]{chapter}

\renewcommand{\glossarysection}[2][]{}

\RedeclareSectionCommands[
    beforeskip=0.3ex plus 0.2ex minus -0.2ex
]{paragraph,subparagraph}

\DeclareNewSectionCommand[
  style=section,
  level=\sectionnumdepth,
  beforeskip=3.25ex plus 1ex minus .2ex,
  afterskip= 1.5ex plus .2ex,
  indent=0pt,
  font=\mdseries\slshape,
  tocstyle=default,
  toclevel=\sectiontocdepth,
  tocindent=0pt,
  tocnumwidth=0pt 
]{Article}

\renewcommand*\addArticletocentry[2]{%
  \addtocentrydefault{Article}{}
    {\IfArgIsEmpty{#1}{}{Article~\makebox[1.5em][l]{#1\protect\autodot}}#2}%
} 

\setkomafont{pagehead}{\footnotesize}
\setkomafont{pagefoot}{\footnotesize}
\setkomafont{pagenumber}{\normalsize\slshape\bfseries}
\setkomafont{section}{\Large\sffamily\mdseries}
\setkomafont{subsection}{\large\sffamily\mdseries}
\setkomafont{subsubsection}{\large\sffamily\mdseries}
\setkomafont{paragraph}{\sffamily}
\setkomafont{subparagraph}{\rmfamily}
\DeclareTOCStyleEntry[entrynumberformat={\bfseries}]{default}{section}
\DeclareTOCStyleEntry[entryformat={\small\sffamily\mdseries}, entrynumberformat={\small\bfseries}, pagenumberformat={\small\rmfamily\slshape}]{default}{subsection}
\DeclareTOCStyleEntry[entryformat={\small\sffamily\mdseries}, entrynumberformat={\small\bfseries}, pagenumberformat={\small\rmfamily\slshape}]{default}{subsubsection}
\DeclareTOCStyleEntry[entryformat={\hspace*{-0.6cm}\small\sffamily\mdseries},entrynumberformat={\small\bfseries},pagenumberformat={\small\rmfamily\slshape}]{default}{figure}
\DeclareTOCStyleEntry[entryformat={\hspace*{-0.6cm}\small\sffamily\mdseries},entrynumberformat={\small\bfseries},pagenumberformat={\small\rmfamily\slshape}]{default}{table}

\makeatletter
\newcommand*{\rom}[1]{\expandafter\@slowromancap\romannumeral #1@}
\makeatother

\lehead[]{\rightmark}
\cehead[]{}
\rehead[]{\leftmark}
\lohead[]{\leftmark}
\cohead[]{}
\rohead[]{\rightmark}

\lefoot[]{}
\cefoot[]{}
\refoot[]{}
\lofoot[]{}
\cofoot[]{}

\newcommand{\printtitle}{Clustering---Basic concepts and methods, J.-O. Kapp-Joswig, B. G. Keller}

\rofoot[\raisebox{125pt}{\raggedleft\printtitle} \quad\rule{0.5pt}{135pt}\quad \raisebox{125pt}{\pagemark}]{\raisebox{125pt}{\raggedleft\printtitle} \quad\rule{0.5pt}{135pt}\quad \raisebox{125pt}{\pagemark}}

\lefoot[\raisebox{125pt}{\pagemark} \quad\rule{0.5pt}{135pt}\quad\raisebox{125pt}{\raggedright\printtitle}]{\raisebox{125pt}{\pagemark} \quad\rule{0.5pt}{135pt}\quad\raisebox{125pt}{\raggedright\printtitle}}

\newcommand{\ivector}[1]{
    {\small\ensuremath{\left(\begin{matrix}\,#1\,\end{matrix}\right)}}%
}

\newlength{\tindent}
\newcommand{\customindent}[1]{%
    \settowidth{\tindent}{%
    \footnotesize\sffamily\bfseries#1%
    }\hspace{\tindent}%
    }

\newcommand{\figureindent}{\customindent{{Figure \thetable\quad}}}

\newlength{\doublespacetypewriter}
\newcommand{\chapterwsub}[2]{%
\chapter[#1]{#1%
    \newline \chaptersubtitle{#2}%
    }%
}

\newcommand{\stdmarginnote}[1]{%
    {\marginnote{\centering\itshape{#1}}}
}

\newcommand{\extref}[2]{%
    ~\mbox{\ref{#1}\textcolor{draculagray}{#2}}%
    }

\newcommand{\captionofbottom}[4]{%
    \captionof{#1}[#2]{\textsf{\textbf{#3}}\newline #4}%
    }

\newcommand{\captionofside}[4]{%
    \captionsetup{indention=5pt}%
    \captionof{#1}[#2]{\textsf{\textbf{#3}} #4}%
    }

\newcommand{\custompar}[1]{%
    \bigskip\noindent\textsc{#1}%
}

\newcommand{\firstsecpar}[1]{%
    \textsc{#1}%
}

\newcommand{\subl}[1]{%
    {\fontsize{9}{9}\selectfont\textsf{\textbf{#1)}}}%
}

\newenvironment{illustration}{%
    \noindent
    \begin{minipage}[t]{\textwidth}
        \vspace{0pt}
    }{%
    \end{minipage}
    }

\mathchardef\mhyphen="2D

\DeclareSIUnit\angstrom{\text {\AA}}

\recalctypearea

\renewcommand{\printtitle}{Clustering---Basic concepts and methods, J.-O. Kapp-Joswig, B. G. Keller}

\begin{document}


\thispagestyle{empty}
\begin{titlepage}

\large
{\LARGE\sffamily\bfseries Clustering---Basic concepts and methods}

\smallskip
\textcolor{draculagray}{Jan-Oliver Kapp-Joswig, Bettina G. Keller\(^\star\)}

\smallskip
\textcolor{draculagray}{28.11.2022}

\vfill\noindent
\textcolor{draculagray}{\(^\star\)Freie Universität Berlin\\Institute of Chemistry and Biochemistry\\Theoretical Chemistry\\Arnimallee 22, 14195 Berlin}

\end{titlepage}

\pagenumbering{arabic}
\chapterwsub{The basics}{Terminology and definitions}
\label{chap:clustering_intro}

\lettrine{C}{}lustering\footnote{The term \textit{clustering} will be
used here either as a verb to describe the act of performing the respective
analysis (to cluster objects), or as a noun to describe the outcome
of the analysis (a clustering of objects).} is an analytic process that identifies
associations of some kind (i.e. \textit{clusters}) among a number of
considered objects.
Phrasing this differently, \enquote{clustering is a synonym for the
decomposition of a set of entities into \textit{natural
groups}}.\cite{Gaertler2005}
The word \enquote{natural} appearing in this quote indicates already
that this definition leaves room for interpretation.\stdmarginnote{general definition}
Clustering as a task has as such no precise, unambiguous goal, and what
a cluster actually is can not be universally defined.
Sometimes, clustering is even reduced to just achieving \enquote{a
grouping of objects} as a common denominator without further
specifications.\cite{Henning2016}
A sorting of arbitrary objects into groups can be done in many different
ways, and how it is done best, does usually depend on the question of
\textit{why} it is done.
The same set of objects can be clustered based on all of its properties
or only a selection thereof, and even if the same properties are used,
different
clustering approaches will in general yield different results.
We will proceed on the common ground, that clustering tries to establish
some sort of relation between the clustered objects, meaning that
objects within the same identified group should be alike with respect to
(some of) their properties, and in reverse, objects in different groups
should be less alike.
This still rather open notion can be expressed as the paradigm of\stdmarginnote{object relationship}
\textit{strong within-cluster relationship} versus \textit{weak
between-cluster relationship}, that should be satisfied by a structure
found among objects through a
clustering.\cite{Gaertler2005}
How different clustering  techniques understand the broader notion of
relationship, will be discussed in section~\ref{sec:similarity_definitions}.
Obviously, relationships among objects can be subjective and finding
relationships has a lot to do with human intuition.
Humans are in a way intuitively very good at the task of clustering
objects visually by distinguishing different kinds of objects according
to spatial proximity, colour, or other---sometimes oblivious---criteria.
Just by looking at a set of objects, we can often immediately see or
feel how individual objects could be appropriately split into
groups.\stdmarginnote{human analysis}
The problem is of course, that not every set of objects can be
conveniently visualised and even worse, our intuition would depend on the
exact way how the visualisation is achieved.
Additionally, human analysis is usually limited to a rather low number
of objects or a low number of simultaneously considered properties, not
to mention that while we may be able to identify groups of objects in an
instance, documentation and isolation of the result to make further use of
it is more often than not extremely tedious.
Figure~\ref{fig:clustering_ducks} should illustrate the ambiguity that usually
accompanies clustering tasks with a casual example.
It shows a photograph of plastic ducks which can be analysed by
clustering in many different ways.
Most important would be the question, what the clustering should be done
for.
Do we want to split the image into areas of different colour so that
each identified cluster contains pixels of related colours?\stdmarginnote{a casual example}
Do we want to identify groups of pixels that form individual ducks so
that each cluster contains only the set of pixels for one respective duck?
Or do we want to identify groups of ducks, that is for example ducks
with a certain orientation or local rafts of ducks?
The overall goal may be accomplished by either considering individual pixels
or entire ducks or parts of the ducks as the set of objects to cluster and
the examined object properties will be different in each case.
Along the line, there are many open detail questions: how fine should
the clustering be? Do we only want to separate red from yellow or do we
want to distinguish different shades of yellow?
Is there a maximum number of clusters we want to deal with?
Should each cluster be restricted to contain a minimum or maximum number
of object members?
Some of these clustering tasks can be in principle well done by human
visual analysis but in practice this is seldom really feasible or
efficient.

\begin{illustration}
\hspace{1em}\begin{minipage}[t]{0.34\textwidth}
    \vspace{0\baselineskip}
    \centering
    \includegraphics[width=\textwidth, frame]{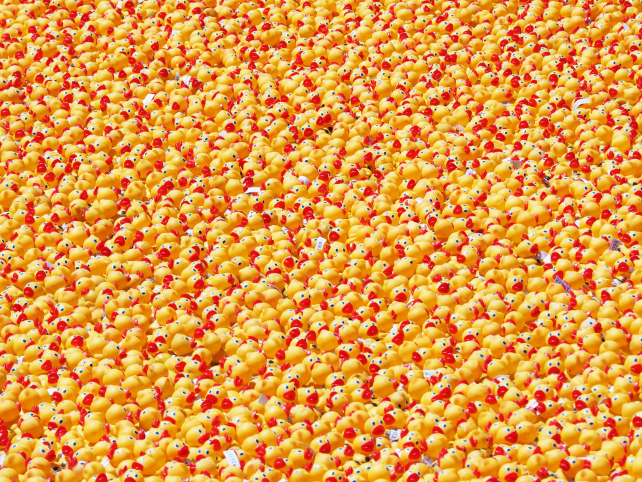}
\end{minipage}\hspace{0.05\textwidth}%
\begin{minipage}[t]{0.55\textwidth}
    \vspace{-0.25\baselineskip}
    \begingroup
    \hypersetup{hidelinks}
    \captionofside{figure}{%
        Clustering an image of plastic ducks}{%
        Clustering an image of plastic ducks}{%
            The picture on the left shows a crowd of plastic ducks. The
            formulation of a clustering task involves several questions:
            what do we want to achieve---a separation into areas of
            different colour? A grouping of pixels belonging to each
            duck? A grouping of related ducks? What do we consider as
            objects---individual pixels? Individual ducks? By which
            property do we want to cluster---colour of pixels?
            Orientation of ducks? For many of these tasks, we have an
            intuitive idea about what the result looks like. Photo by
            \href{https://unsplash.com/@marcuslenk?utm_source=unsplash&utm_medium=referral&utm_content=creditCopyText}{\textcolor{draculagray}{Marcus
            Lenk}} on
            \href{https://unsplash.com/s/photos/rubber-duck?utm_source=unsplash&utm_medium=referral&utm_content=creditCopyText}{\textcolor{draculagray}{Unsplash}}.
            }
    \endgroup
    \label{fig:clustering_ducks}
    \vspace{0.2\baselineskip}
\end{minipage}
\end{illustration}
Computer aided clustering schemes try to produce fast and reliable
groupings of possibly very large numbers of objects based on objective,
quantitative criteria.\stdmarginnote{computer analysis}
As such, computer analysis is supposed to complement intuition driven
human analysis.
It turns out, however, that it is not an easy endeavour to find
automatic clustering schemes that are universally adequate.
A vast number of clustering techniques has been proposed, based on
various underlying principles, expressed in many different method
frameworks, and realised through an even larger number of concrete
implementations of the respective algorithms.
The reason for this might be, that clustering commonly serves a specific
practical purpose and the question of what a clustering
scheme should do and the notion of what a cluster should be, depends on
the formulation of the clustering problem---which in turn varies
strongly among disciplines.
Different clustering approaches have been developed before different
application backgrounds, making different assumptions about the
clustered objects and making different demands on the clustering results.\cite{Estivill-Castro2002}
The matter is complicated by the fact that clustering approaches are
hard to compare with each other and that in general it cannot be
argued about the \textit{right} or \textit{wrong} of a specific
clustering result.
Clustering schemes produce clusterings in accordance with their
underlying design.
In this sense, any clustering result is basically
\textit{correct}---disregarding of course that clustering
implementations may contain bugs that make them incorrect.
This does not mean, however, that any clustering result is also
\textit{relevant} or \textit{meaningful} in any situation.
The question of what is a good clustering is biased by the view and
beliefs of whoever uses it.\cite{Estivill-Castro2002}
Consider the six different groupings of ball-like objects in
figure~\ref{fig:clustering_imprecise}.
Based on a personal opinion, you may find one of the groupings most
persuasive but without further context all of them are equally
appropriate.
\begin{illustration}
\hspace{1em}\begin{minipage}[t]{0.49\textwidth}
    \vspace{0pt}
    \includegraphics[]{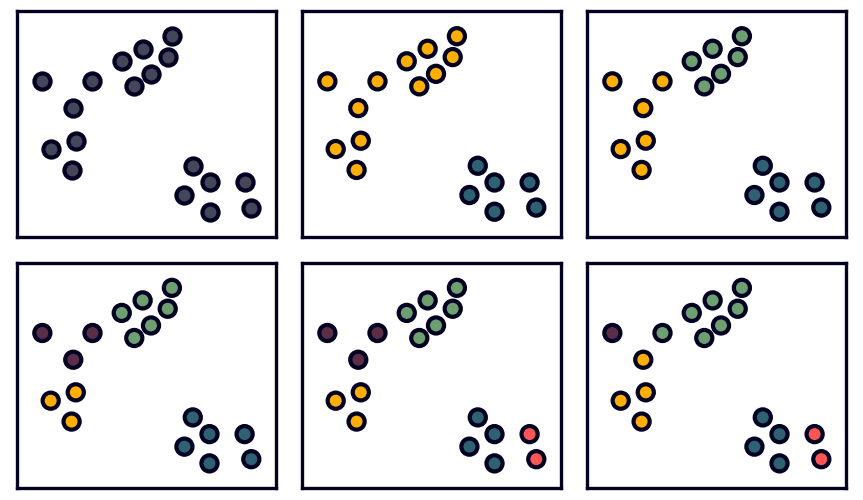}
\end{minipage}%
\begin{minipage}[t]{0.36\textwidth}
    \vspace{-0.25\baselineskip}
    \captionofside{figure}{%
        Clustering is imprecise}{%
        Clustering is imprecise}{%
        The formulation of what a clustering should achieve and the
        definition of what a cluster should be is ambiguous and depends
        on the perspective of the user. All the clusterings on the left
        can be legitimate (including the trivial case of just one
        group).}
    \label{fig:clustering_imprecise}
\end{minipage}
\end{illustration}

Eventually, the usefulness of a clustering result with respect to the
initial intention decides whether it is indeed valid.
The validity of a clustering result can be effectively judged only
either by an expert through careful analysis of the obtained clusters
(called \textit{manual} evaluation) or by a further use of the
clustering in yet another application to deliver a result that can be
better evaluated (\textit{indirect} evaluation).
Manual evaluation puts the focus on human intuition (or let's call it
expertise for that matter) back on the table.\stdmarginnote{manual evaluation}
Expert users need to have a certain expectation about the result of a
clustering that is predicated on an intuitive or experience-based idea
of the identified clusters, which may also be called some sort of domain
knowledge.
A clustering is good, if it meets the expectation.
Clustering schemes can be chosen or results can be refined in order to
match the expectation as closely as possible.
Things become difficult, though, if manual inspection, which mostly
means visual inspection, is not feasible for example when a clustering
contains many groups or if groups are split based on many object
properties.
A fundamental issue about expectation is also that it could be unjustified.
If a clustering does not meet the expectation, it could be either the
clustering or the expectation that needs adjustment.
It can also be a problem, if there is no expectation.
In many practical situations, it is not exactly known what kind of
clustering is most suitable for a given set of objects.
Generally speaking, clustering is often employed in the hope of
\textit{finding} some sort of structure in the set of clustered objects.
What is actually done by a clustering process is that a certain
structure in accordance with the design of the chosen clustering method is
\textit{imposed} upon the objects.\cite{Aldenderfer1984}
Clustering always produces a structure as a kind of hypothesis about the
clustered objects\cite{Estivill-Castro2002} but there is  no guarantee
that this structure is actually well-founded or---to make a connection
to the introductory definition of clustering---the \textit{natural}
structure among the objects.
Indirect evaluation can partly counter the vagueness that we are
confronted with when the quality of clusterings should be assessed.
Instead of comparing clusters obtained by a specific procedure to a
subjective expectation that may still be flawed, it can be
helpful to\stdmarginnote{indirect evaluation} get objective feedback from an
application which the clusters are used for.
The clustering result itself is of secondary importance if it proves
functional for something that should be done with the clustering.
The result of a subsequent step validates or invalidates the
clustering step.
Unfortunately, such indirect evaluation is not always practical, for
example because the groups obtained through a clustering are themselves
the result of interest or because the application they are used for does
not give a definite positive or negative response.
It can also be problematic, if the validating application gives falsely
positive feedback, that is if things seem to work well but they actually
do not.
Similarly, things may work well in an application for the wrong reason
which makes it difficult to infer anything about the quality of a
clustering in general.
An example for indirect evaluation with a direct connection to the
content of this thesis would be the construction of kinetic Markov-models
from molecular simulations.
In this context, clustering is used to define conformational (micro-)states
on top of which the model is estimated.
In principle, a good clustering is one that allows the construction of a
good Markov-model.
In reality, the question whether one has got a good model is not that
straightforward, though, in particular not if the aim is to rank
clustering results quantitatively.
Moreover, also a suboptimal clustering, for example one where certain
conformational states (i.e. relevant information) are missing, may give
a seemingly good model.
There are two other forms of clustering validation that are worth
mentioning:
\textit{internal} and \textit{external} evaluation.
Internal evaluation tries to assess the quality of a clustering result
based on the corresponding grouping of objects
itself.
Making assumptions about how identified clusters should be constituted
in the ideal case, clustering results can be quantified with respect to
these assumptions.\stdmarginnote{internal evaluation}
In fact there exists a whole bunch of statistical measures that can be
used for checking to what extent a clustering satisfies some idealised
theorization.
A few of them will be discussed in section~\ref{sec:kmeans} in the context
of \textit{k}-means---a particular clustering approach for which internal
validation criteria can be well applied.
There is, however, the potential risk of running into a circular
argumentation.
A conjecture about what ideal clusters are, is likely to be one that
could be used as the underlying principle of a clustering method in the
first place.
Clustering schemes that find clusters according to some criterion for how
the resulting clusters should be organised will of course be considered
favourable by a validation that uses the same criterion.
Clustering schemes on the other hand that are based on different
assumptions, will be probably considered less favourable.
An agreement with a certain validation criterion is therefore only
meaningful if different concrete clustering algorithms, based on the
very same principle, are compared to find out which one produces
clusterings in closest agreement with the initial clustering objective.
In other words, internal validation can proof if a specific algorithm or
implementation is successful in realising an underlying principle.
Conversely, an agreement or disagreement of a clustering result with a
validation criterion tells us basically nothing if it is not the intend
of the clustering method to satisfy the validation criterion in the
first place.
In any case, internal validation measures may just be nonsensical
because the assumption they are based on are not aligned with the actual
nature of a specific object set, and the validity of the validation
technique itself often needs to be scrutinised.
External validation uses a reference clustering to compare that to the
result of a specific clustering method.
It assumes that for a given set of objects the \textit{true} groups are
known so that the groups obtained in a clustering can be judged by the
number of individual objects correctly assigned to a certain
group.
Because it is in general impossible to know the true group structure of
arbitrary object sets, external validation uses exemplary benchmark sets
for which it is presumed that the inherent groups are obvious or at least
apparent to the expert that created the example.\stdmarginnote{external validation}
These sets are often synthetic and relatively simple.
Benchmarking different methods of clustering against an array of
representative test object sets, can give valuable insight to the
fitness of a method for a given purpose.
It can tell us, if a clustering solves a designed problem in the desired
fashion and can reveal major differences between clustering
techniques.
It can also be used to check if a specific clustering algorithm produces
results in line with its own principles.
On the other hand, a good clustering performance in benchmarks can not
be transferred one-to-one to the assessment of realistic objects sets.
It can only give a hint if a clustering method is in principle suitable
for the analysis of a specific object set under the constraint that the
set in question resembles the benchmark set but it can not validate the
actual clustering result.
And yet again, benchmark sets are based on certain assumptions about
identified clusters---assumptions that may be flawed because they are
based on intuition or because they were made to comply with a specific
purpose or idea about clustering.
Clustering is a complex and sometimes confusing topic that entails a
fair amount of tension between objectivity and intuition---between
theoretical conciseness and practical relevance.
It can be understood from multiple standpoints and its proper usage
depends on perspective, which makes it not surprising that the terminology
in the context of clustering is often not very consistent.
Despite all this, a combination of computational clustering techniques
based on objective criteria with human inspection and decision making
can be very powerful.\cite{Sips2009}
In practice, clustering is often a tool for exploration and a way to get
some relational insight about the considered objects from different angles.
It cannot be considered a tool that provides definitive answers about
the actual group structure of the objects in all cases.
Acquiring knowledge through clustering, is often an iterative and
incremental process.
Clustering is applied to various problems in a wide array of scientific
and industrial fields.\cite{Han2000}
A very early documented example goes back to 1855 when during a massive
cholera outbreak in London, cases of death were tracked on a city
map\stdmarginnote{clustering applications} and revealed major sources of
infection where they accumulated in clusters.\cite{Snow1855}
Of course neither the term cluster in the present formal sense nor the
respective analysis has been established at this time, although an
intuitive comprehension of object groups may as well existed already
much earlier.
For instance, \cite{Henning2016} names Aristotle's classification of
living things as one of the first-known clusterings.
Clustering as an acknowledged computational tool
has its roots in the 1960s beginning with biological
taxonomy.\cite{Aldenderfer1984,Sokal1963}
The clustering of molecular
objects, that is for example conformational snapshots obtained from \gls{md}
simulations, experienced an upswing in the
1990s.\cite{Karpen1993,Shenkin1994,Torda1994}

\custompar{To cluster objects usually means} to put group \textit{labels} on
individual objects that indicate a \textit{membership}.
Clustering procedures can be technically discriminated based on how this
assignment is made:\stdmarginnote{forms of object labelling}
in the simplest case, each object will be given exactly one label, e.g.
an integral number (object \(a\) belongs to group 1, object \(b\) to
group 2, and so forth).
If in the end all of the initially considered objects are labelled, the
clustering is called \textit{complete}, or \textit{exhaustive}, or is
described as \textit{full} clustering.
The property of assigning objects to one and only one group makes a
clustering \textit{exclusive}, \textit{hard}, \textit{strict} or
\textit{crisp}.
If on the contrary, some of the objects may be left out of the
assignment---that is they are treated as noise or outliers---the
clustering can be called \textit{partial} or explicitly a clustering
\textit{with outliers}.
Furthermore, the counterpart of hard clustering would be either
\textit{overlapping} if objects are allowed to be a member of more than
one cluster, or \textit{soft} (also \textit{fuzzy}) if objects are
assigned to clusters  with a certain degree of membership.
\textit{Probabilistic} label assignments are more or less the same as
fuzzy just with the additional flavour that group memberships are seen
as probabilities while the true memberships of objects to groups are
assumed to be crisp.\cite{Henning2016}
To assign group labels to objects is also the aim of
\textit{classification} which describes the association of objects
into one of multiple \textit{classes} or \textit{categories}.
Traditionally, clustering can be therefore seen as a form of
classification and both are in fact widely used as synonyms in
literature especially from a statistical standpoint.\cite{Cormack1971}\stdmarginnote{clustering{\newline}vs.{\newline}classification}
Arguably, they are not referring to the same thing, though.
On the one hand, classification through clustering is possible as the
clusters identified by clustering can be interpreted and used as
categorical classes.
On the other hand, clustering is not necessarily always classification
because the resulting clusters can be interpreted and used differently,
e.g. as a form of discretisation or condensation.
Moreover, classifications can be done not only through clustering but
through a variety of other techniques.
In machine learning terms, clustering and classification are
predominately distinguished more strictly.\cite{Henning2016,Jain2010}
Clustering can be understood as an \textit{unsupervised} learning
process through wich objects are labelled without any kind of labels
already being present that could guide the procedure.
Clustering basically invents the labels in the first
place.\cite{Estivill-Castro2002}
Classification of objects on the contrary is equivalent to object
labelling with present labels for those or other objects that are used
to control the process, which is understood as \textit{supervised}
learning.
Classification uses labels that have been established through something
else.
If clustering is used for classification, it may be referred to as
unsupervised classification to discern it from forms of supervised
classification.
It should be noted, however, that this differentiation is softened by
the fact that there is indeed such a thing as semi-supervised and
supervised clustering.\cite{Aggarwal2014,Finley2005}
Coming back to the somewhat silly duck example at the beginning in
figure~\ref{fig:clustering_ducks}, one could ask now the question if
humans are actually very good at intuitive clustering or rather at
classification.
Can we identify individual ducks because we can cluster them ad-hoc or
because we already know from experience what a duck looks like?
The difference between the process of grouping objects into groups that
are de facto unknown and created while the grouping
happens---independent of previous groupings---and that of grouping
objects into groups that are already fixed may be subtle but it
contributes to the overall bewilderment that may at times accompany
clustering as an analysis tool.
In any case, it can make a rather big difference for the choice of
clustering methods whether they should be used to classify data or not.
Up to here, we rather abstractly talked about \enquote{objects} as the
entities that are being subjected to a clustering.
The next section will try to clarify what this actually means concretely
for computational clustering---using either the concept of objects as
points embedded in a metric space or as nodes in a graph network.

\section{Data sets and representations}
\label{sec:data_sets_and_repr}

\textsc{The topic of clustering} that is concerned with finding
groupings of objects based on relationships between these objects, leads
us to the concept of \textit{data}.
To cluster objects means to cluster data of some specific shape or form.
The rather open term \textit{data set}, i.e. a collection of
\textit{data}, is unfortunately in turn not defined very sharply and its
use varies among disciplines.
According to the UNECE cited in the OECD glossary of statistical
terms,\cite{OECD2008} \enquote{data is the physical representation of
information in a manner suitable for
communication}\cite{UNECE2000}.\stdmarginnote{general definiton}
Depending on the actual context, the kind of stored information can be
very different.
As far as we should be concerned throughout this chapter, a data set is
a collection of identically structured \textit{records} where each
record is basically a listing of values for arbitrary variables.
Following a particularly practical definition in the IBM z/OS
documentation, a record in a data set \enquote{is the basic unit of
information used by a program}\cite{IBMzos}.
As such, records can be usually understood as tabular data, which makes
a data set essentially a data table (a file) in which each column stands
for a variable and each row represents a single coherent group of values
for these variables---neglecting the complexity that is posed by a
multitude of file-formats that structure this information in different
ways.
Alternative names for records are \textit{examples} or \textit{samples}.
An individual variable within a  record can also be referred to as a
\textit{field} or \textit{feature}, and its values (a single
\textit{datum}) can be of numerical, ordinal, or nominal nature---or
actually of any other imaginable abstract type.
A classic example is Ronald Fisher's \textit{Iris plant} data set which
collects 150 records of flowers for which the sepal and petal length and
width (the dimensions of two different leaf types) have been measured
(see table~\ref{tab:iris}).\cite{Fisher1936}\stdmarginnote{Iris data}
For each measured flower, additionally the botanical class (one of
\textit{setosa}, \textit{versicolour}, or \textit{virginica}) is known.

\begin{illustration}
    \hspace{0.03\textwidth}%
    \begin{minipage}[t]{0.4\textwidth}
    \vspace{0pt}
        \begin{tabular}{rrrrrl}
            \toprule
            ID & \(s_{l}\) & \(s_{w}\) & \(p_{l}\) & \(p_{w}\) & class \\
            \midrule
            1 & 5.1 & 3.5 & 1.4 & 0.2 & \textit{setosa}\\
            2 & 4.9 & 3.0 & 1.4 & 0.2 & \textit{setosa}\\
            3 & 4.7 & 3.2 & 1.3 & 0.2 & \textit{setosa}\\
            4 & 4.6 & 3.1 & 1.5 & 0.2 & \textit{setosa}\\
            \dots &&&&& \\
            \bottomrule
        \end{tabular}
        \vspace{0.25\baselineskip}
    \end{minipage}\hspace{0.04\textwidth}%
    \begin{minipage}[t]{0.51\textwidth}
        \vspace{-0.275pt}
        \captionofside{table}{%
        Extraction of Fisher's \textit{Iris plant} data set}{%
        Extraction of Fisher's \textit{Iris plant} data set}{%
            Each entry is marked with a unique record ID and identified
            with a category (class). The quantities \(s_\mathrm{l}\)
            (sepal length), \(s_\mathrm{w}\) (sepal width),
            \(p_\mathrm{l}\) (petal length), and \(p_\mathrm{w}\) (petal
            width) are measured in centimeters.}
         \label{tab:iris}
    \end{minipage}
\end{illustration}

In the following we will have a closer look at how to represent, visualise,
and analyse a tabular data set like this before the background of clustering.
Fisher's \textit{Iris plant} data set has four real-valued numeric features
with identical physical dimension (a length in centimetres).
These features as such span
a feature space in which each record can be represented as a single
point, i.e. a feature vector in four dimensions.
The data set of \(n = 150\) entries can be generally denoted as
\begin{equation}
    \mathcal{D} = \{x_1, ..., x_n\}\,,
\end{equation}
where each data point is a sample from the \(m = 4\) dimensional feature
space \(x_i \in X\) with \(x_i = \ivector{x_{i,1}, ..., x_{i,m}}\).
In this special case, the feature space coincides with
\(\mathbb{R}^4_{>0}\).\stdmarginnote{feature space}
For clustering analyses, it is generally beneficial if the feature space
is a real space \(X \subset \mathbb{R}^m\) because then basic
mathematical operations with respect to the samples in a data set (in
particular distances between points) are well-behaved, and in fact
there are clustering methods that do explicitly depend on this.
Figure~\ref{fig:iris_data_points_true_labels} shows a visualisation of
the \textit{Iris} data as two separate 2-dimensional projections for
the sepal and the petal features.
Each data sample appears in the plots as a single point at coordinates
corresponding to the values of the respective orthogonal features, and
coloured according to the known biological flower class of the sample.

\begin{illustration}
    \begin{minipage}[t]{0.76\textwidth}
        \vspace{0pt}
        \includegraphics{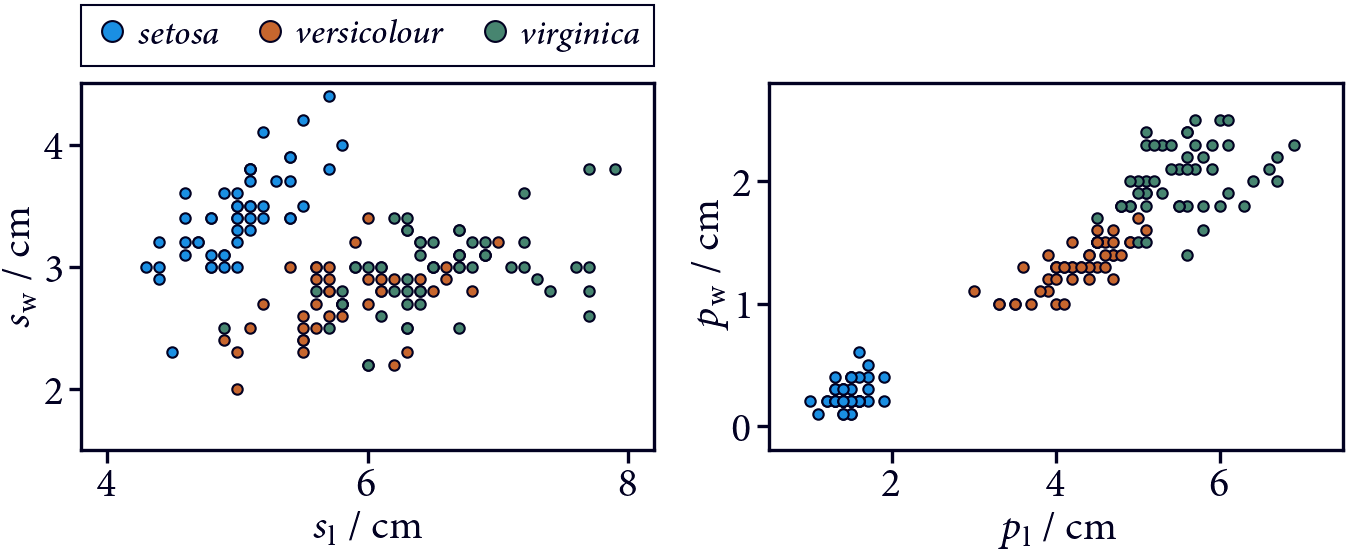}
    \end{minipage}%
    \begin{minipage}[t]{0.24\textwidth}
        \vspace{\baselineskip}
        \captionofside{figure}{%
            \textit{Iris plant} data set with biological classes}{%
            \textit{Iris plant} data set with biological classes}{%
                Data points
                plotted in the 4-dimensional feature space of \(s_\mathrm{l}\)
                (sepal length), \(s_\mathrm{w}\) (sepal width), \(p_\mathrm{l}\)
                (petal length), and \(p_\mathrm{w}\) (petal width) with their
                biological classification indicated by colour.
                }
    \label{fig:iris_data_points_true_labels}
    \end{minipage}
\end{illustration}

For the representation of this tabular data set in a computer program,
the data can take the general form of a \(n\times m\) matrix
\(D\),\stdmarginnote{point coordinates} so that there is one row for
each data point and one column for each feature and the matrix element
\(D_{ij}\) holds the \(j\)th attribute of the \(i\)th data point.
Different programming languages provide their own concrete data
structures for this.
In Python, the native choice of data structure would be a list of lists
which would look for the \textit{Iris} data like:

\pagebreak
\begin{lstlisting}
  D = [
    [5.1, 3.5, 1.4, 0.2,],
    [4.9, 3.0, 1.4, 0.2,],
    [4.7, 3.2, 1.3, 0.2,],
    [4.6, 3.1, 1.5, 0.2,],
    ...
  ]
  coordinates_i = D[i]
\end{lstlisting}
This can be replaced by a \tminline{ndarray} with the third
party library NumPy, or with a
\tminline{DataFrame} in Pandas, but
there is a wide bouquet of other structures to represent this kind of
information.
Usually, the important characteristic property of these
\enquote{matrix}-like data structures is that they are
\textit{indexable} (allows access to the element \(ij\)) or
\textit{iterable} (allows to loop over all points).
Implementations of clustering methods may make strict requirements on
how the data is stored and how individual points (or better their
coordinates) are accessed.
While for clustered objects identified with points in a (metric) data
space different cluster methods can make direct use of point coordinates,
it is often not the location of the points itself that is of interest but
only the spatial distance between them.\stdmarginnote{pairwise distances}
So instead of representing a data set as a \(n\times m\) matrix of
coordinates, it can be possible (or required) to have a \(n\times n\)
\textit{distance matrix} of pairwise distances where each element
\(D_{ij}\) holds the distance between point \(i\) and point \(j\).
In principle, the same concrete data structures as for point coordinates
can also be used for distances, so using the Euclidean distance for the
\textit{Iris} example one would have something like:
\begin{lstlisting}
  D = [
    [0.00, 0.54, 0.51, 0.65, ...],
    [0.54, 0.00, 0.30, 0.33, ...],
    [0.51, 0.30, 0.00, 0.24, ...],
    [0.65, 0.33, 0.24, 0.00, ...],
    ...
  ]
  distance_ij = D[i][j]
\end{lstlisting}
Similar to a data structure for point coordinates, central properties of
distance data structures are that individual points (or rather their
distances to other points) can be accessed by some form of indexing and
that one can iterate over all objects in the data set.
It should be noted, though, that the memory complexity for distance
matrices of \(\mathcal{O}(n^2)\) can prohibit the explicit storage of
all matrix elements for larger data sets, so that sparse data structures
should be leveraged, which for example avoid the storage of 0-valued
elements and the duplicate storage of symmetric elements.
In this way, the number of physically stored distances can be reduced to
at most \((n^2 - n) / 2\).
Distances can also be
given implicitly as a distance function while still only point
coordinates are physically stored.
A distance metric \(d : X \times X \rightarrow \mathbb{R}\) is a function
that maps two points \(x\) and \(y\) of the data space to a value so that
the following is fulfilled:
\begin{align}
    d(x, y) &\geq 0 & &\mathrm{non{\mhyphen}negativity}\label{eq:nonnegativity}\\
    d(x, y) &= d(y, x) & &\mathrm{symmetry}\\
    d(x, y) &= 0 \,\Leftrightarrow\,  x = y & &\mathrm{indiscernibility}\label{eq:indiscernibility}\\
    d(x, y) &\leq d(x, z) + d(z, y) & &\mathrm{triangle\ inequality}\label{eq:triangle_neq}
\end{align}
For sampled data in a data set \(\mathcal{D}\), the identity of
indiscernibles \ref{eq:indiscernibility} can usually not hold because a
data set can contain duplicates.
If two samples have identical coordinates, a distance between them
should be zero, but from a zero distance does not follow that two points
are the same sample so that equation~\ref{eq:indiscernibility} becomes \(d(x, y)
= 0 \,\Rightarrow\,  x = y\).
It is a practical question, if duplicate samples should be removed from
a data set prior to a clustering.
In the context of clustering, the concept of pairwise object distance is
construed within the broader concept of object similarity that will be
addressed further in section~\ref{sec:similarity_definitions}.
If distance evaluations are decoupled from the general internals of a
clustering method, it can be said that a clustering operates on a latent
space that is the actual data space can be unknown or is regarded
irrelevant for the grouping into clusters.
Taking the representation of a data set through pairwise distances one
step further, it is also possible to look at the data in terms of
neighbourhoods.\stdmarginnote{point neighbourhoods}
A clustering may depend on the information if two points are considered
neighbours of each other rather than on how far two points are away from
each other exactly.
There are two basic, practically used forms of defining neighbourhoods
derived from a distance function: \textit{fixed radius near} and
\textit{k-nearest} neighbourhoods.
Using a fixed distance (radius) \(r\), the sampled neighbours of a point
\(x\) in the data set can be denoted as the set \(\mathcal{B}_r\)
\begin{align}
    \mathcal{B}_r(x) &= \{y \in \mathcal{D}~|~d(x, y) < r\}\quad \mathrm{open{\mhyphen}ball}\\[-18pt]
    \shortintertext{or\vspace{-8pt}}
    \mathcal{B}_r(x) &= \{y \in \mathcal{D}~|~d(x, y) \leq r\}\quad \mathrm{closed{\mhyphen}ball}
    \label{eq:closed_ball_neighbourhood}
\end{align}
that is the collection of samples that lie within or below \(r\)
measured from the location of \(x\).
While it may make a fundamental mathematical difference if
\(\mathcal{B}_r\) is defined to be the open-ball or closed-ball around
\(x\), it is usually considered a technicality when it comes to
clustering.
The same goes for the question if \(y = x\) should be contained in the
neighbourhood or not, which means if formally a point \(x\) is its own
(closest) neighbour.
Note that the neighbourhood of a point is in this sense directly the near
neighbourhood with respect to \(r\) and not any set of points containing
\(\mathcal{B}_r\).
Instead of a fixed distance, the \(k\) nearest samples to a point
can be used to define neighbourhoods as the set \(\mathcal{B}_k\)
\begin{align}
    \mathcal{B}_k(x) &= \{y \in \mathcal{D}~|~d(x, y) \leq d_k(x)\}
    \label{eq:k_nearest_neighbourhoods}
\end{align}
where the \(k\)-nearest distance \(d_k\) from a point \(x\) is the
distance for which there are \textit{at least} \(k\) points at \(d(x, y)
\leq d_k\) and \textit{at most} \(k - 1\) points at \(d(x, y)
< d_k\).
Again it may be optionally required that \(y \neq x\).
Other definitions of neighbourhoods---derived from distances or
not---are conceivable.
For the choice of data structures to represent neighbourhood information
there are multiple options among which there are two fundamentally
different approaches.
The first would be to use a \(n \times n\) matrix comparable to a
distance matrix but with binary entries indicating pairwise neighbour
relationships in a true/false manner, which could result in the
following for the \textit{Iris} example when a fixed radius \(r = 0.52\)
is used on the previously shown distances:
\begin{lstlisting}
  D = [
    [1, 1, 0, 1, ...],
    [1, 1, 0, 0, ...],
    [0, 0, 1, 0, ...],
    [1, 0, 0, 1, ...],
    ...
  ]
  is_neighbour_ij = D[i][j]
\end{lstlisting}
This type of matrix can be called an \textit{adjacency matrix}.
Since it contains a lot of 0-valued elements in some cases, this type of
information is prone to be stored in sparse data formats.
Alternatively, it is possible to represent the same information in the
form of a sequence of neighbourhoods, for example by
keeping a list of point indices to refer to the neighbours of each
point:
\begin{lstlisting}
  D = [
    [0, 1, 3, ...],
    [0, 1, ...],
    [2, ...],
    [0, 3, ...],
    ...
  ]
  neighbours_i = D[i]
\end{lstlisting}
The term \textit{neighbour list} is sometimes used in this context but
it is not exactly optimal because it can be ambiguous whether it refers
to the neighbours of an individual point (i.e. a single neighbourhood)
or the full list of neighbourhoods.
It is usually again an important property of these data structures that
points (that is their neighbours) can be indexed and that one can
iterate over all points.
As it is the case for distances, it may be possible that neighbourhood
determination can be decoupled from the actual clustering.
A universal pattern behind the representation of a data set either in
terms of object attributes (point coordinates), pairwise distances, or
neighbourhood relations, is to understand a data set as a graph in which
each object is a vertex and distances or neighbour relations are
indicated by possibly weighted edges.\stdmarginnote{graph data}
Thinking about data sets as graphs is really more a mindset and it is
not tied to a specific data structure.
In fact, all the presented matrix-like data structures can be viewed as
manifestations of graphs making the node-edge nature of data objects
more or less explicit.
The graph picture is more general than a certain conceptualisation
of a specific type of information because it unifies them all.
A node in a graph, i.e. an object in the data set, may have attributes
that may in turn correspond to coordinates in a metric space---but this
is not actually required.
Edges between nodes can encode distances or neighbourhood relations,
which may be derived from object coordinates or may be the primary
source of information, as a general form of connectivity but this is
again optional (the graph does not need to be complete).
Note that the terms distance, adjacency, and as we will see later
(dis)similarity or also \textit{affinity}, and the associated matrix
types can become a little bit blurry in general here.
All have in common that they correspond to edge weights of an (implicit)
graph and while it can make sense to use one of them over the other
depending on the situation, they are also often used interchangeably.
It depends on the clustering method if it is practically advantageous to
think about the data as a graph and which kind of information is captured
in it.
On the other hand, there are certain data structures that are especially
valuable for the representation of graph data because they alleviate
certain operations beyond the access of individual points and iterations
over the data set.
In particular, these operations are membership lookups, the addition and
removal of data objects, the splitting and merging of data (sub)sets,
and the modification of inter-object relations.
A simple, explicitly graph-like representation of the \textit{Iris} data
using a Python dictionary for the set of nodes and a Python set for the set
of edges (indicating binary neighbourhood connections) could look like
this:
\begin{lstlisting}
  D_nodes = {
    0: [5.1, 3.5, 1.4, 0.2,],
    1: [4.9, 3.0, 1.4, 0.2,],
    2: [4.7, 3.2, 1.3, 0.2,],
    3: [4.6, 3.1, 1.5, 0.2,],
    ...
  }

  D_edges = {
    (0, 1), (0, 3), ...
  }
\end{lstlisting}
Many specialised implementations of graph data structures are available
in different programming languages, the \tminline{networkx}
module being only one example for a third party Python package for
exactly this purpose.
A plot of the \textit{Iris} data in terms of a graph is shown in
figure~\ref{fig:iris_data_graph}.

\begin{illustration}
\begin{minipage}[t]{0.5\textwidth}
  \vspace{0pt}
  \includegraphics[trim=20 15 20 15, clip, width=\textwidth]{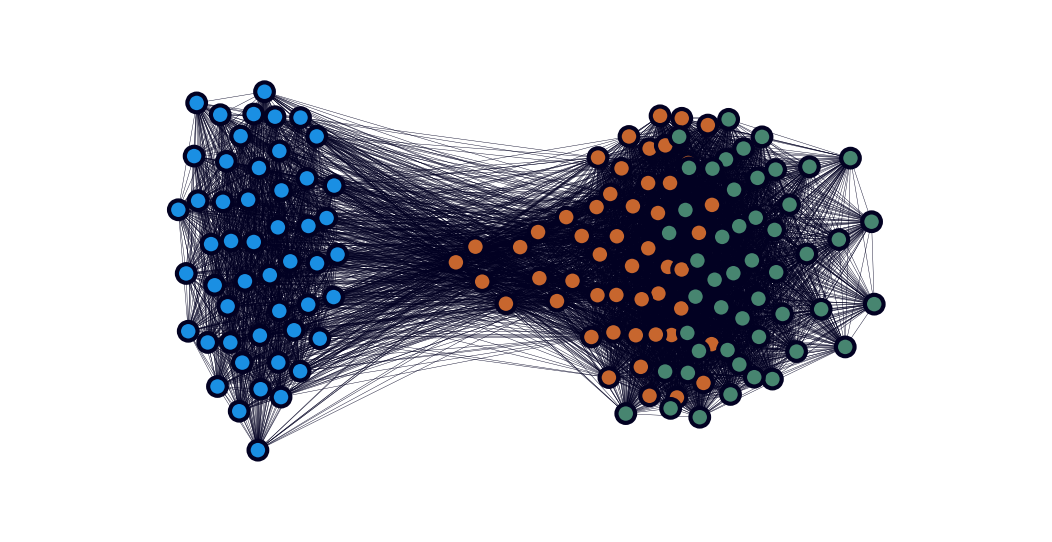}
\end{minipage}%
\begin{minipage}[t]{0.48\textwidth}
  \vspace{0pt}
  \captionofside{figure}{%
    \textit{Iris plant} data set as a graph}{%
    \textit{Iris plant} data set as a graph}{%
        Data points plotted as nodes in a graph coloured by their
        biological classification. The edges are scaled in length by
        weights corresponding to the Euclidean distance between the data
        points. Distances \(d(x, y) > 3\) have been omitted.}
  \label{fig:iris_data_graph}
\end{minipage}
\end{illustration}

\custompar{We stated that the \textit{Iris} data} feature space is a real space, and
so it can be the case for other data sets, in particular for \gls{md}
data where the considered data space is often a configurational space
with (projected) cartesian coordinates.
Generally, however, the feature space associated with a data set does
not have to be restricted to real valued components as the records in a
data set can be composed of basically arbitrary
variables.\stdmarginnote{non-numeric features}
The graph perspective on the clustered objects becomes even more
appropriate then because graph nodes can be equipped with any attribute
and non-standard attribute combinations do not need to be forced into a
strict matrix-like layout.
Care needs to be taken, however, in these cases about how basic
mathematical operations can be defined.
Without going into great detail of working with non-numeric data and the
problems that it can involve, here are a few general points to keep in
mind.
It is often possible and good advice to convert the features of a data
set in a way that makes them numeric.
For example, binary categorical values (e.g. yes/no, true/false,
present/absent) can usually be represented by \(0\) and \(1\)
respectively.\stdmarginnote{binary features}
It just may need to be considered that to compute a distance between
points in a binary space, the Euclidean distance may be not very
appropriate and other distance measures should be preferred.
Take for example, the following data set of three points with four binary
attributes
\begin{lstlisting}
  D =[
    [1, 0, 1, 1],
    [0, 0, 0, 1],
    [0, 0, 1, 1],
    ]
\end{lstlisting}
The Euclidean distances of \(d_{0,1} = \sqrt{2}\) and \(d_{0,2} = 1\)
between the points are not very intuitive while the Manhatten (or in
this case equivalently Hamming) distances \(d_{0,1} = 2\) and \(d_{0,2}
= 1\) can be well interpreted as the number of features in which two
points differ.
Things become more difficult with nominal categorical values (e.g.
countries DE/FR/GB) where the only valid operations are absolute
comparisons (e.g. \(\mathrm{DE} = \mathrm{DE}\),
\(\mathrm{FR} \neq \mathrm{GB}\)).\stdmarginnote{nominal features}
A common approach to transform these, is to introduce a number of dummy
features, one for each categorical value that a nominal feature can adopt.
This is called a one-hot encoding.
A single nominal feature comprising three categories would be replaced
by three binary features like shown here with an example of four data
points:
\begin{lstlisting}
  D_nominal = [
    ["a"],
    ["b"],
    ["c"],
    ["a"],
    ]

  D_one_hot = [
    [1, 0, 0],
    [0, 1, 0],
    [0, 0, 1],
    [1, 0, 0],
  ]
\end{lstlisting}
It is now possible to use the Manhatten distance again so that points
differing in one nominal feature have a distance of \(d = 2\) from each
other.
Note, however, that the Hamming distance works well with categorial data
even without the encoding.
There are other possible comparisons, e.g. via the S{\o}rensen-Dice or
Jaccard index, just to name a few.
It is arguably most difficult to adequately represent ordinal,
categorical values (good / neutral / bad, low / rather low / medium / rather
high / high).\stdmarginnote{ordinal features}
These can be either treated in the same way as nominal values which
will, however, ignore the fact that there is an ordering in the
categories and distances between each pair of categories are not all
equal (e.g. \(d(\mathrm{good}, \mathrm{bad}) \neq d(\mathrm{good},
\mathrm{neutral})\)).
Alternatively, they can be mapped to numerical values (e.g. good: 3, neutral: 2,
bad: 1) to reflect the ordering (that is \(\mathrm{good} >
\mathrm{neutral} > \mathrm{bad}\)) but it may be difficult to ensure
that the numeric distances (e.g. here \(d(\mathrm{good}, \mathrm{neutral}) =
d(\mathrm{bad}, \mathrm{neutral}) = 1\)) are aligned with the true, intrinsic
distance of the categories.
For special data objects or attribute values, it may be furthermore
necessary to select specialised treatments of comparisons.
There is for example the Levenshtein distance to compare strings.
Especially problematic may be mixed data spaces with partly numeric,
partly categorical features (more on this further below).
\custompar{A typical problem with feature spaces} can occur when
individual features have very different dynamic ranges, or different
physical dimensions or meaning.
In particular, this can be the case for mixed data.
Let's first have a look at the influence of feature
ranges.\stdmarginnote{feature ranges}
Figure~\ref{fig:iris_data_standardise} shows the histogramed features of
the \textit{Iris} data set and it can be noticed that the distributions
for the lengths features (\(s_l\), \(p_l\)) cover larger intervals than
the respective width features (\(s_w\), \(p_w\)).
For the calculation of distances between data points, this can have the
consequence that a difference with respect to a length has a larger
influence than a difference with respect to a width, just because widths
differences are consistently smaller than length differences.
In other words, individual features may dominate distance measures.

\noindent
\begin{illustration}
\begin{minipage}[t]{0.6\textwidth}
    \vspace{0pt}
    \includegraphics[trim=1 0 0 0, clip]{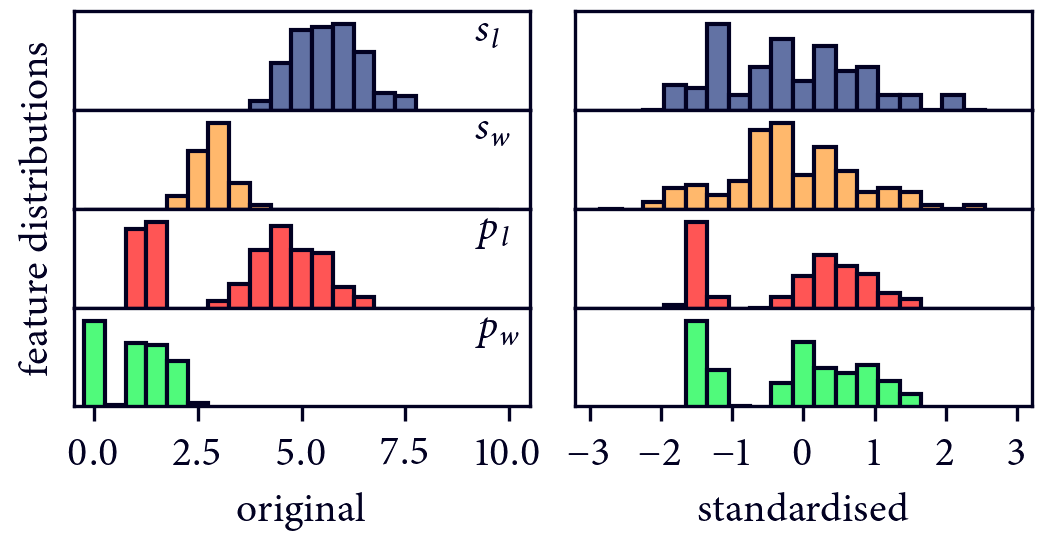}
\end{minipage}\hspace{0.01\textwidth}%
\begin{minipage}[t]{0.39\textwidth}
    \vspace{-0.25\baselineskip}
    \captionofside{figure}{%
    \textit{Iris plant} data feature standardisation}{%
    \textit{Iris plant} data feature standardisation}{%
        Feature distributions before (left) and after (right)
        standardisation of the features through z-scaling, i.e. removal
        of the mean and normalisation by the standard deviation. Note
        that it is \textit{not} advised to do this in this case.}
    \label{fig:iris_data_standardise}
\end{minipage}
\end{illustration}

To counter the effect of disproportionate distribution ranges of
individual features, we can apply a standardisation that will in some
sense make each feature equally important.
A typical standardisation protocol is found in the so called z-scaling.\stdmarginnote{z-scaling}
This will transform feature values (here \(x_j\) is the \(j\)th
component of a single data point) by subtracting the mean of the
respective feature \(\mu_j\) and dividing by the feature's standard
deviation \(\sigma_j\)
\begin{equation}
    x_j^\prime = \frac{x_j - \mu_j}{\sigma_j}\,.
    \label{eq:z-scaling}
\end{equation}
As the result, each standardised feature will have a mean of \(\mu_j^\prime = 0\)
and a standard deviation of \(\sigma_j^\prime = 1\).
Alternative protocols are min-max scaling\stdmarginnote{min-max scaling}
that normalises a feature to a given closed interval \([\min_j{^\prime},
\max_j{^\prime}]\) (e.g. \([0, 1]\)) by subtracting the minimum feature
value and dividing by the feature range
\begin{equation}
    x_j^\prime = \frac{x_j - \min_j}{\max_j - \min_j} (\max{_j}{^\prime} - \min{_j}{^\prime}) + \min{_j}{^\prime}
\end{equation}
and max-abs scaling\stdmarginnote{max-abs scaling} that divides by the maximum
absolute feature value and effectively normalises to the interval \([-1, 1]\)
\begin{equation}
    x_j^\prime = \frac{x_j}{\max(|\max_j|, |\min_j|)}\,.
\end{equation}
The question is now, if a standardisation or normalisation is actually
advisable for the \textit{Iris} data set.
All features in this set have the same physical dimension and by
re-scaling the features, the relative proportions of the features are
distorted.\stdmarginnote{always normalise?}
By scaling the widths to the same ranges as the lengths, we equalise
their influence on distance calculations but this may lead to
false interpretations because the influence of widths and lengths is
not the same in reality.
Another way to think about it, is that both width and length are coupled
quantities that are two sub-features of say an area-feature.
By re-scaling the sub-features the jointly described feature is skewed.
The \textit{Iris} data set is therefore actually a counter example in which
case a strict feature-wise standardisation is not recommended.
If the ranges are to be scaled, this should be done proportionally.
It would be a possibility to scale the sepal features (\(s_l\), \(s_w\))
and the petal features (\(p_l\), \(p_w\)) as separate pairs, equalising
the ranges corresponding to different leafs while maintaining the
proportions of features describing the same leaf but this is still
debatable.
Another example in which case independent feature-wise normalisation is
not appropriate are molecular configurational spaces given as let's say
\(m\) real-valued features.
It is possible to scale all \(m\) dimensions proportionally to a new
range but setting all of them individually to the same range distorts
the structural information of the data.
Feature standardisation and normalisation should only be applied if
separate features have very different meaning, for example \enquote{age},
\enquote{height}, and \enquote{weight} in a data set containing samples
of different persons.
These features may cover very different ranges (e.g. height: \([150,
200]\), age: \([30, 60]\), etc.) and a combined distance measure to which
each feature should contribute equally calls for a re-scaling of each feature
to say the interval \([0, 1]\).
Different feature ranges are especially encountered in mixed data sets.
An example would be a molecular feature space that contains not only
real-valued dimensions but also binary features (e.g. interaction
indicators).\stdmarginnote{mixed data}
The binary features are naturally confined to the interval \([0, 1]\)
and to equate their influence on distance calculations, a proportional
re-scaling of numeric spatial coordinates may be an option.
We face, however, a much more crucial problem here because the question
is, what would be a good distance measure to combine binary indicators
with spatial coordinates?
Obviously, using the Euclidean distance to which a binary feature
contributes a value of 0 or 1 would be difficult to balance against the
influence of the numeric features, that is it would be hard to decide
what the relative importance of the presence or absence of a single
interaction is compared to a structural change along a specific spatial
dimension.
In this case it may be advisable to use a set of different distance
metrics suitable for a comparison along only a subset of the features
and to average over the weighted contributions from these distances.
Such an approach of computing partial distances that can be re-combined
into a single distance is formalised for example in the Gower distance.\cite{Gower1971,Bishnoi2020}
In the \textit{Iris} data set, each record carries a categorical value
that associates it with a biological flower class besides the four measured
numerical features.
This nominal feature can in principle be included into the feature space
as well if it is seen as a source of information on which basis the data points
should be grouped into clusters.\stdmarginnote{reference labels}
Here we want to treat the class field differently, namely as a source of
reference.
We can interpret the points of the data set as labelled data and treat
the class-field as a true classification, which can be compared
to classifications that are based on a clustering.
In other words, we can use the class field for an external validation of
clusterings for the \textit{Iris} data and we will consider a clustering
good if it is able to reproduce the reference labels.
This is based on the premise that firstly the class field represents
indeed true information, i.e. we assume that there is no mistake in the
assigned biological class, and second we will presume that it is actually
possible to reproduce the reference labels using the information of the
remaining features.
As a last remark, it would also be imaginable to use the ordering of the
records (the record ID in table~\ref{tab:iris}) as a feature on its own.
In this particular case it might be not very meaningful because the ordering
of the samples is arbitrary.\stdmarginnote{geometric vs. kinetic clustering}
It may, however, be the case in other data sets that the ordering encodes
information itself, as for example in \gls{md} data were each data point
is a structural snapshot at a given point in time.
Clusterings disregarding the ordering of data points that only use the
positions of objects in a feature space can be referred to as geometric
clusterings while kinetic clustering on the other hand tries to
incorporate the temporal relation between clustered objects.
No matter how we choose to represent objects in a data set to be
analysed by a clustering and no matter which object properties we consider
for this, a clustering \(\mathcal{C} = \{C_1, ..., C_k\}\) is always a
decomposition of \(n\) objects into \(k\) subsets
\((C_j \in \mathcal{C}) \subseteq \mathcal{D}\).
In the next section we will discuss different clustering approaches in a
general fashion while selected methods and algorithms are presented in
chapter~\ref{chap:clustering_methods}.

\section{Definitions of similarity and clustering categories}
\label{sec:similarity_definitions}

\firstsecpar{At the beginning of this chapter}, clustering was described as an
analytical process to identify groups of objects that are characterised
by strong relationships between objects within the same group and weak
relationships between objects in different groups.
This conception is wilfully very open and a relationship
can be basically anything.
Practical clustering procedures need to transform it into something
tangible and need to specify how relationships are actually defined and
determined.
Based on how this transformation is done and how the clustering works,
clustering procedures can be roughly
categorised.\stdmarginnote{clustering categorisation}
These categories should, however, not be seen too dogmatic and are rather an
orientation.
Throughout the literature, substantially different categorisations are
proposed---and criticised.\cite{Peng2018,Henning2016,Estivill-Castro2002,Jain2004}
We already encountered one form of categorisation previously in this
chapter with the assessment of how clusterings assign labels to objects
(e.g. in a strict or fuzzy manner).
This is a very practical way to look
at it but it tells us little about the actual clustering.
In a way, it is primarily a modelling decision if a clustering uses a
certain type of label assignment and methods can be flexible in this
regard.
Clustering methods can be understood as being comprised of three
different aspects.
The heart of each clustering technique is (or rather should be) a
\textit{model} that is some idea of what potential clusters in a group
of objects actually are, although the model may not be always obvious
for every method.\stdmarginnote{cluster model}
Categorisations based on cluster models are popular and we will address a
few of them.
Most paramount are \textit{connectivity}- and
\textit{prototype-based} models.
It should be noted, though, that the borders between models can be
blurry and that their association with a specific category may only
reflect or emphasise one model property.
A good way to describe the cluster model of a given method can also be to
think about the kind of output that is produced.
A certain model can be paired with an \textit{inductive principle}, which
may be either a mathematical formalism---an \textit{objective function}
that a model should strive to optimise---or rather a more or less formal
description of how a model should be
constructed.\stdmarginnote{inductive principle}
Such an inductive principle provides an opinion on what is a \textit{good}
clustering.
Explicit categorisations of inductive principles are rather rare, which
may be owed to the fact that for many clustering methods the underlying
principle is not exactly easy to grasp so that the focus simply shifts
towards the model.
When an inductive principle does not manifest itself in a mathematical
formalism, it may also become unclear where the model and the underlying
principle differ.
Inductive principle and model need to match each other but in principle
the same model can be evaluated in the light of different objectives
and one inductive principle may apply to different kinds of models.
Lastly, a clustering method needs to be realised through an
\textit{algorithm} while in turn different \textit{implementations} are
possible for the same algorithm.
If a clustering is actually applied, it is an implementation of an
algorithm that eventually needs to be chosen for
it.\stdmarginnote{algorithms}
Clustering algorithms and their concrete implementations
produce a model for the clustered data while an inductive
principle suggests what the best model should be.\cite{Estivill-Castro2002}
Methods can be exact in their formulation of an underlying principle but
concrete algorithms may in many cases only provide approximations.
Especially, when there is no concise mathematical formulation for what
an algorithm should yield, algorithms are only heuristic protocols to
construct good models.
Algorithms are in general very convertible and the same context (model
and inductive principle) can lead to a variety of algorithms and
realisations.
Categorisations of clustering methods that claim to be focused on the
model or the underlying principle are sometimes scrambled with a focus
on algorithmic or implementation details, that is in particular the type
of data structures that are used or produced.
Examples are the categories of \textit{graph based} or \textit{grid based}
models.
Graph based puts an emphasis on the fact that the output of a clustering
is a graph or that the method explicitly uses a graph representation of the
data.
With some effort, however, most if not all clustering methods could be
transformed to use some kind of graph structure---particular ones may
just be more obviously suited to be expressed in graph theoretical
terms.
We would therefore argue that a description of a model as graph based
(possibly as a subcategory of connectivity-based) is not very
informative and foremost a property of the algorithmic formulation or
implementation.
Whether a clustering method uses a form of grid, is also more a question
of procedure rather than of what the model for the data is.
Of course it can be the case that the cluster model decidedly is grid
based and a grid is the final result of a clustering (possibly with the
grid cells as a certain kind of prototype) but in the way it is
normally used, the grid plays often a rather auxiliary role (compare
section~\ref{sec:density_based_grid}).
Many clustering methods can be formulated with or without involving a
grid.

\custompar{A prominent concept to substantiate object relationships} is
to leverage the term \textit{similarity}.
Similarity is in fact so abundantly used that many alternative
definitions of clustering include it in their basic
formulation.\stdmarginnote{object similarity}
It can for example be found that the aim of clustering is to
identify groups among objects that \textit{maximise intra-cluster
similarity} and \textit{minimise inter-cluster
similarity}.\cite{Estivill-Castro2002,Han2000}
We think that such a statement can be problematic because of two
reasons: first, against what is suggested here, clustering can by far
not always be expressed as an optimisation problem in which a certain
quantity is minimised or maximised.
It may of course well be true that clustering is intrinsically an
optimisation (or at least that an objective to optimise for is in general
desirable) and that an objective function can just not be established or
discovered, or is too hard to solve.
But the statement does in this form
not aptly describe clustering in reality where it can be studied from a
rather procedural level.
And second, while \enquote{similarity} is used here without obligations,
the term has actually a distinct meaning and by far not every clustering
approach does rely on it literally.
We have to be careful to not bloat the concept of similarity to
an extend where it just replaces \enquote{relationship} in the broad
initially proposed definition and becomes essentially hollow.
It remains legitimate to say, that clustering has the aim to identify
groups of \enquote{similar} objects when it is supposed to mean
generally \enquote{alike} or \enquote{somehow related} but similarity as
such should probably be treated with a bit of caution.
Clustering procedures that use similarity as an underlying idea are
based on a similarity function \(s(x, y)\) that maps two objects to a
single value.
Similarity is something that can be measured and compared
quantitatively, which means it can take either a low or high value.
Instead of similarity, the same can be expressed inversely using a
\textit{dissimilarity} function \(d(x, y)\).
Such a function can be just a common distance metric (compare
eq.~\ref{eq:nonnegativity} to \ref{eq:triangle_neq}).
A short distance between objects with respect to the space they are
embedded in corresponds to low dissimilarity, i.e. high similarity.
Because of the frequently made analogy between dissimilarity and
distance, \textit{proximity} can be used synonymously to similarity.
It should be noted, however, that similarity measures are not
necessarily always classic distance measures.
Similarity and dissimilarity functions can be much less formal than
distance metrics in the sense that neither non-negativity, symmetry, nor
satisfaction of the triangle inequality are strictly required, although
especially symmetry may be generally helpful.
Possible measures of similarity include divergences, correlations,
mutual information, kinetic distances, but also generic notions of
pairwise object connectivity as the number of shared neighbours (like in
the \gls{commonnn} scheme described in section~\ref{sec:commonnn_theory}).
It can be also something arbitrary like the number of
text messages that are send from user to user in a social network.
Thinking about data sets in terms of graphs, similarity can be expressed
as anything that can serve as the weight of an edge connecting two nodes
in the graph.
Similarities can be confined to arbitrary value ranges.
They can for example be expressed in a binary form, i.e.
\(1\equiv\mathrm{\enquote{similar}}\),
\(0\equiv\mathrm{\enquote{not~similar}}\).\stdmarginnote{similarity values}
Normalised distances (dissimilarities) can be converted to a similarity
via \(s(x, y) = 1 - d(x, y)\), so that similarity is restricted to the
interval \([0, 1]\), but it is also possible to have similarities in the
interval \([0, \infty]\) with a relation \(s(a, b) = d(a, b)^{-1}\)
(with \(d(a, b) = 0 ~\Rightarrow~ s(a, b) = \infty\)).
In principle, similarities can be obtained through any (non-linear)
mapping of distances, say radial functions like for example a generalised
Cauchy function \(s(x, y; a, b) = 1 / (d(x, y)^a + b)\).
A combined similarity/dissimilarity measure could furthermore take on
values in an interval as say \([-1, 1]\).
Clusterings that employ a similarity concept, can produce models in
accordance with the idea that based on the similarity of
two objects \(x\) and \(y\), it can be decided if they should be part of
the same cluster.
One should refrain, however, from the tempting conclusion that in
reverse two points \(x\) and \(y\) are similar (in the sense of a high
pairwise similarity value) in general if they are in the same
cluster.\stdmarginnote{connectivity clustering}
Also, an actual maximisation or minimisation of the similarity can not
be claimed for all similarity based clustering methods.
The property of linking pairs of objects through
similarity, affiliates methods with the
\textit{connectivity-based} category of clusterings.
To use a set of representative connections between objects is a kind of
model for a data set where clusters are groups of inter-connected
objects.
Inductive principles for connectivity-based models can select a set of
relevant connections (based on similarities) and can be rather subtle.
Conceptually, pairwise similarity to guide clustering models
can for example take a strong and a weak form.
If for any point \(x\) in the data set that is a member of cluster
\(C_i\) the condition is fulfilled that for any other point \(y\), also
a member of \(C_i\), the similarity \(s(x, y)\) is larger than the
similarity \(s(x, z)\) of \(x\) and any other point \(z\) in cluster
\(C_j\) with \(i\neq j\), this clustering contains strongly similar
groups.
In a graph picture of the data, this can be related to the identification
of strongly connected sub-graphs.
In other words, the strong version requires that points within the same
cluster are strictly more similar to each other than to points in
different clusters.
Note that this condition can be fulfilled for more than one grouping of
the same object set---in particular it is true for the trivial cases of
having only one group (no actual clustering) and having each object in
its own group (over clustering).
For an actual optimisation of some sort, additional requirements would
need to be made as for example that the minimum similarity within a cluster
should be as large as possible (that is the maximum dissimilarity between
two points within the same cluster is minimal).
The weak version of this principle, on the other hand, would only require
that for any point \(x\) in cluster \(C_i\) the most similar point
\(y\) in the data set is also a member of cluster \(C_i\).
In the graph picture, this is equivalent to a search for connected
components.
Note again, that without any further specification this condition can
be fulfilled for more than one grouping of the same object set, as well.
A connectivity-based element can be found in many clustering methods and
their models.
A common trait of such models is that they do not necessarily make any
assumptions about the shape, size, number, or spreading of the clusters
that should be found in a data set.
These characteristics will be just an implication of how individual
objects are connected to each other.
Connectivity-based clustering has a predisposition to be formulated
using graph concepts.
As mentioned already, clusters can be understood in this way as
connected sub-sets within the graph for the whole data set.
The term \textit{connected component} reoccurs multiple times in the
context of connectivity-based clustering.
A universal clustering paradigm within the framework of data graphs
is\stdmarginnote{graph sparsity} to state that the aim of clustering is to
\textit{maximise within-cluster density} and to \textit{maximise
between-cluster sparsity}. where in this context within-cluster density
means many significant connections of objects within the same group and
between-cluster sparsity refers to few connections of low importance
between objects in different groups.\cite{Gaertler2005}
Pairwise similarities between objects are used for example in agglomerative
clustering, a family of clustering methods that will be discussed in
section~\ref{sec:linkage_clustering} in more
detail.\stdmarginnote{similarity clustering}
This clustering approach uses, in addition to similarity, a so called
\textit{linkage} criterion that states how similarity is further treated
to assign objects to the same group.
Interestingly, agglomerative clustering does in general not maximise the
within-cluster similarity while minimising the similarity between
clusters overall in the sense of an optimisation according to a global
objective function.
Instead it provides just an iterative protocol that yields a hierarchy of
clusterings by making a series of locally optimal decisions in the sense
of the linkage formulation.
As such, the model it produces for the data is a tree of clusterings in
accordance with a connectivity-based principle.
A single optimal solution can be selected from the model hierarchy only
by further requirements for example on the number of identified clusters
or using a threshold on the similarity.
Another broad family of clustering methods that make use of pairwise
similarities is spectral clustering, which will be briefly described in
section~\ref{sec:spectral}.
Through eigenvalue decomposition of a similarity matrix, precisely its
graph Laplacian, the input data is embedded into a lower dimensional
space that can reveal the cluster structure.
The objects in a data set can then be clustered in this embedding using
a method of choice.
Therefore, spectral embedding is arguably only a helpful transformation
that facilitates a clustering and not an actual clustering technique
itself.
Conceptually, it is tightly related to the general idea of rearranging a
similarity matrix into a block-diagonal form in which clusters appear as
blocks of elements with high similarity values.
Besides clustering methods with a connectivity-based model of how
clusters should be identified, which is more or less directly tied to
object similarity (i.e. connections can be found based on a similarity
definition---a connection can be seen as a form of similarity), there
are methods that have a different idea about this.\stdmarginnote{homogeneity}
One widely-used alternative approach is to infer that a data set can be
described as a composition of a number of equally behaved
clusters.
The paradigm for finding clusters can for these methods sometimes be
expressed as a \textit{maximisation of within-cluster homogeneity}
or generally as the idea of splitting
an inhomogeneous data set into more homogeneous subsets.
A similarity concept may play a role here as well but similarity between
objects is not fundamental.\stdmarginnote{prototype clustering}
Because of the shift in the notion of object relationship from pairwise
similarity to a comparison of objects with some shared
representative, these methods can be
categorised as \textit{prototype-based}.
Such a method could for example hypothesise that a data set could be
modelled by a set of \(k\) representatives \(\mu_i\) so that the following
objective function is satisfied:
\begin{equation}
  \substack{\mathrm{min}\\\mathcal{C}} \left\{\sum_{i=1}^k \sum_{x \in C_i} d(x - \mu_i)\right\}\,.
  \label{eq:partitioning_objective_function}
\end{equation}
This objective states that the best model for the data is the one that
partitions the objects in such a way that the sum over some distance
function computed for all points within a cluster and the cluster
representative is minimal over all clusters.
Each data object
will be associated to its best fitting representative.
A popular variant of this is the \(k\)-means method when the
distance function \(d\) is the squared Euclidean
distance and the used representatives are cluster centroids
(\textit{centroid based} is a special type of prototype-based), meaning
the arithmetic mean points of each cluster.
This method will be addressed in section~\ref{sec:kmeans}.
In contrast to connectivity-based methods, prototype-based clustering
does commonly make more or less strict assumptions about the number,
shape, size, and general distribution of ideal clusters.
Another example for a prototype-based clustering approach are
\glspl{gmm} also known under the family of expectation
maximisation algorithms (see section~\ref{sec:gmm} for an example).
Sometimes these will be explicitly referred to as \textit{distribution
based} models.
The idea behind these, is that a data set can be modelled by a set of
overlapping normal distributions.\stdmarginnote{expectation maximisation}
Each data point is assumed to be sampled from one of these underlying
distributions (the prototype they will be associated with) and the aim
of the clustering is to identify the set of distributions that represents
the observed data in the best possible way.
The inductive principle can in these cases often be formulated as
a likelihood maximisation.
Other related methods may take different kinds of distributions as their
basis.
Here it becomes particularly clear that similarity between objects plays
a very subordinate role and is overruled by the aim to find homogeneous
groups within the data of which each can be described by a single simple
distribution.
In categorisations, these clustering methods are sometimes also called
\textit{model based}, which, however, makes little sense before the
background that essentially every clustering implies a cluster model of
some form.
Yet another principle of clustering models, which should be especially
important to us, comprises the category of \textit{density-based}
clusterings.
The central idea from this point of view is that clusters cover regions
in the data space in which the object density is relatively high, and
that are separated from each other by relatively low
density.\stdmarginnote{density clustering}
Already early on in the history of clustering, this was taken as
a \enquote{natural} notion of clusters.\cite{Carmichael1968}
Presumably, it is the \enquote{discontinuity in \enquote{closeness}}
that makes us intuitively perceive groups of objects when we see them.
A density-based model provides a prescription for how object density can
be estimated and consequently how clusters of high density can be
identified.
As such it is normally predicated that the data set is sampled from an
unknown underlying probability distribution.
For the actual clustering, it is, however, usually not enough to have a
solely density-based model.
Eventually, a notion of how object density is evaluated needs to be
paired with either a connectivity-based element so that clusters are
identified as pairwise connected (groups of) objects of relatively high
density, or with a prototype-based element so that cluster
representatives are aligned with data regions of high density.
Prototype-based models can be described as being focused on mode
seeking, which means they usually try to identify the maxima of
a data distribution.
Note, however, that density-based clustering in general is sometimes
referred to as \textit{modal clustering}.
Examples for prototype-based, density-based clusterings are mean-shift
and density-peaks (see section~\ref{sec:density_peaks}).
Connectivity-based models on the other hand can be related to the
concept of \textit{level-sets} (see section~\ref{sec:level_set_method})
of the approximated probability density.
From this perspective, clusters are universally the connected components
of a super level-set.
Examples for connectivity-based, density-based clustering methods are
DBSCAN (section~\ref{sec:dbscan}), Jarvis-Patrick
(section~\ref{sec:jp_clustering}) and \gls{commonnn} clustering
(section~\ref{sec:commonnn_theory}).
As connectivity-based clustering in general, these clusterings make
no assumptions on the shape, size, or number of identified clusters.
Since the notion of level-sets is intrinsically hierarchical, they also
produce in essence a hierarchical tree of clustering results much like
in linkage clustering.
Taking a short detour, density-based clustering schemes are also most
valuable for the identification of molecular conformational states from
molecular simulations.
The notion of a \textit{conformation} as an ensemble of molecular
structures associated with a potential energy minimum (a region of high
Boltzmann density) separated from other minima by transition regions (of
low Boltzmann density) has a very natural and intuitive correspondence
to the density-based clustering assumption.\stdmarginnote{molecular
conformations}
The property of these clusterings that they find the high density
regions independent of the spatial layout of these regions allows to
identify molecular conformations truly as what they are, detached from
purely structural information.
A conformational ensemble can be structurally quite diverse as long
as the respective atomic arrangements are separated by no (or very low)
energetic barriers.
Although density-based clustering is in principle geometric, i.e. does
not make explicit use of temporal information, the result in a molecular
context can be interpreted as kinetically relevant.
A low energetic separation of molecular structures can be equated with
fast interconversion, whereas conformations identified as different
clusters will be in relatively slow exchange.
The definition of a conformation as being associated with an energetic
minimum is of course somewhat blurry (especially for complicated
potential energy surfaces), but so is the definition of density-based
clusters.
The complete conformational ensemble of a molecule can be seen as a
hierarchy of sub-ensembles.
With increasing granularity, one can split the whole ensemble first into
the most clearly separated sub-ensembles (the ones with the highest
barriers in between them) and continue by further splitting along the
smaller separations.
Along the way, individual conformations can be considered with varying
extent, shrinking from the complete region they exist in up to the
transition boarders, down to a splitting point or until they are reduced
to only the closest structures to the minimum.\stdmarginnote{core sets}
The term \textit{core set} is often used in this context to denote those
data points that are a \enquote{sure} member of a cluster.

\custompar{Sometimes, clusterings are categorised} by the fact whether
they produce a \textit{hierarchical} or \textit{flat}
result.\footnote{Instead of flat, the opposite of hierarchical is
regularly also referred to as \textit{partitional}. We will avoid this,
however, since there can be some confusion with the term
\textit{partition} used to refer to a crisp, non-overlapping, possibly
also exhaustive clustering (e.g. full partitioning vs. partial
clustering).}
It is true that individual clustering methods can be
predominately\stdmarginnote{hierarchical vs. flat} hierarchical or flat:
agglomerative clustering is intrinsically hierarchical while
\textit{k}-means gives a single flat partitioning.
We argue, though, that it is not profound to a method to be either
of it but that it is rather a consequence of the existence or missing of
an explicit objective function according to which a single clustering
result can be optimised.
Flat clusterings can be often applied quasi-hierarchical by reusing
the method on a subset of the data identified as a cluster.
Hierarchical clusterings can be used quasi-flat by taking a certain
level of the hierarchy as a single partition, i.e. by for example applying
a termination criterion.
\Gls{commonnn} clustering is a good example for a method that can be
practised in a flat manner (using a specified density threshold) or
fully hierarchically.
Another form of categorisation judges clustering methods by the parameters
that they depend on.
Prototype-based methods are typically
\textit{parametric},\stdmarginnote{(non-)\\parametric clustering} which
means that for example the number of clusters or their constitution
needs to be pre-defined.
An optimal solution to the clustering problem is sought given the
constrains of the mandatory parameters.
The outcome of the clustering can depend non-trivially on these
parameters.
In contrast to this, \textit{non-parametric} clusterings do not depend
on parameters that pre-define the number of obtained clusters or their
constitution.
They can still use parameters but these are seen as tuning parameters
to set the granularity of the process.
The outcome of the clustering usually varies systematically with these
parameters.
Parametric clustering can be related to flat clustering that produces a
single result while non-parametric clustering can be related to
hierarchical clustering in the sense that a screening of the granularity
parameter produces a hierarchy of clusterings.\cite{Henning2016}
%


We would like to close this section with mentioning a very different, yet
tempting approach to categorise clustering methods, that is to compare them
in terms of the output they produce on example data sets.\cite{Jain2004}
Using an index of external validation, it can be evaluated how well the
cluster label assignments obtained through two different methods agree,
without assuming that either one of them represents the ground
truth.\stdmarginnote{clustering clusterings}
By comparing a set of clusterings in a pairwise fashion, one does
essentially build up a similarity matrix since the used index of
external validation can be interpreted as a similarity measure between
clustering results.
Based on this matrix it is possible to cluster clusterings and to find
groups of seemingly related algorithms.
This perspective offers a purely assignment centred approach beyond
categorisations based on how clustering methods assign labels, the
structures they produce, how they operate, or what the cluster model is.
Any categorisation can only provide an idea about how a specific method
may be distinguished from others but the myriad of available clusterings
may after all be judged solely by whether they find the result that fits
the application best.



\chapterwsub{Clustering methods}{Many tools for the same purpose}
\label{chap:clustering_methods}

\lettrine{T}{}he pool of available clustering methodologies to choose
from is very large.
As outlined in section~\ref{sec:similarity_definitions}, they may differ
in the kind of cluster model they try to adapt to the clustered data,
the presence or absence of an explicit objective (that can be optimised
in a mathematical sense), and in the algorithms that
will eventually make them usable.
Technically, a lot of clustering algorithms are actually a combination
of more than one fundamental algorithm, which can lead to a possibly even
larger number of detailed algorithmic variations and implementations.
This can make things difficult if we want to compare clustering methods
from a bird's eye perspective since what can be ultimately compared are
always only individual algorithms and specific implementations thereof.
While in principle different realisations of the same method
should yield the same if not very similar results, this may not be true
in reality in any case, for example when they employ
constrains or make decisions in inconclusive situations.
On top of contingent qualitative differences, individual implementations of
basically the same algorithm can vary drastically in their efficiency.\cite{Kriegel2017}
The name used to refer to a clustering can either describe the basic
idea that the method is based on, or rather a specific algorithm, or even
just a concrete implementation.
In the following, we will discuss a few commonly used clustering methods with
corresponding algorithms and (rough) example implementations.
The selection is influenced by whether knowledge about them is
potentially useful to fully understand \gls{commonnn} clustering and to
put it into context.
For each presented method, we will try to point out what is the cluster
model, what could serve as an inductive principle, what are typical
variations, and what are suitable cases of applications, that is for
which data sets is a method in general suitable.





\section{Linkage clustering}
\label{sec:linkage_clustering}

\firstsecpar{Let's start our short survey of clustering procedures} with
one of the ancestral methods that has its origins back in the early
1950s.\cite{Graham1985}
\textit{Single-linkage} clustering is an archetype of connectivity-based
clustering, which makes use of pairwise object similarities or
dissimilarities to find groups of objects.
The method aged rather well and is still widely applied.
As one of the standard clustering techniques, it has been used for the
development of a multitude of other clustering methods or as a building
block in more complex procedures.
In a nutshell, single-linkage clustering is based on the following idea:
the two most similar objects in a given data set should be grouped into the
same cluster.
The argument can be continued, so that the next two most similar objects
in the data set should also be grouped into the same cluster, followed by
the next most similar pair of objects, and so forth.\stdmarginnote{principle}
This reasoning is intrinsically hierarchical.
It leads to the somewhat
unfortunate use of \textit{hierarchical clustering} as a synonym for
single-linkage clustering and closely related methods.
In a bottom-up fashion, objects are iteratively merged into larger and
larger clusters at decreasing granularity of similarity until eventually
all objects end up in the same cluster.
At each step, merged objects are less and less similar.
An alternative term to describe this bottom-up approach of this method is
\textit{agglomerative} clustering.
Figure~\ref{fig:single_linkage_nutshell} illustrates this process with
a simple example.

\begin{illustration}
        \figureindent\includegraphics{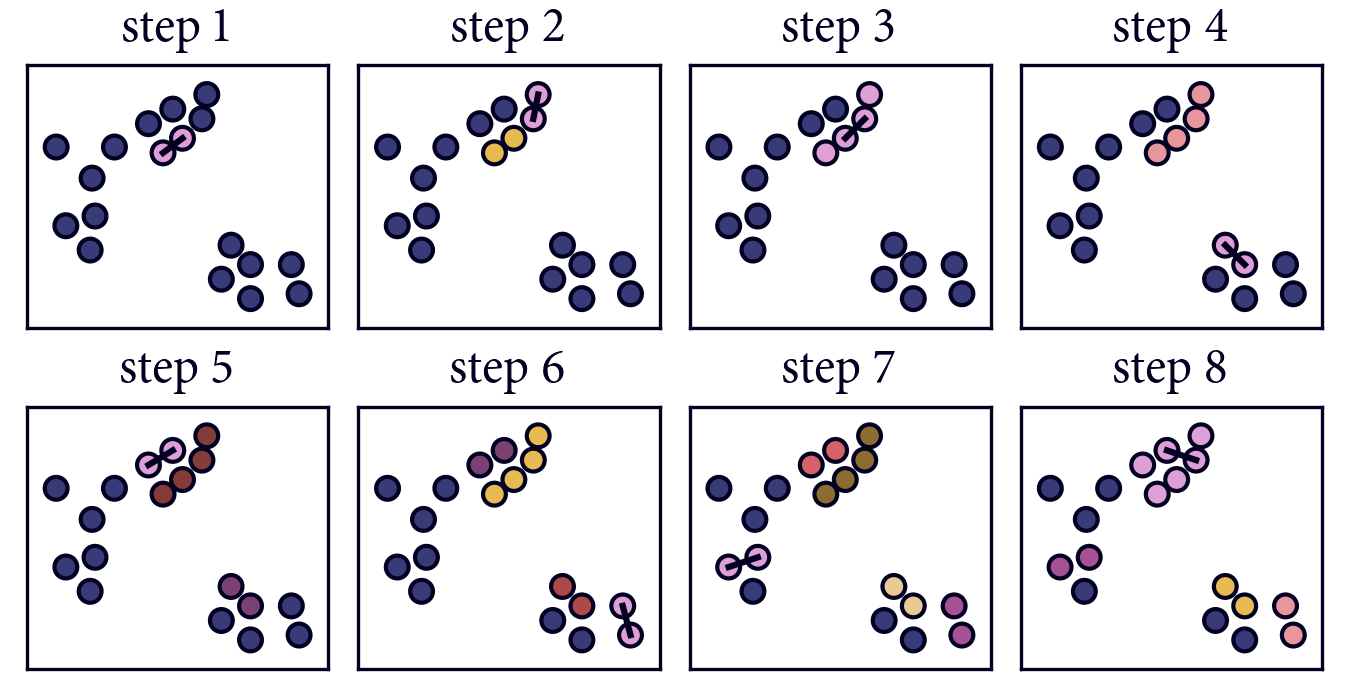}
        \captionofbottom{figure}{%
            Single-linkage clustering in a nutshell}{%
            Single-linkage clustering in a nutshell}{%
            Data points are merged iteratively into clusters \textit{bottom-up},
            two data points at a time. In step 1, the two closest (most similar)
            data points are connected and form the first cluster. In step 2, the
            next two closest points are connected and form a second cluster.
            In step 3, the two next closest points are already part of separate
            clusters in which case the two sub-clusters are merged into a bigger
            super-cluster. The process is continued until all points are part of
            the same cluster or until a stoppage criterion is reached. A user
            might be interested in either the hierarchy of clusters or clusters
            at certain stages.}
    \label{fig:single_linkage_nutshell}
\end{illustration}

A hierarchy of clusterings with atomic objects at the bottom and a
single cluster containing all objects at the top is called a
\textit{full} hierarchy.
The connectivity-based model that will be produced by single-linkage
clustering is essentially a rooted tree of clusterings.
An individual clustering result may be selected from the hierarchy for
example by adding the constraint that a specific number of clusters
should be obtained.\stdmarginnote{stoppage criterion}
Such an additional demand on the cluster result can be seen as a \textit{stoppage}
or \textit{threshold criterion}.
Instead of the (full) hierarchy, the preferred model for the data
will consequently be a flat clustering.
Figure~\ref{fig:iris_data_points_agglomerative_single_3} shows an application
of single-linkage clustering to the \textit{Iris} data set.
Note, that the used similarity measure is just the Euclidean distance here.
A specification of three target clusters does, however, not yield a result
that is in good agreement with the expectation.
We can observe that the orange cluster has swallowed up the green cluster
almost entirely.
Such behaviour is not untypical for single-linkage clustering that has
therefore a bad reputation for being sensitive to chaining effects, i.e.
to spurious data points that prevent larger clusters from separating.

\begin{illustration}
    \centering
    \begin{minipage}[t]{0.75\textwidth}
        \vspace{0pt}
        \includegraphics{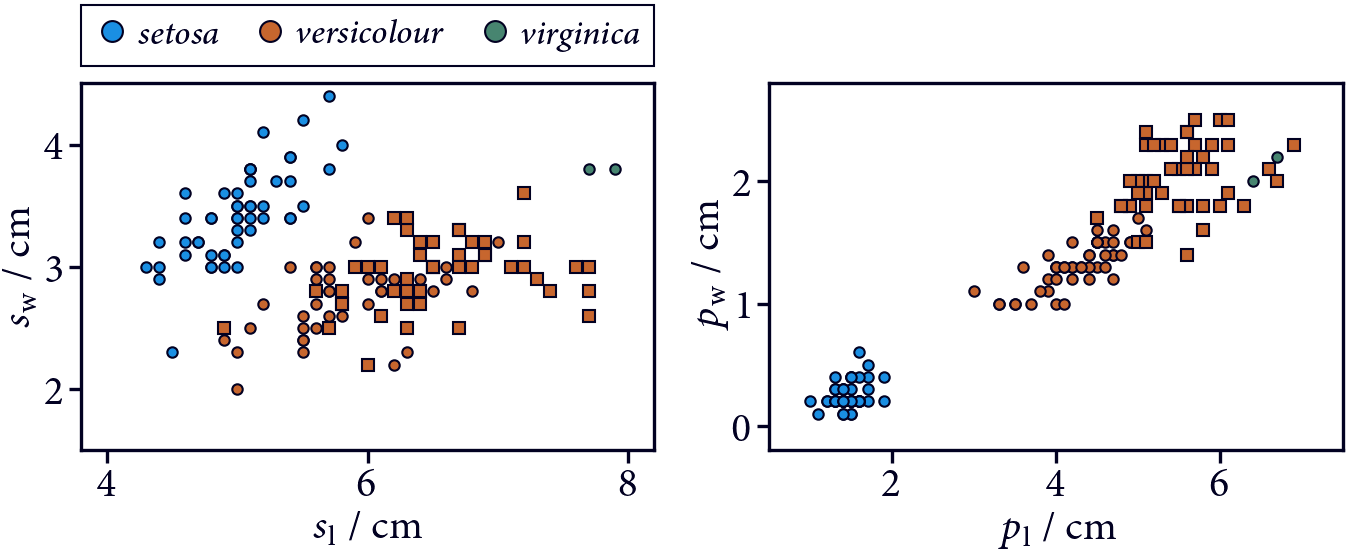}
    \end{minipage}
    \begin{minipage}[t]{0.75\textwidth}
        \vspace{0pt}
        \captionofbottom{figure}{%
            \textit{Iris} data set single-linkage (3 clusters)}{%
            \textit{Iris} data set single-linkage (3 clusters)}{%
                Data points (compare
                figure~\ref{fig:iris_data_points_true_labels}) with cluster
                labels found by agglomerative clustering
                (\cminline{sklearn.cluster.AgglomerativeClustering}
                using single-linkage). \SI{68}{\percent} of the cluster
                labels match the true classification labels. Non-matching
                assignments are marked with squares.}
            \label{fig:iris_data_points_agglomerative_single_3}
    \end{minipage}
\end{illustration}

Alternatively, the hierarchical tree of clustering results can be
investigated to select a specific clustering at an appropriate distance
(similarity) threshold.
A typical way to plot single-linkage hierarchies to this effect is to use
a dendrogram as shown in figure~\ref{fig:iris_dendrogram_agglomerative_single}.
In such a plot, the similarity measure is put on one axis while the indices
of individual data points are shown on the other one.\stdmarginnote{dendrogram}
A merge of points and respectively clusters into a bigger cluster is represented
by two legs of a tree that meet at a specific similarity level.
From this, we can identify a similarity level at which (among others)
three clusters are obtained that agree better with the expectation (see
figure~\ref{fig:iris_data_points_agglomerative_single}).
The cluster hierarchy can give deep insight into the structure of a data
set.
Of course it is also possible to select a combination of clusters as
branches of the tree with differing similarity thresholds.
It may, however, become increasingly difficult to fully grasp these
hierarchies when larger data sets are considered.

\begin{illustration}
    \includegraphics{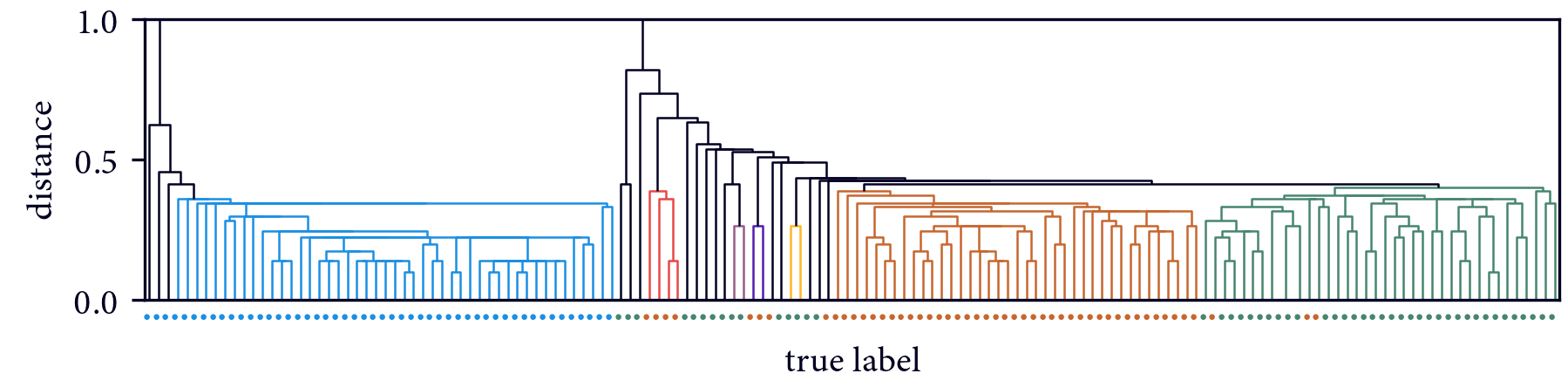}
    \captionofbottom{figure}{%
        \textit{Iris} data single-linkage dendrogram}{%
        \textit{Iris} data single-linkage dendrogram}{%
        Cluster hierarchy (compare
        figure~\ref{fig:iris_data_points_agglomerative_single_3}).
        At the lower end of the dendrogram are individual data
        points where a coloured dot indicates the true class label.
        The legs of the tree show which points are merged into
        clusters with a respective distance (similarity) at which
        the merge occurs. Selection of a distance threshold
        \(d=0.41\) results in seven clusters (coloured legs) among
        which the three largest clusters match the true
        classification rather well.}
    \label{fig:iris_dendrogram_agglomerative_single}
\end{illustration}

\begin{illustration}
    \centering
    \begin{minipage}[t]{0.75\textwidth}
        \vspace{0pt}
        \includegraphics{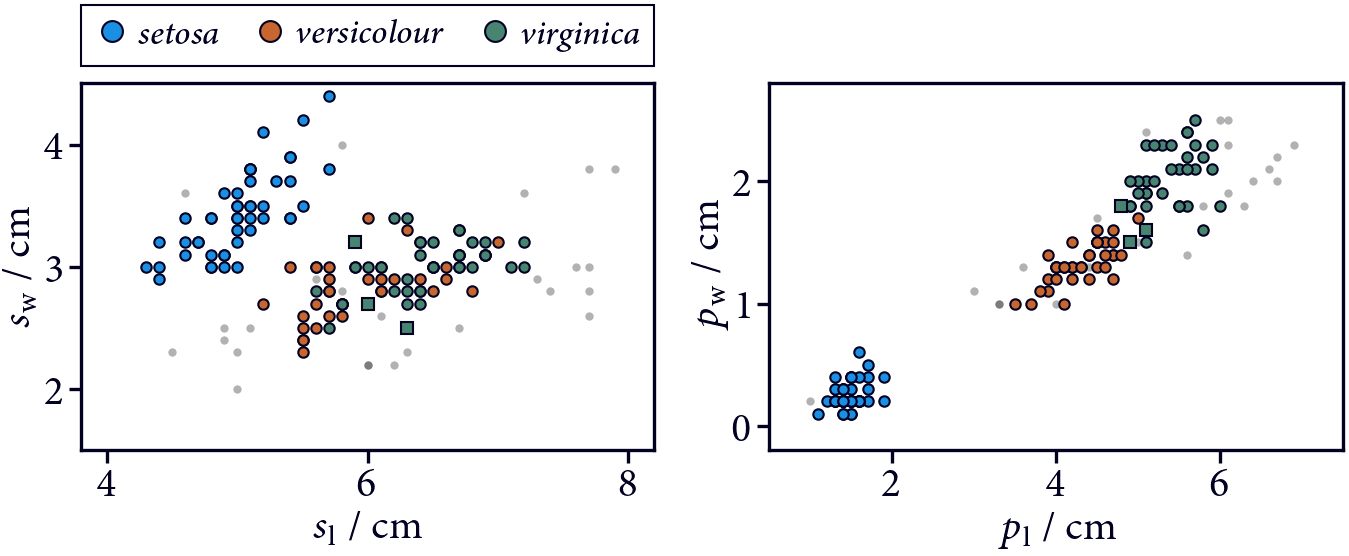}
    \end{minipage}
    \begin{minipage}[t]{0.75\textwidth}
        \vspace{0pt}
            \captionofbottom{figure}{%
            \textit{Iris} data set single-linkage (distance threshold)}{%
            \textit{Iris} data set single-linkage (distance threshold)}{%
                Data points (compare
                figure~\ref{fig:iris_data_points_true_labels}) with cluster
                labels found by single-linkage clustering after choosing a
                distance threshold from the cluster hierarchy
                (compare~\ref{fig:iris_data_points_agglomerative_single_3}).
                \SI{81}{\percent} of the cluster labels match the true
                classification labels (\SI{98}{\percent} if smaller clusters
                are neglected as noise). Non-matching assignments are marked with
                squares.}
        \label{fig:iris_data_points_agglomerative_single}
    \end{minipage}
\end{illustration}

The way how such a dendrogram can be communicated deserves special
attention and a good example for that is given by SciPy with the
\tminline{scipy.cluster.hierarchy}.
Let's say we have a set of \(n\) data points.\stdmarginnote{SciPy hierarchy}
In each single-linkage clustering iteration, two clusters \(a\) and
\(b\) are joined to form a new parent cluster \(c\).
Clusters are labelled with successive integers (starting with 0), where
the first \(n - 1\) clusters are the singletons represented by the \(n\)
data points in the set.
There will be \(i = n - 1\) iterations, giving rise to the new clusters
\(\{n, n + 1, ..., 2n - 2\}\) that will be collected in rows
of a \(i \times 4\) hierarchy matrix \(Z\).
The elements \(Z_{i, 0}\) and \(Z_{i, 1}\) contain the labels of the
clusters that are merged to give the \(c = (n + i)\mathrm{th}\) cluster.
\(Z_{i, 2}\) holds the distance value of the respective merge and
\(Z_{i, 3}\) holds the total number of data points in the new cluster.
Hierarchies in this format can be subjected to a lot of Scipy's
hierarchy related functionalities (including the plotting of
dendrograms) and it may thus be beneficial to adhere to this format in
general.
As a plus, storage of such a matrix is relatively efficient.
It may be, however, a disadvantage in certain situations that at each
iteration only the information on which clusters are merged is kept
while the information on which points were responsible for the merge is
lost.

\begin{illustration}
    \hspace{0.03\textwidth}%
    \begin{minipage}[t]{0.33\textwidth}
    \vspace{0pt}
    \begin{tabular}{rrrrrr}
        \toprule
        i & \(c\) & \(a\) & \(b\) & \(d_{ab}\) & size \\
        \midrule
        0 & 5 & 0 & 1 & \(d_0\) & 2 \\
        1 & 6 & 2 & 3 & \(d_1\) & 2 \\
        2 & 7 & 4 & 5 & \(d_2\) & 3 \\
        3 & 8 & 6 & 7 & \(d_3\) & 5 \\
        \bottomrule
    \end{tabular}
        \vspace{0.25\baselineskip}
    \end{minipage}\hspace{0.01\textwidth}%
    \begin{minipage}[t]{0.61\textwidth}
        \vspace{-0.275pt}
        \captionofside{table}{%
        SciPy single-linkage hierarchy format}{%
        SciPy single-linkage hierarchy format}{%
            Example \(Z\) matrix for a set of 5 data points.
            In iteration \(i\), cluster \(c\) is formed by merging
            \(a\) and \(b\) at distance \(d_{ab}\).
        }
         \label{tab:scipy_hierarchy}
    \end{minipage}
\end{illustration}

\custompar{Since single-linkage clustering does in essence} only make use of
pairwise similarities and not of individual object properties,
single-linkage can be understood as being detached from the (metric)
data space that objects may be embedded in.\stdmarginnote{latent space}
In this sense, the data space is treated as a latent space.
Object coordinates with respect to the data space may well be necessary
to compute similarities in the first place but as far as the clustering
is concerned similarities could have a discretionary source.
In principle, single-linkage clustering can be adopted for arbitrary
similarity concepts.
Besides single-linkage, which merges clusters at each step by the
minimum distance between existing clusters, there are also other types
of linkages.
In \textit{complete-linkage} for example, the maximum distance between
existing clusters is taken to decide which clusters are merged next.\stdmarginnote{other linkages}
\textit{Average-linkage} uses the mean distance between clusters.
A linkage uses a similarity measure and provides a prescription of how
to extend the similarity estimate to non-singleton clusters.
In a way, a linkage criterion is also a kind of (greedy) clustering
objective, suggesting an optimal cluster merge at each hierarchy step.
Depending on the linkage, the choice of meaningful distance functions
may be limited.
Traditionally, linkage clustering is thought of and implemented
based around a distance or similarity matrix.
At each step, the minimum value in this matrix decides over which points
are merged.\stdmarginnote{implementation}
For the next step the matrix has to be reduced by the just
used elements and respectively modified with the distances to the newly
formed cluster.
A generic way to realise most common linkage schemes at this point is
given by the Lance-Williams dissimilarity update
formula.\cite{Lance1967,Lance1967a}
A procedure like this has a rather costly runtime complexity of
\(\mathcal{O}(n^3)\) with respect to the number of points in the data
set \(n\).
If a distance matrix is explicitly stored during the clustering, it
also has a memory demand on the order of \(\mathcal{O}(n^2)\).
It should be noted, though, that in some cases leveraging priority
queues to store and retrieve similarities in the needed order is
preferable over the pure matrix centred approach.\cite{Krznaric2002}

For single-linkage clustering specifically, there exist also other neat
solutions.
SLINK clustering deserves to be mentioned here for example.\cite{Sibson1973}
In the context of \gls{commonnn} clustering, which will be discussed
later, however, one fact is of particular
interest:\stdmarginnote{single-linkage and \glsfmtlong{mst}}
all the information that is required for the single-linkage clustering of
a data set is contained in a \glsfirst{mst} for the data.\cite{Gower1969}
This basically means that the clustering problem can be substituted by
the well studied problem of building \glspl{mst}.
Through a revision of the edges in a \gls{mst} in order of there
similarity weight, a hierarchy can be constructed that is equivalent
to a single-linkage clustering result.
As a last remark, it is in principle also conceivable to use
\textit{divisive} top-down strategies instead of the described
agglomerative approaches to split an initially un-clustered data
set into smaller and smaller sub-clusters.\stdmarginnote{divisive clustering}
Because there exists a combinatorial number of \(2^{n-1} - 1\) possible
solutions for each cluster split, where \(n\) is the number of objects in
the cluster, divisive methods are, however, computationally very
difficult to operate and are usually heuristically heavy as for example
in DIANA clustering.\cite{Kaufman2009}
Recall that the number of possible merges in agglomerative
clustering only amounts to \(n(n-1)/2\), where \(n\) is the current
number of clusters available for a merge.
Divisive strategies play, however, an important role in the context of
the interesting question of how to cut graphs in an optimal way under
various constraints.
Consider for example the \textit{minimum cut} problem of splitting
a graph into components by removing a minimum number of edges or
edges with a minimal total weight.\cite{Stoer1997}\stdmarginnote{graph cuts}
A variation of this would be to find a \textit{normalised cut} that is a
split into components where the cost of the split amounts to the total
weight of the removed edges normalised by the total edge weight of the
resulting components,\cite{Shi2000} which favours cuts into \enquote{balanced}
components.
Spectral clustering for example, which will be addressed in the
following section, can be considered an approximate solution to the
relaxed normalised cut problem.\cite{VonLuxburg2007}
Another common example for graph cut problems is \textit{sparsest
cut} where the total weight of removed edges is normalised by the number
of nodes in the smallest of the resulting components, which favours
components of equal size.\cite{Arora2009}

\section{Spectral clustering}
\label{sec:spectral}

\firstsecpar{Pairwise data object similarities are the fundament} of
another broad family of clustering procedures referred to as
\textit{spectral} methods.
As mentioned in section~\ref{sec:similarity_definitions}, a spectral
embedding of a data set does not actually represent a clustering itself
in the classic sense but rather tries to provide a convenient
transformation of the data space into a lower dimensional representation
in which a clustering can be done easier or the cluster structure
becomes more obvious.
In general, a spectral clustering requires a similarity matrix \(S\)
(often also called affinity matrix in this context) as input.
From that, one constructs the graph Laplacian \(L\)
as\cite{VonLuxburg2007}\stdmarginnote{graph Laplacian}
\begin{equation}
    L = D - S
    \label{eq:graph_laplacian}\,,
\end{equation}
where \(D\) stands in for the diagonal degree matrix \(D_{ii} = \sum_j S_{ij}\).
Note that the elements of the similarity matrix correspond to edge
weights \(w(i, j)\) between a pair of data objects \(i\) and \(j\) when
the data is thought of in a graph representation (recall that the degree
of a graph node is the total weight of edges connecting the node to
other nodes).
\(S\) is generally required to describe an undirected graph, i.e. it has
to be symmetric which does also make \(L\) a symmetric matrix.
Furthermore, the used similarities need to be well-behaved, which means
they are not allowed to be negative and need to be actual similarities,
not distances for instances.
A typically used similarity is for example found in symmetric
\(k\)-nearest graphs in which a pair of data points is
considered\stdmarginnote{similarity examples} similar (\(w = 1\)) if one
of the points is a \(k\)-nearest neighbour of the other or if both
points are mutually among their \(k\)-nearest neighbours.
Another example for a continuous similarity are Gaussian neighbourhoods
\(w(i, j) = \exp(-||x - y||^2 / (2 \sigma^2))\) for a given
\(\sigma\)-value, here with \(x\) and \(y\) as the coordinates of data
point \(i\) and \(j\).
We are then interested in the (smallest) eigenvalues and corresponding
eigenvectors of \(L\).
In practice there are a few possible variations on how to find those and
on how to normalise \(L\) beforehand,\cite{VonLuxburg2007,Weber2004} but
let's ignore this here for the sake of simplicity.
\(L\) has \(n\) (the number of graph nodes) non-negative eigenvalues
\(\lambda_1 \leq \lambda_2 \leq, ..., \leq \lambda_n\), with \(\lambda_1
= 0\) of which the corresponding eigenvector is a constant one-vector.
Laxly spoken, the neat thing about this is that the eigenvalues can be
interpreted as a cost of separating the input graph into clusters.\stdmarginnote{evaluation}
The \(k\)th eigenvalue gives us a separation of the graph nodes into
\(k\) disjoint sets, that is if we are interested in an isolation of two
clusters, we have to consider the 2nd eigenvalue \(\lambda_2\).
Zero-valued eigenvalues (no cost) are an indicator for the number of
connected components (initially disjoint sub-graphs) in the input graph,
i.e. if there are two zero-valued eigenvalue \(\lambda_1 = \lambda_2 =
0\) the graph contains two connected components that are not connected
to each other.
From the associated eigenvectors, we can get the cluster memberships
of individual graph nodes by evaluating their projection onto the first
\(k\) chosen eigenvectors.
This is typically done for example by a \(k\)-means clustering (see
section~\ref{sec:kmeans}) in the resulting low dimensional projection.
Figure~\ref{fig:spectral_embedding} shows an example for a spectral
embedding of the \textit{Iris} data set.
In figure~\extref{fig:spectral_embedding}{a} we can see the eigenvalue spectrum
for a Laplacian constructed from \(k\)-nearest neighbour similarities.
Since we are interested in three clusters to be isolated, we focus on the first
three smallest eigenvalues.
While figure~\extref{fig:spectral_embedding}{b} shows the input
similarity matrix (that exhibits a block diagonal form because the input
point samples are ordered by their true classification label),
figure~\extref{fig:spectral_embedding}{c} illustrates the 2-dimensional
projection of the data points onto the 2nd and 3rd eigenvector.
In this low dimensional projection, the group structure of the data
becomes well identifiable.

A particular form of spectral clustering, which plays a considerable
role before the background of \gls{md} and the estimation of kinetic
Markov models to identify clusters of molecular metastable conformations,
is found in \glsfirst{pcca}.\cite{Weber2004,Deuflhard2005}
This method solves an eigenvalue problem for a stochastic matrix \(T\),
namely for example a transition probability matrix modelling a molecular
trajectory as a Markov chain.\stdmarginnote{\glsfmtshort{pcca}}
The eigenvalues of this matrix start with \(\lambda_1 = 1\) (the Perron
eigenvalue) and one is typically interested in the next smaller
eigenvalues for the identification of kinetic clusters.
The eigenvectors of \(T\) are the same as those of a corresponding graph
Laplacian \(L\) and so is the way how one uses a low dimensional
projection onto these eigenvectors for the separation of weakly
connected sub-blocks within \(T\).
The clustered objects are in this case the Markov states on top of which \(T\)
was constructed and the obtained clusters can be interpreted as metastable
conformational states between which transitions are slow (i.e. rare).
\gls{pcca} as a method does as such, however, not refer to the spectral
embedding part but rather to the procedure of how to analyse the
eigenvectors as an alternative to the wide use of \(k\)-means.
In its basic form, it exploits the sign structure of individual eigenvectors
to assign states to one of two groups while iteratively considering higher
eigenvectors.
The more recent robust implementation does rely on the property of the
eigenvectors that these ideally form a \(k\)-dimensional simplex if one
considers \(k\) eigenvectors at the same time.\cite{Roblitz2013}

\begin{illustration}
    \includegraphics[]{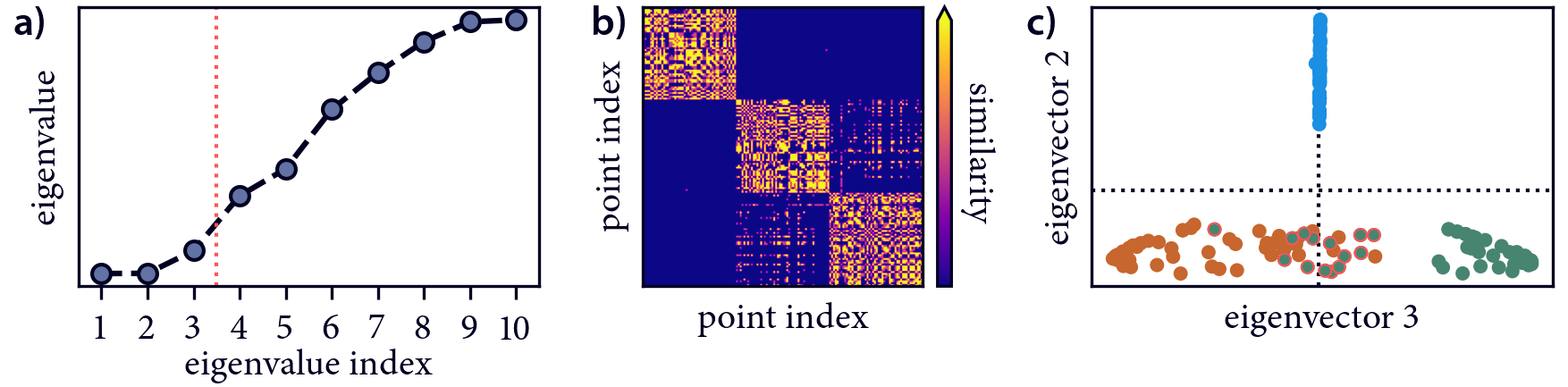}
    \captionofbottom{figure}{%
        Spectral embedding of the \textit{Iris} data set}{%
        Spectral embedding of the \textit{Iris} data set}{%
            \subl{a} The first ten eigenvalues in the sorted eigenvalue
            spectrum of the graph Laplacian. We want a partitioning of
            the data  into 3 components based on the first three
            eigenvalues. \subl{b} Affinity (similarity) matrix in
            block-diagonal form with matrix elements coloured by value.
            \subl{c} Embedding of the data into the reduced space
            defined by the 2nd and 3rd eigenvector of the Laplacian.
            Points are coloured by their true classification label. The
            identification of clusters can be done by separating the
            points along each eigenvector (e.g. below and above 0
            indicated by the dashed lines) or by using another
            clustering approach, e.g. \(k\)-means (see
            section~\ref{sec:kmeans}). Points highlighted with a red
            outline will probably be assigned to the wrong cluster.}
    \label{fig:spectral_embedding}
\end{illustration}

\begin{illustration}
    \centering
    \begin{minipage}[t]{0.75\textwidth}
        \includegraphics{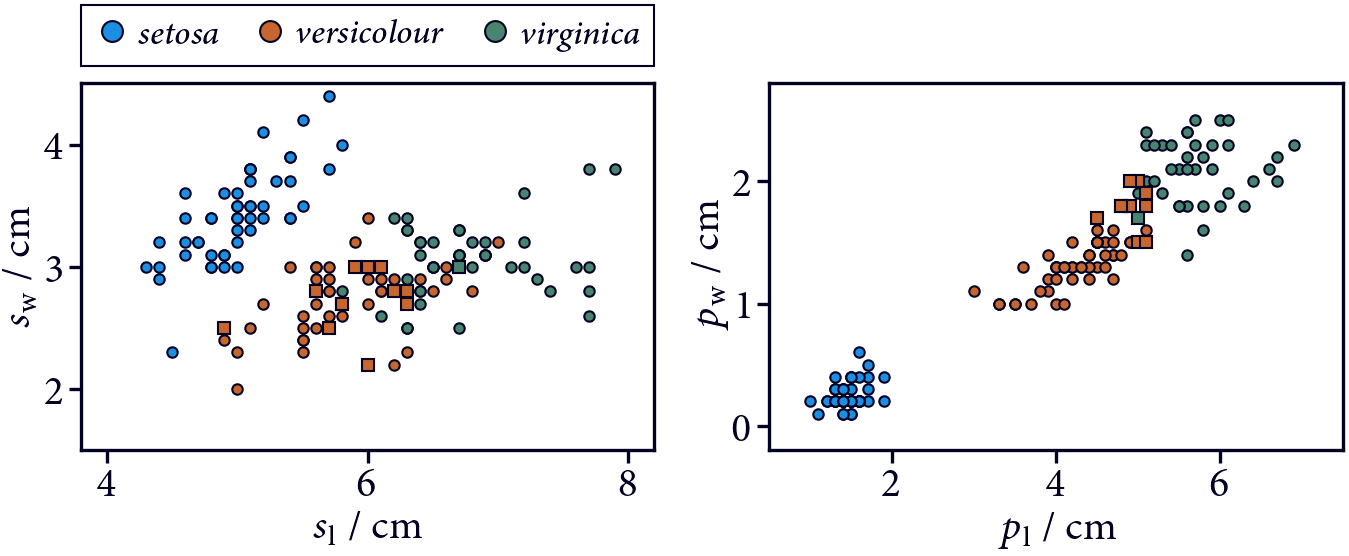}
    \end{minipage}
    \begin{minipage}[t]{0.75\textwidth}
    \captionofbottom{figure}{%
        \textit{Iris} data set spectral clustering (3 clusters)}{%
        \textit{Iris} data set spectral clustering (3 clusters)}{%
        Data points (compare
        figure~\ref{fig:iris_data_points_true_labels}) with cluster
        labels found by spectral clustering
        (\cminline{sklearn.cluster.SpectralClustering}
        using symmetric \textit{k}-nearest neighbour similarities with
        \(k=26\)). \SI{91}{\percent} of the cluster labels match the
        true classification labels. Non-matching assignments are marked
        with squares.}
        \label{fig:iris_data_spectral}
    \end{minipage}
\end{illustration}

\section{\textit{k}-Means clustering}
\label{sec:kmeans}

\firstsecpar{The \textit{k}-means method} is arguably the most popular of all
clustering procedures.\cite{Henning2016}
Being not much younger than single-linkage clustering, \textit{k}-means
has quite a long history and has been studied extensively from various
perspectives.\cite{Bock2008}
Its popularity may be not least due to a solid mathematical framework.
The fundamental idea behind this method, is that a set of objects \(\mathcal{D} =
\{x_1, ..., x_n\}\) can be partitioned into a fixed number of \(k\)
subsets \(\mathcal{C} = \{C_1, ..., C_k\}\), so that \(\bigcup_{i=1}^k C_i =
\mathcal{D}\) and \(C_i \cap C_j = \emptyset~\forall~i \neq j\), and
where each \(C_i\) is characterised by a representative \(\mu_i\), so
that \(\mathcal{D}\) can be modelled by the set of representatives
\(\mathcal{M} = \{\mu_1, ..., \mu_k\}\).
The inductive principle underlying the \textit{k}-means method is a
particularisation of equation~\ref{eq:partitioning_objective_function}
\begin{equation}
    \substack{\mathrm{min}\\\mathcal{C}} \left\{\sum_{i=1}^k \sum_{x \in C_i} d(x - \mu_i)\right\} = \substack{\mathrm{min}\\\mathcal{C}} \left\{\sum_{i=1}^k \frac{1}{2|C_i|}\sum_{x,y \in C_i} d(x - y)\right\}
    \label{eq:kmeans_objective_function}
\end{equation}
with \(d\) being the squared Euclidean distance in \(m\) dimensions
\begin{equation}
    d(a, b) = d_{\mathrm{Euclidean}}^2 = \sum_{i=1}^m (a_i - b_i)^2\,.
    \label{eq:squared_Euclidean}
\end{equation}
Algorithms implementing \textit{k}-means strive to optimise the partition
\(\mathcal{C}\) according to equation~\ref{eq:kmeans_objective_function}, which
states literally: \enquote{\textit{pick the model (the set of \(k\)
representatives) that minimises the total squared
error}},\cite{Estivill-Castro2002} where \textit{error} needs to be understood
as the deviation of objects from the representative for the cluster they where
assigned to.
It is important to emphasise, that the squared Euclidean distance is used
here, which requires that the clustered data points \(x \in \mathcal{D}\) are
given as feature vectors in \(\mathbb{R}^m\).
The representatives that minimise in this case the sum of squared distances
for a given partition \(\mathcal{C}\) are the arithmetic mean points
of each cluster \(\mu_i = 1 / |C_i|\sum_{x \in C_i} x\).
These so called \textit{centroids} constitute a special form of
prototype-based model.
A minimisation of the sum of squared distances between points in a
cluster and the mean is related to a minimisation of within-cluster
variance via \(|C_i|\sigma_i^2 = \sum_{x \in C_i}
d_{\mathrm{Euclidean}}^2(x - \mu_i)\).
A minimisation of intra-cluster variance implies a maximisation of
inter-cluster variance according to the law of constant
variances.\cite{Kriegel2017}
In turn, it is also equivalent to a
minimisation of the squared pairwise distances between points of the
same cluster as stated by the right side of
equation~\ref{eq:kmeans_objective_function}.
Although we are reluctant to say that \textit{k}-means is
similarity-based, it is true that each point in the data set will be
assigned to the cluster with the closest (that is most similar) centroid
and that points in the same cluster can be characterised by a shared
similarity to the same centroid.
Also, a minimisation of squared pairwise distances between points of the same
cluster can be understood as a maximisation of within-cluster similarity on
average.
Note, however, that two individual points in the same cluster can still
be less similar to each other than to points in a different cluster.
In particular, for any given point the most similar points may not always be
found within the same cluster.
As pointed out in section~{\ref{sec:similarity_definitions}}, the
paradigm of the method may be more aptly described as a maximisation of
within-cluster homogeneity.
The dependence of \textit{k}-means on the squared Euclidean distance
(\(L_2\) norm) is non-negotiable.
To exchange the distance metric that evaluates the deviation of objects
and their cluster representatives means to change the inductive
principle underlying the method
(equation~\ref{eq:partitioning_objective_function}) and implies a different
kind of prototype-based model for the data.
The mean minimises the sum of squared deviations within clusters.
Using for example the Manhatten distance (\(L_1\) norm) instead, results
in an objective for the identified clusters that is optimised by their
medians---not their means.
The median minimises the sum of absolute deviations within clusters.
This is incorporated in other clustering methods like \textit{k}-medians
or in a generalised form for arbitrary metrics in PAM
(\textit{k}-medoids), where the prototypes are constrained to be among
the clustered objects themselves so that \(\mu_i \in
\mathcal{D}~\forall~i\).\cite{Schubert2021}
It is, however, possible to transform the clustered data in a way so
that the Euclidean distance after the transformation corresponds to a
different distance before the transformation, which allows the
application of \textit{k}-means for example in the context of cosine
similarity, covariance (Mahalanobis distance), and correlation.
Equation~\ref{eq:kmeans_objective_function} can be computationally quite hard
to satisfy exactly.\cite{Mahajan2012}.
A classic algorithm that provides an iterative, locally optimal solution
to the problem was presented by Lloyd,\cite{Lloyd1982} but there are
many other proposed heuristic algorithms.\cite{Tarsitano2003}
It is based on the rational of two alternately applied optimisation steps:
first, given a set of current centroids \(\mathcal{M}\), the currently best
partition of the data \(\mathcal{C}\) can be constructed by assigning each data
point to the closest centroid so that \(C_i = \{x \in D\,|\,d(x, \mu_i)
\leq d(x, \mu_j)~\forall~j \neq i\}\).
Second, given a current partition \(\mathcal{C}\), the currently best centroids
can be computed as the mean of the data points in each cluster.
Starting with an initial set of centroids, the two steps can be repeated until
for example the positions of the centroids or the within-cluster variances or
the cluster assignments are converged, or if a maximum number of iterations is
reached.
It is generally considered an asset of the \textit{k}-means method that this
standard algorithm is easy to implement and computationally cheap.
A single iteration requires only one distance calculation for each pair of
\(n\) data points and \(k\) cluster centres while a single distance calculation
requires a summation over contributions in \(m\) dimensions.
The overall runtime complexity can therefore be given as \(\mathcal{O}(nkmi)\)
with the total number of iterations \(i\).
If \(k\) and \(m\) are fixed, the cost of the algorithm scales basically linear
with the number of data points, assuming that the number of needed iterations is
relatively small as well.
A problem with the Lloyd algorithm in this basic form, is that the found
solution is not necessarily globally optimal and that it depends on the
starting condition, that is on the initially chosen positions of the
cluster centres.
For randomly placed starting centroids, the final result is
non-deterministic.
Also, the number of necessary iterations to reach convergence can vary
and there is the danger of picking unlucky candidates that may
lead to empty clusters.
Figure~\ref{fig:kmeans_nutshell} demonstrates this with a simple example.
A commonly accepted workaround, is to run the algorithm multiple times
with different starting conditions and to select the best result
according to equation~\ref{eq:kmeans_objective_function}, which would be
the one in figure\extref{fig:kmeans_nutshell}{b} (the \textit{k}-means
objective is denoted by \(J_2\)).
Many \textit{k}-means implementations have this already built in and
proceed like this by default.
Alternatively, swapping procedures can be employed to escape local
minima.

\begin{illustration}
\begin{minipage}[t]{0.67\textwidth}
    \vspace{0pt}
    \includegraphics{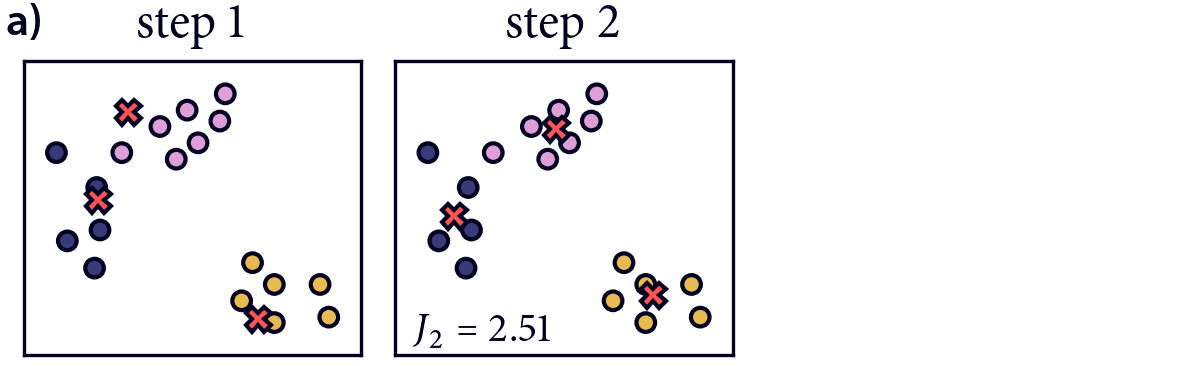}

    \vspace{2pt}
    \includegraphics{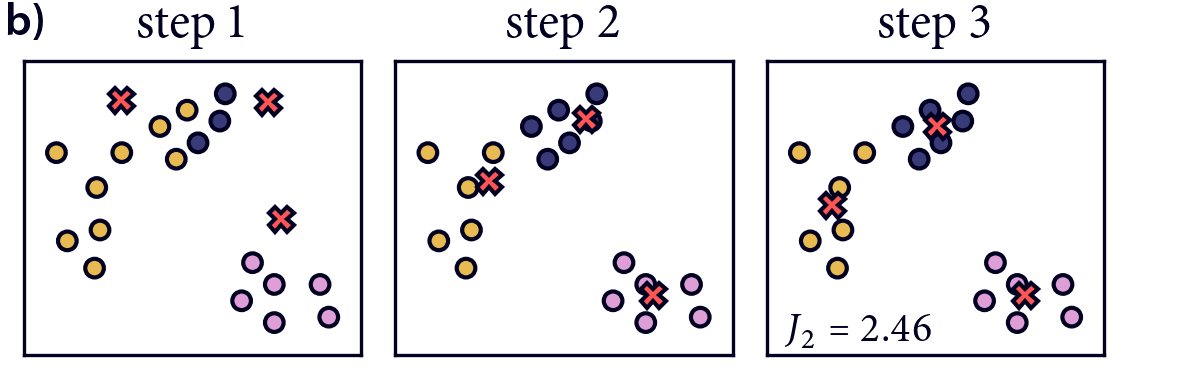}
\end{minipage}%
\begin{minipage}[t]{0.33\textwidth}
    \vspace{0pt}
    \captionofside{figure}{%
        \textit{k}-Means in a nutshell}{%
        \textit{k}-Means in a nutshell}{%
            Clustering with two different sets of \(k=3\) randomly
            placed initial centroids in \subl{a} and \subl{b}. While the
            results are stable with respect to the assignments of the
            cluster in the lower-right corner, they do not agree for the
            other two clusters. Convergence in terms of changing
            centroid positions is reached more quickly in \subl{a}.}
    \label{fig:kmeans_nutshell}
    \vspace{0.5\baselineskip}
\end{minipage}
\end{illustration}

Nonetheless, since the choice of the starting centroids is so critical, a
lot of effort went into the development of improved initialisation
schemes.\cite{Celebi2013}
Among the most successful strategies, is what is known as the
\textit{k}-means++ method.\cite{Arthur2007}
Starting with an empty set of selected starting centroids
\(\mathcal{M}\), the first centroid to be added is chosen as a random
point \(x \in \mathcal{D}\).
By defining the function \(d_{\mathrm{min}, \mathcal{M}}(x)\) as the
minimum distance between a point of the data set to already selected
centroids in \(\mathcal{M}\), the next centroids \(x^\prime\) are added
one by one with probability \(d_{\mathrm{min}, \mathcal{M}}(x^\prime)^2
/ \sum_{x \in \mathcal{D}} d_{\mathrm{min}, \mathcal{M}}(x)^2\).
This should ensure that the starting centroids are well distributed over
the whole data set to speed up convergence and decrease the chance to
pick unlucky candidates.
Another potential drawback of \textit{k}-means, is that the number of
clusters that should be identified through the clustering needs to be
specified beforehand.
If the aim of the clustering is to isolate distinct groups of data
points, we ideally need to have an idea about how many groups there are
in the first place.
When a guess is difficult to make, a typical strategy would be to try
clusterings with different values of \(k\) and to pick the best result.
But how could we judge what would be best?
Equation~\ref{eq:kmeans_objective_function} is not very suitable to compare
clusterings with different numbers of clusters as the within-cluster
variance does always decrease with larger \(k\).
It can still be used, though, by trading within-cluster variance against
the assumption that a low number of \(k\) is generally better, in the
spirit of optimising this criterion with a minimum complexity in the
obtained model.
Figure~\ref{fig:elbow} shows a plot of this objective versus \(k\)
for the \textit{Iris} data set.
One could now choose the \(k\)-value in the plot up to which the sum of
squared distances of data points to the centroid of their cluster
decreases relatively fast, and after which the decrease is substantially
diminished, which is given for \(k=3\) (although admittedly not very clearly) in
the present example.
This is called the \enquote{elbow} method, as the optimal value for \(k\) is
presumed to be found where the plotted line shows a kink in reminiscence
of a bent arm.

\begin{illustration}
\begin{minipage}[t]{0.4\textwidth}
    \vspace{0pt}
    \includegraphics{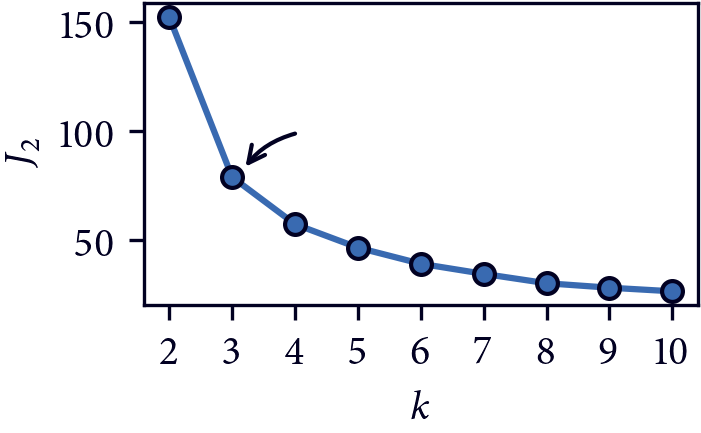}
\end{minipage}\hspace{0.02\textwidth}%
\begin{minipage}[t]{0.58\textwidth}
    \vspace{-0.25\baselineskip}
    \captionofside{figure}{%
        Choosing the \textit{k} in \textit{k}-means (elbow plot)}{%
        Choosing the \textit{k} in \textit{k}-means (elbow plot)}{%
            The plot shows the sum of squared deviations of points from
            their respective cluster center, i.e. the evaluation of the
            \textit{k}-means objective \(J_2\), versus the number of
            identified clusters. The within-cluster variance is also
            termed \textit{inertia} or \textit{distortion} in the
            literature. The optimal \textit{k} is indicated by the
            arrow.}
    \label{fig:elbow}
\end{minipage}
\end{illustration}

Alternatively, the silhouette score can be used for a similar analysis
(figure~\ref{fig:silhouette}).\cite{Rousseeuw1987}
The silhouette score for a single data point is defined as
\begin{equation}
    s_\mathrm{silhouette}(x) = \frac{\bar{d}_{\mathrm{nearest}}(x) - \bar{d}_{\mathrm{same}}(x)}{\mathrm{max}\left(\bar{d}_{\mathrm{same}}(x), \bar{d}_{\mathrm{nearest}}(x)\right)}
\end{equation}
with \(\bar{d}_\mathrm{same}\) as the mean distance between a point
\(x\) and all other points in the same cluster, and
\(\bar{d}_\mathrm{nearest}\) as the mean distance between a point \(x\)
and all other points in the next nearest cluster (by centroid distance).
The score is bounded by \(-1\) and \(1\) where values close to \(1\) are
better in the sense that within-cluster distances are low while
between-cluster distances are high.
In the present example, a preference for \(k=2\) or \(k=3\) is conveyed
by the fact that many data points have silhouette coefficients above the
average (indicated by the dotted vertical line) and the individual clusters
show a distinct elbow-like characteristic (few points with low scores and
many points with high scores).
The relative sizes of the clusters can also be taken as an indicator, although
it depends on whether equally sized clusters are actually desirable.

\enlargethispage{\baselineskip}
A third validation technique is illustrated in
figure~\ref{fig:calinksi} with the Calinski-Harabasz score, also known
as the variance ratio criterion.\cite{Calinski1974}
In contrast to the two previous indicators, this score is optimal for
a given clustering if it is maximised, and one should look for a
maximum in the plot of it versus \(k\).
The score is defined for a partitioning of \(n\) data points into \(k\) subsets as
\begin{equation}
    s_\mathrm{Calinski-Harabasz}(x) = \frac{\mathrm{trace}(\Sigma_\mathrm{inter})}{\mathrm{trace}(\Sigma_\mathrm{intra})}\frac{n - k}{k - 1}
\end{equation}
with the within-cluster covariance matrix (the sum of covariance matrices for individual clusters)
\begin{equation}
    \Sigma_\mathrm{intra} = \sum_{i=1}^k \sum_{x \in C_i} (x - \mu_i)(x - \mu_i)^\intercal
\end{equation}
and the between-cluster covariance matrix (the covariance of the cluster
centres weighted by cluster size) where \(\mu_\mathcal{D}\) is the mean of the complete data set
\begin{equation}
    \Sigma_\mathrm{inter} = \sum_{i=1}^k |C_i| (\mu_i - \mu_\mathcal{D})(\mu_i - \mu_\mathcal{D})^\intercal\,.
\end{equation}

\begin{illustration}
    \vspace{0pt}
    \includegraphics{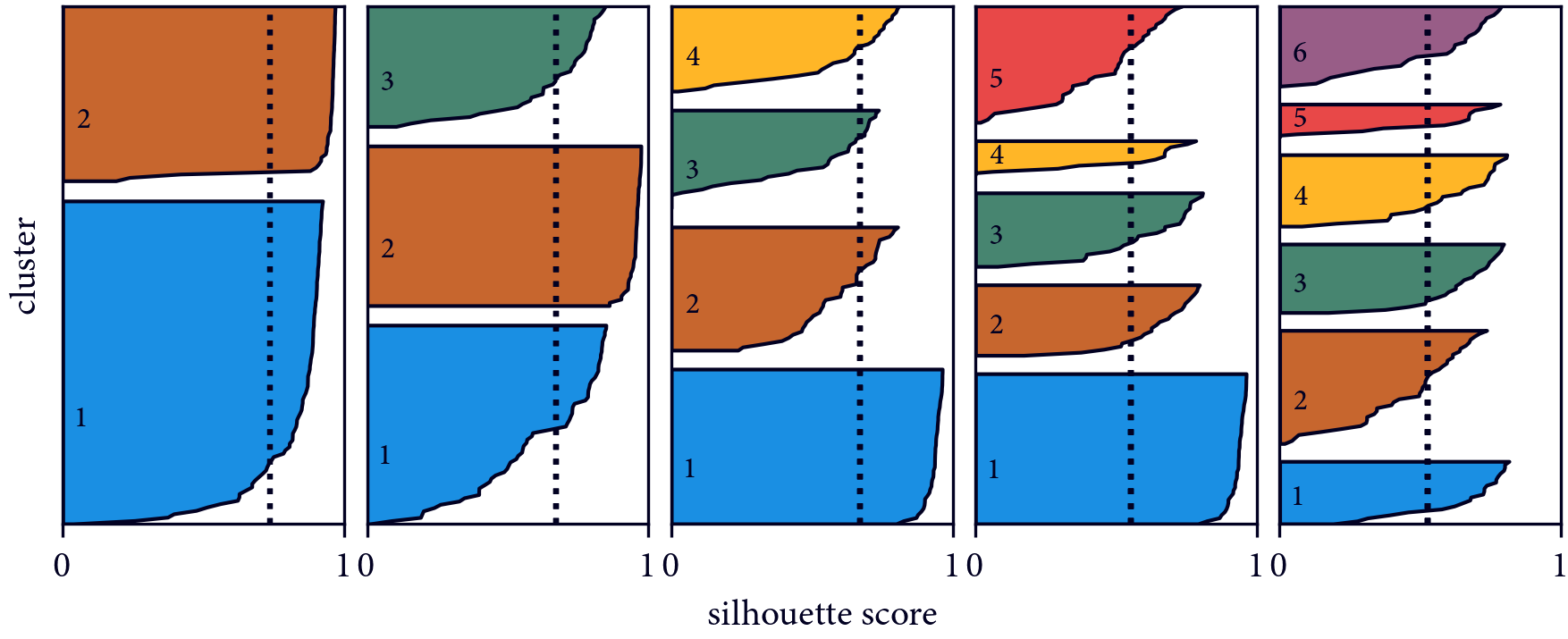}
    \captionofbottom{figure}{%
        Choosing the \textit{k} in \textit{k}-means (silhouettes)}{%
        Choosing the \textit{k} in \textit{k}-means (silhouettes)}{%
            Silhouette coefficients of the obtained clusters for \(k =
            2, 3, 4, 5,\) and 6. The profile of the scores for points in
            specific cluster is generally considered good if it shows a
            clear kink and the majority of data points has a score above
            average (dotted vertical line). The decision between \(k =
            2\) and \(k = 3\) remains somewhat ambiguous in this case.
            Relatively equal cluster sizes speak in favour of \(k = 3\)
            while the individual cluster profiles are more distinctly
            elbow-like for \(k = 2\).}
    \label{fig:silhouette}
\end{illustration}
\vspace{\baselineskip}

This score is high if within-cluster variances are low and clusters are well separated.
In the present example, the Calinski-Harabasz score strongly prefers the
clustering with \(k=3\).

\begin{illustration}
\begin{minipage}[t]{0.4\textwidth}
    \vspace{0pt}
    \includegraphics{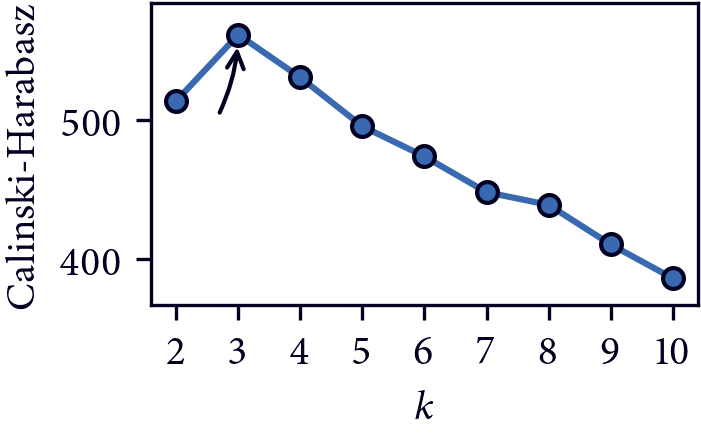}
\end{minipage}\hspace{0.02\textwidth}%
\begin{minipage}[t]{0.58\textwidth}
    \vspace{-0.25\baselineskip}
    \captionofside{figure}{%
        Choosing the \textit{k} in \textit{k}-means (Calinski-Harabasz)}{%
        Choosing the \textit{k} in \textit{k}-means (Calinski-Harabasz)}{%
        The clustering with \(k=3\) is clearly preferred because the
        score is maximised for this number of clusters.}
    \label{fig:calinksi}
\end{minipage}
\end{illustration}

The three presented evaluation techniques are all internal validation criteria, that
is they judge the obtained clusterings based solely on the grouping itself.
As mentioned in the introduction of this chapter, internal validation works well
if we have a concrete idea about how ideal clusters should be constituted.
The \(k\)-means approach has a strong opinion about this and the used validations
have strong opinions on their own that are essentially very similar.
Identified clusters are in general ideally compact, convex (globular), equally sized
and well-separated.
With this assumption in mind, we can make a judgement about which
\(k\)-means clustering agrees best with our expectation.
We would like to stress again that this validation tells us very little about the
actual quality of the clustering as the true group structure of the data set
can still differ from our expectation.
A clustering not aligned with the ideal image of clusters represented by
these validations will be ranked poorly even if it may reflect the true nature
of the data better.
For the \textit{Iris} data set, we are in the luxurious situation of having true
group assignments available, on the premise of course that the provided
expert classification into groups of plant species is indeed true.
In this case, we can also validate the obtained clusterings by comparing them
to the given group assignments using a set of external validation criteria, for
which a few examples are condensed in figure~\ref{fig:external_scores}.

\begin{illustration}
\begin{minipage}[t]{0.5\textwidth}
    \vspace{0pt}
    \includegraphics{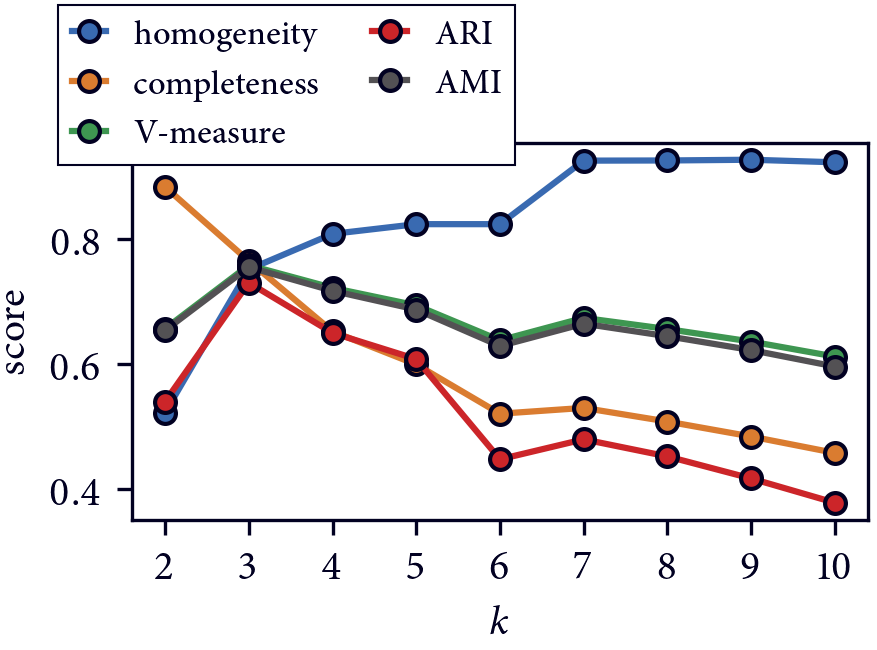}
\end{minipage}\hspace{0.02\textwidth}%
\begin{minipage}[t]{0.48\textwidth}
    \vspace{2\baselineskip}
    \captionofside{figure}{%
        Choosing the \textit{k} in \textit{k}-means (external scores)}{%
        Choosing the \textit{k} in \textit{k}-means (external scores)}{%
            While the homogeneity and completeness
            score are not very telling on their own, a combination of the two in
            the V-measure strongly favours the clustering with \(k = 3\), which
            is confirmed by the adjusted Rand index (ARI) and the adjusted
            mutual information (AMI).}
    \label{fig:external_scores}
\end{minipage}
\end{illustration}

Here, the homogeneity score assesses to what extent each cluster
contains only members of a single class.
The score is bounded by \(0\) and \(1\) where \(1\) is best.
Note, that this is very similar
to homogeneity in terms of low cluster variance, which is optimised in
\(k\)-means under the belief that low-variance clusters represent
distinct (that is homogeneous) classes.
Indeed we see the homogeneity score increasing with larger values for
\(k\).
Complementary, the completeness score measures to what extent all
members of a given class are assigned to the same cluster.
Like the homogeneity score, completeness is bounded by \(0\) and \(1\)
where \(1\) is best, as well.
We see this score decreasing with larger values for \(k\) as more and
more homogeneous groups appear, which, however, reproduce the same
class.
It feels natural to balance these two scores against each other, which is
formalised in the V-measure as the harmonic mean of homogeneity and
completeness.\cite{Rosenberg2007}
For this score, we see an optimal (maximal) value for the
\textit{k}-means clustering with \(k=3\).
Besides that, we can use the Rand index to define a similarity measure between
two clusterings (the obtained one and the underlying truth).
This index considers all pairs of objects in a data set and counts how
many pairs are correctly or incorrectly assigned to the same or to
different clusters based on the provided expected assignments.
The adjusted version of the index (ARI) contains a correction for matching
assignments by chance.\cite{Steinley2004}
The \textit{k}-means clustering with \(k=3\) is most similar to the
expert classification.
Finally, we can use the mutual information between the obtained and the
true cluster label assignments to assess how much information about the
underlying classes is captured by a given clustering, or more generally
speaking how much two clusterings agree.
The adjusted mutual information (AMI) does again account for
expected agreements by chance.\cite{Vinh2010}
We can see, that it is fairly straightforward to quantitatively compare
clustering results with an assumption of ground truth.
The problem with this is just that reference assignments are almost
never available for practically interesting data sets and it would be
very brave to speculate that a \textit{k}-means clustering with \(k=3\)
is universally a good choice even for data sets that are most similar to
the considered case.

Figure~\ref{fig:iris_data_kmeans} shows the result of a \textit{k}-means
clustering with \(k = 3\) on which we could have settled using the
presented validation schemes or because we knew how many clusters there
should be in the first place.
The agreement with the expectation is quite good, as the expected
clusters can be well approximated by compact, globular, well-separated
clusters of roughly equal spatial extent---which is the ideal view of
clusters from the \textit{k}-means perspective.
The \textit{k}-means approach is not suitable for data sets where there
is reason to suspect that clusters can differ from this idealisation, at
least if the aim is to generate a clustering where each cluster represents
a separate interpretable class of objects.
The approach is still valid and widely-used if there is a clear focus on
homogeneity, meaning when it is accepted that true object groups can be
split into multiple \textit{k}-means clusters.
A good example for this, are \gls{md} data sets.
For these it can generally not be assumed that clusters of molecular
conformations are globular, and low within-cluster variance may not be
actually desirable.
Still, \textit{k}-means can be used with a potentially very large number
of clusters with the intention of data discretisation or
condensation.
As such it can for example be used to prepare a state-space for a
\gls{msm} analysis, or as a preliminary clustering step to reduce
the size of a data set from \(n\) points to \(k\) points so that further
(clustering) steps can just operate on the obtained cluster centres.
%

\begin{illustration}
        \includegraphics{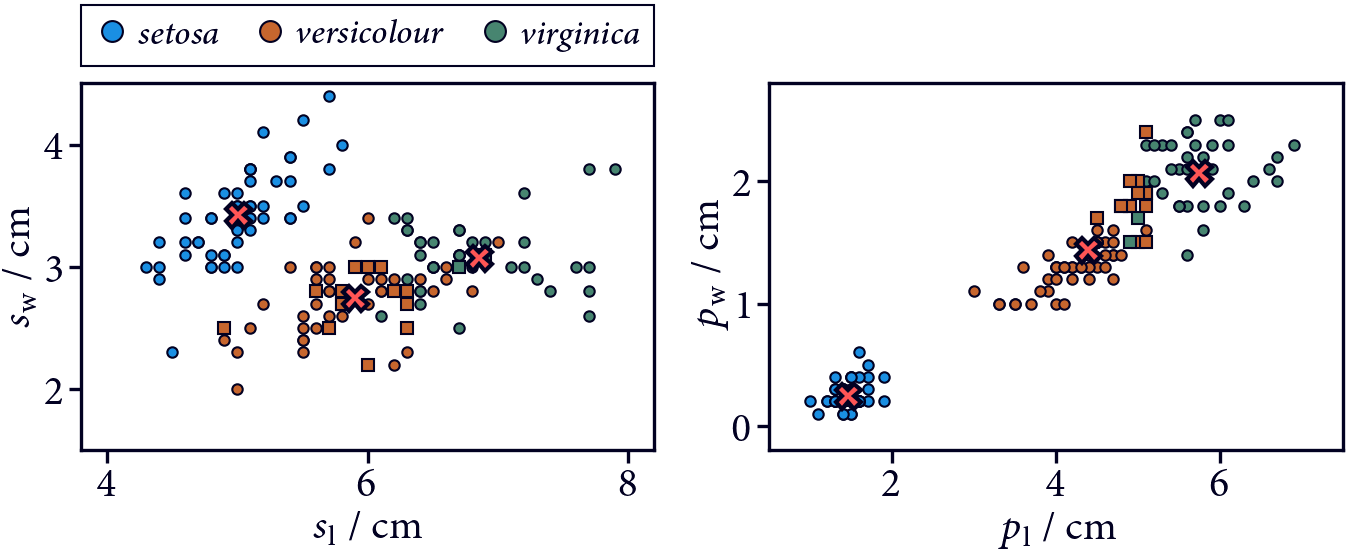}
        \captionofbottom{figure}{%
        \textit{Iris} data set \textit{k}-means (3 clusters)}{%
        \textit{Iris} data set \textit{k}-means (3 clusters)}{%
            Data points (compare
            figure~\ref{fig:iris_data_points_true_labels}) with cluster
            labels found by \textit{k}-means clustering
            (\cminline{sklearn.cluster.KMeans})
            with \(k = 3\). \SI{89}{\percent} of the cluster labels
            match the true classification labels. Non-matching
            assignments are marked with squares. Cluster centres are
            drawn with crosses.}
    \label{fig:iris_data_kmeans}
\end{illustration}

\section{\glsfmtlongpl{gmm}}
\label{sec:gmm}

\firstsecpar{Clustering from a statistical point of view} is nicely
represented by \glspl{gmm} or distribution-based mixture models in
general.\cite{McLachlan2000}
If one has reason to believe that the observed samples in a data set are
the manifestation of several overlapping well behaved probability
distributions, say normal distributions, then it would be only logical
to try to reproduce the observed data by fitting a certain number of
said distributions to the data.\stdmarginnote{model distributions}
It turns out that for the \textit{Iris} data set, this is actually
an excellent approach.
The measured leaf proportions of plants belonging to the three different
classes are indeed likely to be more or less normally distributed, which
means that a good approximate model for the data can be achieved by a
combination of three independent Gaussian distributions (one for each
plant class).
Figure~\ref{fig:iris_data_points_gmm} illustrates the result of this.

\begin{illustration}
    \begin{minipage}[t]{0.75\textwidth}
        \vspace{0pt}
        \includegraphics{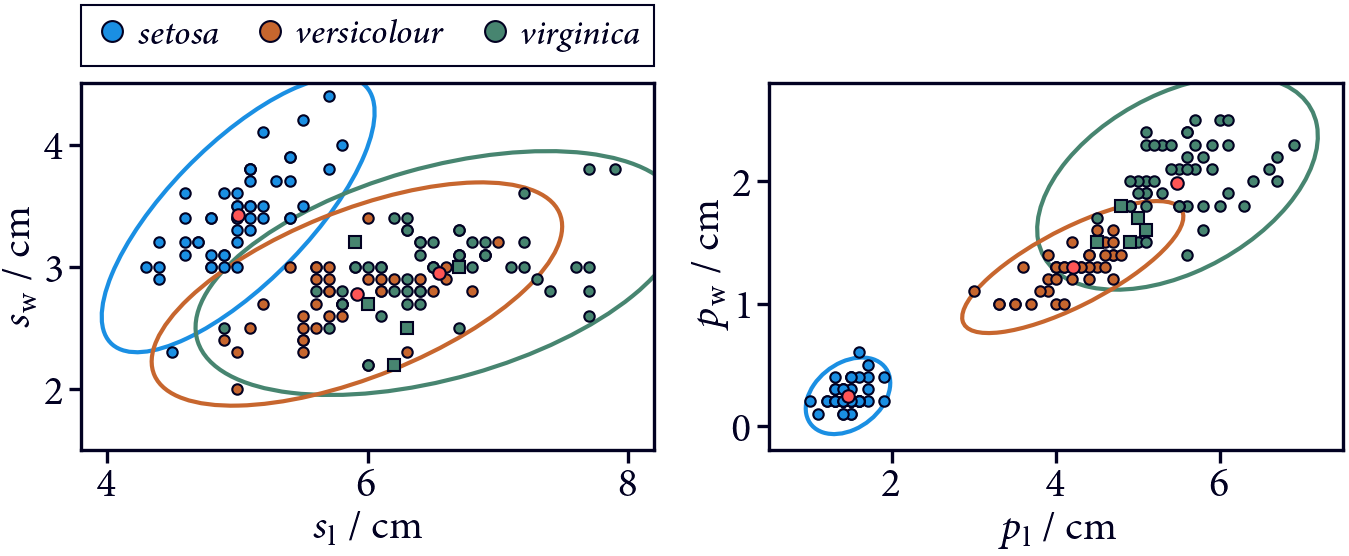}
    \end{minipage}
    \begin{minipage}[t]{0.75\textwidth}
        \vspace{0pt}
        \captionofbottom{figure}{%
            \textit{Iris} data set GMM (3 clusters)}{%
            \textit{Iris} data set GMM (3 clusters)}{%
                Data points (compare
                figure~\ref{fig:iris_data_points_true_labels}) with
                cluster labels predicted from a \gls{gmm}
                (using \cminline{sklearn.mixtures.GaussianMixture}).
                The means of the used
                distributions are marked by red dots while the
                circumference of the ellipses corresponds to three
                standard deviations. \SI{97}{\percent} of the cluster
                labels match the true classification labels.
                Non-matching assignments are marked with squares.}
            \label{fig:iris_data_points_gmm}
    \end{minipage}
\end{illustration}

In principle, mixture models are a form of prototype clustering where
each data point is associated with the prototypical distribution it was
probably sampled from---or with all of the underlying distributions with
respective probabilities.\stdmarginnote{prototypes}
What is achieved, is the decomposition of a full population of samples
into \(k\) homogeneous sub-populations.
There is no directly used similarity concept apart from that points
identified with the same sub-population can be considered similar and
that a sample probability kind of measures a similarity to a specific prototype.
Conceptually, the problem of finding the distributions that model the
observed data best can be formulated in terms of a maximum likelihood
approach.
We want to find the distributions that are most likely to have produced\stdmarginnote{maximum likelihood}
the data set.
Practically, a locally optimal solution to this can be found with
the expectation maximisation algorithm.\cite{Dempster1977}
Under the condition that we know beforehand how many distributions the data
should be modelled with, one can start with an initial guess for
the parameters (mean and variance) of each distribution.
This initial guess can for example be generated with \(k\)-means.
Then the probabilities for data points being sampled from each
distribution are calculated and weighted so that the total probability of
each sample over all distributions sums up to 1.
The parameters are iteratively varied to improve them until convergence.
A clear weaknesses of mixture models is that while it can be a very good
fit for data that truly originates from distributions that are similar
to the chosen ones, it will more or less fail if one picked an inappropriate
prototypical distribution.
In other words, without decent knowledge about the clustered data, it
may be hard to make a well founded decision about this.
The same goes for the number of chosen distributions, although similar
techniques as for \(k\)-means exist on how to tweak the number of
underlying distributions, e.g. via the Bayesian information criterion
(BIC-score).\cite{Scrucca2016}

\section{Density-based clustering using histograms}
\label{sec:density_based_grid}

\firstsecpar{So far we have discussed classic examples for
connectivity-based} and prototype-based clustering methods that may or
may not employ an explicit concept of similarity.
Let's now proceed to \textit{density-based} clustering, meaning methods
that primarily identify clusters as groups of densely packed objects,
rather than certainly behaved object groups.
This type of clustering philosophy is most valuable for the clustering
of molecular conformations and an overview over commonly used techniques
should level the field to fully understand the \gls{commonnn} clustering
approach (see section~\ref{sec:commonnn_theory}.
A clustering method that is on first glance almost trivially simple but
yet very illustrative and with modifications also very powerful, makes
use of (regular) spatial grids.\stdmarginnote{grid-based clustering}
We want to have a look at this method, to demonstrate some of the basic
ideas and variations of density-based clustering and to build the bridge
to connectivity-based and prototype-based cluster models as well as to
flat partitional versus hierarchical clustering.
It turns out that density-based as a trait alone can not provide a model
that is sufficient to cluster data.
It always has to be eventually combined with a notion of connectivity or
prototyping.
Figure\extref{fig:grid_intro}{a} shows a data set of points scattered
in two dimensions.
A density-based view on the data suggests the existence of two
intertwined, sickle-shaped clusters as cohesive regions of high data point
density, separated by low (zero) density.
Formally, we expect the sampled data to be generated by an unknown probability
density function \(\rho : X \rightarrow R_{\geq0}\).
A density-based model for the data depends on an estimate of the true
probability density based on the scattered samples.
One way to achieve such an estimate is to impose a grid of cells with
constant volume upon the data (figure\extref{fig:grid_intro}{b}) and
to count the number of points that fall into each cell, which means
nothing else than to construct a histogram on the data
(figure\extref{fig:grid_intro}{c}).

\begin{illustration}
    \includegraphics{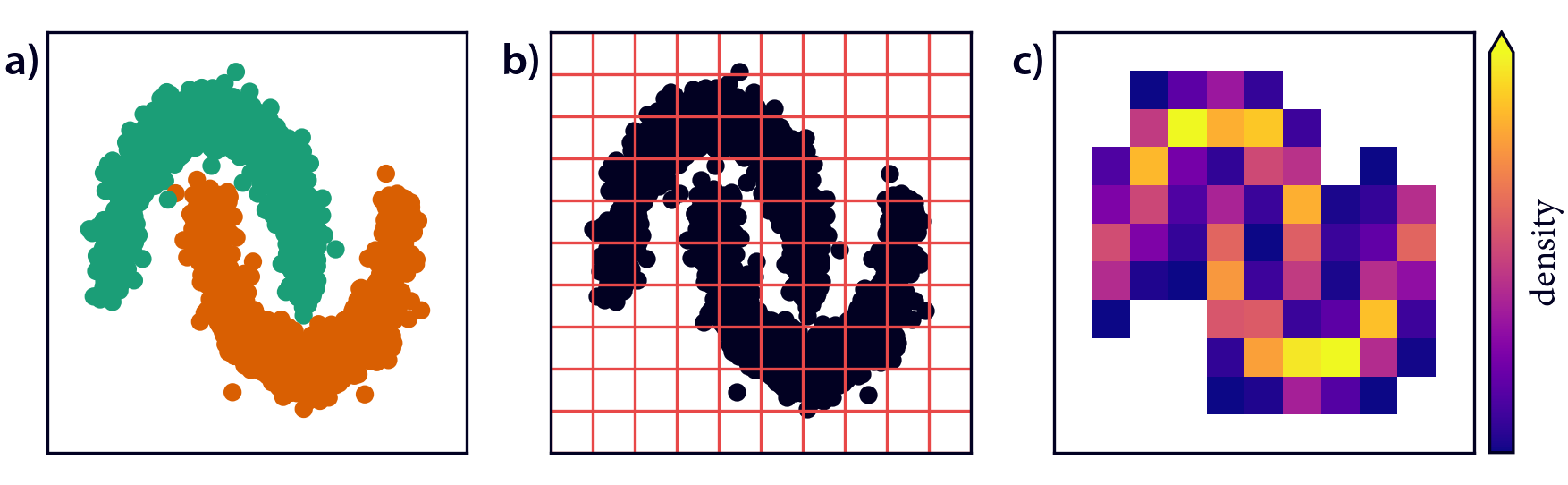}
    \captionofbottom{figure}{%
        Density-based clustering using histograms}{%
        Density-based clustering using histograms}{%
            \subl{a}
            The scikit-learn \textit{moons} data set (2000 points in 2D)
            with true cluster labels indicated by green and orange
            color. \subl{b} A regular grid with 10 bins in each
            dimension imposed upon the data. \subl{c} Cell-wise point
            density estimate (histogram).}
    \label{fig:grid_intro}
\end{illustration}

The process of choosing a grid of cells that bins the data points can
itself already be interpreted as a simple form of clustering in the
sense of a discretisation.
By setting the width of each cell to a constant \(\epsilon\) and by
limiting the extend of the grid in each dimension with \(x_{i,
\mathrm{low}}\) and \(x_{i, \mathrm{high}}\), one obtains a number of
\smash{\(n_{\mathrm{bins}, i} = \mathrm{ceil}\left((x_{i, \mathrm{high}} -
x_{i, \mathrm{low}}) / \epsilon\right)\)} cells in each dimension or
\(\prod_i n_{\mathrm{bins}, i}\) cells in total.\stdmarginnote{regular
grid}
Alternatively, the number of bins per dimension can be set, which determines the bin
width \(\epsilon\).
A cell is uniquely addressed via its location on the grid by an indexing
tuple, e.g. \(i_\mathrm{cell} = (i_1, i_2)\) in the shown 2D-case, or an
indexing function \(i_\mathrm{cell} = i_1 n_{\mathrm{bins}, 1} + i_2\),
which is equivalent to a cluster label for all data points in the respective
cell.
The cells themselves or for example the midpoints of the cells
constitute a form of prototype that individual points are identified
with.
There are many alternative approaches to impose a grid upon objects in a
data set.
%
%
Another rather sophisticated approach to tile objects recursively into
bounding boxes is for example the construction of an \(R\)-tree.\cite{Guttman1984}
It should be noted, though, that for grid cells of unequal volume,
density can not just be estimated as the number of points per cell but
has to be normalised by the cells volume.

The discretisation of the data points into grid cells can in a way be seen
as a pre-clustering of the data.
We use this clustering as a support that provides a way to approximate
the underlying probability density \(\rho\).\stdmarginnote{connectivity}
Note that the grid dimension controls the resolution at which the density is
estimated.
Having a density estimate for portions of the data space based on our
notion of density as shown in figure\extref{fig:grid_intro}{c} is,
however, only half of what is necessary to derive density-based
clusters.
We also need a model for how our final clusters should be constructed,
continuing on the density information.
This could be done conveniently in this case by establishing a
connectivity-based model.
We could for example say that adjacent cells on the grid should be
assigned to the same cluster if they both represent regions of
relatively high density.\stdmarginnote{threshold}
\enquote{Relatively} high implies that we could define a \textit{density
threshold} above which density is considered high.
Two adjacent grid cells within which the threshold is exceeded will be
regarded as connected.
An intuitive way to illustrate this, is to think of the (pre-clustered)
data in terms of a graph structure in which each grid cell is a vertex
and unweighted edges are drawn for pairs of adjacent cells in case their
density is higher than the threshold (figure\extref{fig:grid_threshold_graph}{a}).
In this picture, clusters become immediately discernable as\stdmarginnote{data graph}
\textit{connected components} of the respective graph, which is
displayed in figure\extref{fig:grid_threshold_graph}{b} for an example
threshold.
The cluster labels obtained for the histogram grid cells can then be transferred
back to the original points.
Figure\extref{fig:grid_threshold_graph}{c} shows
the final result of this, which is in agreement with the initially expected
clustering (compare figure\extref{fig:grid_intro}{a}).
A particularity of the clustering is that grid cells below the density
threshold that are not connected to any other cell can be labelled as
noise, i.e. as not part of any cluster.

\begin{illustration}
    \includegraphics{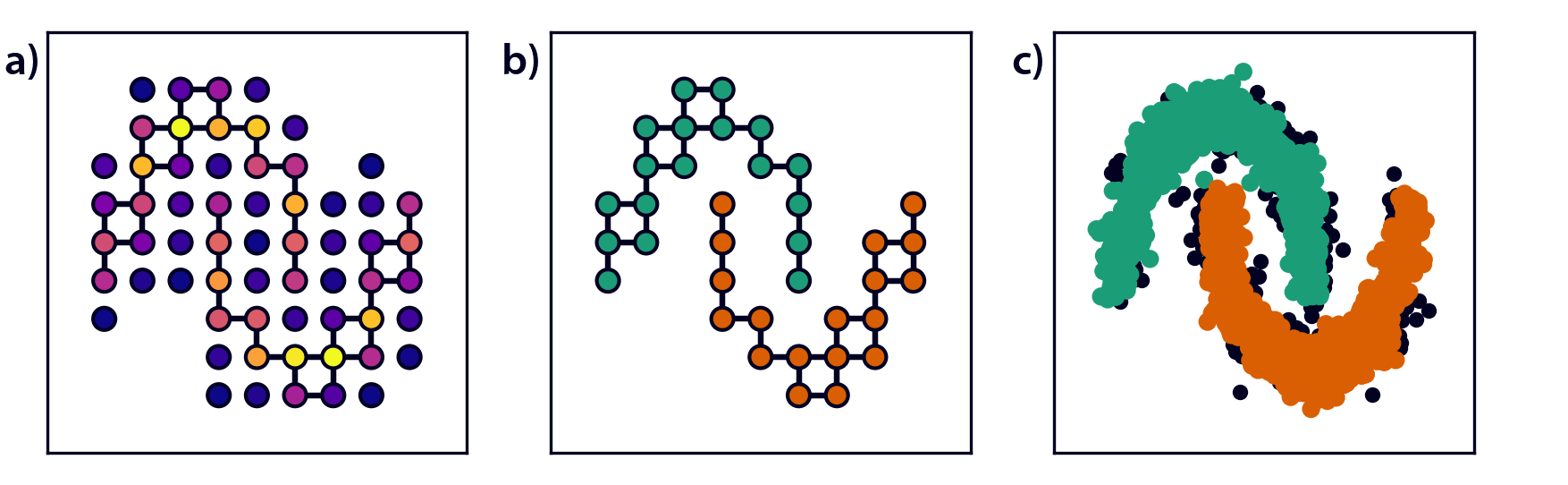}
    \captionofbottom{figure}{%
        Density-based clustering using histograms and a density
        threshold}{%
        Density-based clustering using histograms and a density
        threshold}{%
            A density threshold is applied to identify connected grid
            cells. \subl{a} Histogrammed data represented as unweighted
            graph with edges connecting adjacent grid cell when both
            exceed the density threshold (e.g. here 15 points per cell). \subl{b}
            Clusters identified as connected components of grid
            cells. Grid cells with low density are not assigned to any
            cluster and are omitted. \subl{c} Cluster labels of the grid
            cells translated back to the original data points. Data
            points not assigned to any cluster (noise) are coloured in
            black.}
    \label{fig:grid_threshold_graph}
\end{illustration}

Let's pause for a second and recapitulate how we could characterise the
presented clustering approach.
To begin with, we are using a type of grid to join individual data objects into
groups.\stdmarginnote{How to describe this clustering?}
Based on this, we could say that our method is grid-based and in fact
two typical methods doing something very similar, STING\cite{Wang1997}
and CLIQUE\cite{Agrawal1998}, are commonly referred to as grid-based
methods.
Note, however, that this naming is chosen to emphasise that the
respective methods are supposed to be eligible for the clustering of
large, high-dimensional data sets.
The employed grids are a neat
trick to condense the contained information, so that the clustering can be
carried out on a pre-clustered data set, which is easier to handle.
The finally produced cluster model on the other hand is not actually
grid-based.
The employed grid is exchangeable without altering the method in its
essence, so grid-based is an implementation focused categorisation.
In contrast, it is vital to the method that the grid cells are used to
estimate object density, so we can say that the method
entails a density-based model.
Further, the density-based model is characterised as
connectivity-based because clusters of high density are identified as
networks of connected (adjacent) dense cells.
Finally, we have a threshold-based criterion.
Threshold-based is a categorisation focused on the inductive principle.
The used threshold on the estimated density suggests what the optimal model for
the considered data would be: a partitioning of the data objects according to
their pairwise connectivity, where the threshold determines which connections
exist and which are neglected.
Each object can be reached directly or
indirectly from any other point in the same cluster just by following the
these connections.
This objective is equivalent to a search for maximally connected components in
the data graph from which low density connections where trimmed off.
Finding connected
components in graphs is a routine task that can be solved efficiently by
well established graph traversal algorithms.
A similar threshold-based approach is used by the popular DBSCAN method
(see section~\ref{sec:dbscan})\stdmarginnote{related methods} in its original formulation and can also
be applied to \gls{commonnn} clustering (see
section~\ref{sec:commonnn_theory}).
These methods differ from the presented grid-based approach primarily in
the way how density is estimated and how dense objects are connected
with each other.
In the following section~\ref{sec:level_set_method}, the unifying concept of
level-sets that underlies all of these methods will be discussed briefly.
The connectivity between object vertices in the implied graph structure
for the data can be interpreted as a form of similarity.
Two connected points are similar with respect to how the connection was
defined in the connectivity-based model.\stdmarginnote{similarity}
When the density threshold is applied to select valid connections,
similarity can be viewed as a binary relation, that is two objects can
either be similar (because they pass the threshold criterion) or not.
In a relaxed view, it would also be possible to state that all points
within the same cluster are similar to each other if one allows the
following reasoning: if two points \(a\) and \(b\) are connected
(similar) and \(b\) is connected to a third point \(c\), then \(a\) is
also (indirectly) connected to \(c\).
Note, however, that this form of similarity is detached from similarity
measures with respect to the data space the clustered data points where
originally embedded in.
This means that two points in the same cluster can be actually very dissimilar
(far away from each other), and vice versa two points in different clusters can
be actually very similar (close to each other), from a different perspective on
similarity.
Interestingly, the actual data space plays only a minor role for the
presented grid-based approach.
The decisive element for the clustering\stdmarginnote{latent space} is the
density estimate that does only require the grid, which in turn is in general
any kind of discretisation---of not necessarily known origin.
There is no direct dependence on another concept of similarity, neglecting
that the needed discretisation of the space may involve one.
In this sense, the clustering can be said to operate on a latent data space.

\custompar{For threshold-based approaches} to work with a density estimate on a
data set, the clustering result is obviously dependent on the choice of the
threshold.
\stdmarginnote{choice of threshold}
For lower thresholds, the two disjoint components of the graph as
shown in figure~\ref{fig:grid_threshold_graph} may for instance become
connected.
In this case, only one cluster will be found.
For even higher thresholds, the two components are expected to shrink since
lower density grid cells on their outer rims will fall below the minimum
density requirement.
Where to put the threshold, can be understood as a tuning parameter.
It is left to the user of the clustering to select an ideal value in the
context of a specific application and possible expectations.
There is no universal concept of what a suitable threshold would be.
Effectively, a variation of the threshold criterion creates a hierarchy of
clusterings of systematically varying granularity.
From this perspective, the applied threshold acts as termination
criterion on an intrinsically hierarchical cluster model.\stdmarginnote{cluster
hierarchy}
It provides a strict inductive principle that prefers a model at a specific
level of the hierarchy and turns the method into a quasi-flat clustering.
Yet, it is also possible to not apply a threshold on the estimated density and
to obtain the actual hierarchy of clusterings instead.
Let's recall that the use of a density threshold gave us a binary
perspective on pairwise object similarity (connectivity) where the
threshold was exceeded.
To relinquish the threshold means to find a quantitative description of
connections between grid cells as weighted edges in the respective data
graph.
In this picture, objects can be more or less similar.
A simple variant of this could be to use the minimum density of two
connected cells as a weight for their connection.
In this way, a pair of cells in which one of the cells represents a low density
data region will consequently have a low edge weight on the connection between
the pair, rendering it of low importance (note that it is in principle
arbitrary whether high density equates to low or high edge weight).
A hierarchy of clustering results is than basically given by reviewing
all connections in the graph in the order of their importance, for
example by beginning with the least important one and dropping one
connection at a time.
As more and more connections are ignored each level in the hierarchy
represents a new clustering.
Starting with all objects in the data set in one cluster when still all
connections are considered, the hierarchy is ending with each data object in a
separate cluster (i.e. as noise) once the last connection has been dropped.
In reverse, it is also possible to begin with adding the most important
connection and to proceed with lower weight edges.
Figure~\ref{fig:grid_intro_hierarchy} shows the data graph with weighted
edges and the resulting hierarchy of clustering results for our example
case.
From this, it is now clear at which density threshold a splitting of the
data set into sub-clusters can be observed exactly.
Note that here clusters need to contain more than two cells to be
counted as such and are considered noise otherwise.
For this demonstration, a connected component search was performed on a
graph at each hierarchy level after adding all edges of a certain
weight.
It should be mentioned that this is in general very inefficient and that
other approaches can be used, for example by leveraging \glspl{mst}.
Using a \gls{mst} of density weighted edges transforms the
clustering into nothing else than single-linkage clustering.

\begin{illustration}
    \figureindent\includegraphics[trim=0 0 0 0, clip]{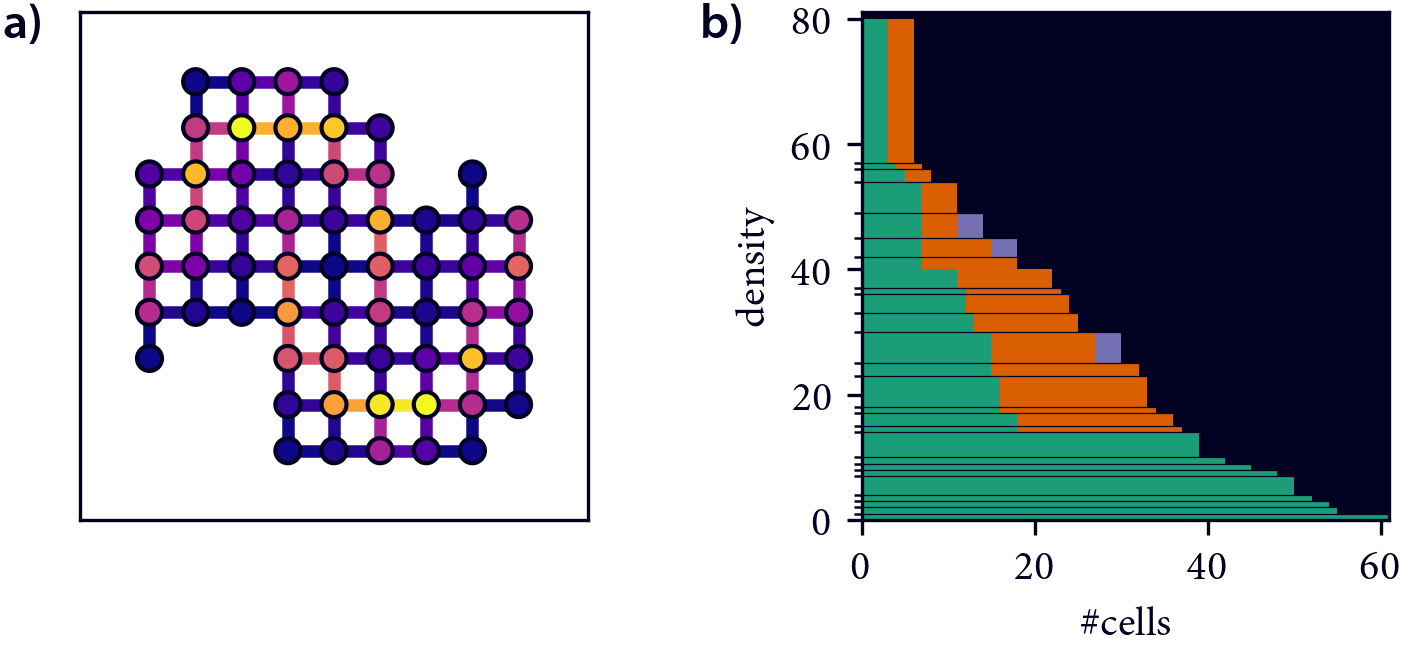}
    \captionofbottom{figure}{%
        Density-based clustering using histograms hierarchically}{%
        Density-based clustering using histograms hierarchically}{%
            Not applying a threshold, the full hierarchy of clusterings
            can be built instead. \subl{a} Data set represented as
            weighted graph with edges connecting adjacent grid cells
            quantitatively. \subl{b} Hierarchy of clusterings. Each edge
            weight on the \(y\)-axis gives rise to a new hierarchy level
            when edges of the corresponding weight are removed from the
            graph (or added). Starting at the bottom with all grid cells
            in one cluster (all connections are present), the graph
            becomes more and more disconnected at higher levels until
            all grid cells are separated at the top. The absolute number
            of cells assigned to each cluster at each level is
            represented with coloured horizontal bars where black stands
            in for noise.}
    \label{fig:grid_intro_hierarchy}
\end{illustration}

\section{Density-based clustering using level-sets}
\label{sec:level_set_method}

\firstsecpar{Connectivity-based density-based clustering} like introduced
in the last section can be uniformly expressed with the help of level-sets.
A level-set of a function \(f(x)\) is in principle just the set of
points \(L = \{x\,|\,f(x) = \lambda\}\) for which the function takes on a
certain value \(\lambda\).
The use of level-sets for the definition of density-based clusters is
usually attributed to Hartigan.\cite{Hartigan1975}\stdmarginnote{level-sets}
Let \(\rho(x)\) be an underlying probability density on a continuous
data space \(X\) (practically usually a subset of \(\mathbb{R}^d\)) from
wich a data set \(\mathcal{D} = \{x_1, ..., x_n\,|\,x_i \in X\}\) can be
sampled.
Then, a level-set of \(\rho\) using a density criterion \(\lambda\)
corresponds to a density iso-surface (in 2D a contour-line) on this
density function.
A super level-set (also often called an upper level-set or just a
level-set) is furthermore defined as \(L = \{x \in X\,|\,\rho(x) \geq
\lambda\}\), i.e. the set of points for which the density exceeds the
specified \(\lambda\)-threshold.
Density-based clusters are the connected components of such a super
level-set.
The sub level-set, the domain of the function for which the density
falls below the threshold, is ignored for the cluster assignment and
treated as a \enquote{noise}, outlier, or \enquote{fluff} region.\cite{Stuetzle2010}
A clustering based on level-sets does not yield a full partitioning of a
data set.
While a clustering can be achieved with a respective \(\lambda\)-value,
the level-set formulation makes it obvious that there is actually
a hierarchy of connected components corresponding to a continuous variation
of \(\lambda\) in the interval \([0, \max(\rho(x))]\).
It can be shown that this leads to a finite level-set tree of clustering
results.\cite{Steinwart2011}
A property of this tree is for example that for any connected component
\(C_\mathrm{child}\) obtained at a density value \(\lambda\), there is
exactly one connected component \(C_\mathrm{parent}\) with\stdmarginnote{level-set trees}
\(C_\mathrm{child} \subset C_\mathrm{parent}\) for any \(\lambda^\prime
< \lambda\), that is there is a strict child-parent relation between
connected components at different levels.
Furthermore, if of two connected components one is not a subset of the
other, they do not overlap at all.
There is a large of amount of level-set tree related theory, on how to
construct and analyse them, and on how to estimate them from scattered
data.\cite{Chaudhuri2010,Klemela2018}
Density-based clustering of point samples is one possibility to estimate
level-set trees for unknown probability density functions.
Of particular interest might also be the connection between
density-based clusters and single-linkage clustering (using
\glspl{mst}).\cite{Hartigan1981,Stuetzle2003,Stuetzle2010}
Figure~\ref{fig:level_set_tree} shows an example of a level-set tree on
a 1-dimensional multimodal distribution for which \(\rho(x)\) is known.
In this case one can construct the tree of connected components
quasi-analytically for different \(\lambda\) values just by finding the
respective level-sets, i.e. the points where the threshold and \(\rho\) intersect.
The plot represents the size of each component and its mean location.\stdmarginnote{1D example}
For an analysis aiming on the extraction of clusters, one is
usually most interested in the \(\lambda\)-values at which the number
of connected components increases.\cite{Steinwart2011}
In the example, this is the case for \(\lambda_1\) and \(\lambda_2\).
Up to \(\lambda_1\) there is only a single connected region under the
curve of the function, which shrinks from the borders when \(\lambda\)
is increased starting from 0.
Above \(\lambda_1\), the minimum between the rightmost peak and the rest of
the function falls below the threshold and leaves the two respective
regions as disjoint sets.
Note that once \(\lambda_3\) is reached, the maximum of the right component
falls below the threshold and the component vanishes.
One single threshold value is not able to select a partition of three
components corresponding to the peaks of the distribution.
The intuitively correct clustering result is a combination of the
components from the orange, red, and green branch of the level-set tree.

\begin{illustration}
    \begin{minipage}[t]{0.51\textwidth}
        \vspace{0pt}
        \includegraphics{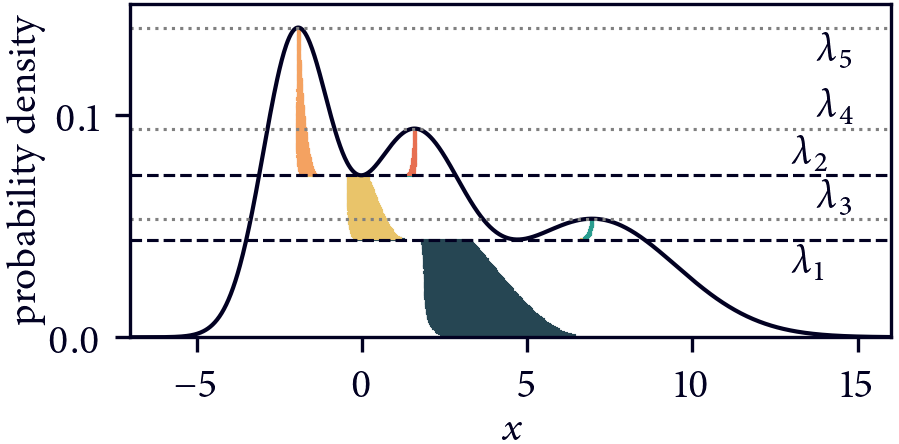}
    \end{minipage}%
    \begin{minipage}[t]{0.49\textwidth}
        \vspace{0pt}
        \captionofside{figure}{%
            Level-set tree for a 1D multimodal Gaussian distribution}{%
            Level-set tree for a 1D multimodal Gaussian distribution}{%
                Slicing the shown probability density distribution along
                a \(\lambda\) threshold value, we obtain corresponding
                clusters as the connected components of the respective
                super level-set. For each component, its size is
                represented by a coloured horizontal bar scaled in width
                by the integral \(\int_{x_1}^{{x_2}} \rho(x)\), where
                \(x_1\) and \(x_2\) are the components bounds. The bars
                are centred around the mean of the component.}
        \label{fig:level_set_tree}
        \vspace{0.5\baselineskip}
    \end{minipage}
\end{illustration}

When density-based clusters should be discovered from discrete point
samples in accordance with the level-set approach, one essentially needs
a density estimate \(\rho^\prime\) for the unknown underlying
probability density \(\rho\).\stdmarginnote{density estimate}
The connected components of the discrete level-set \(L^\prime = \{x
\in \mathcal{D}\,|\,\rho^\prime(x) \geq \lambda\}\), that is the identified clusters,
are an approximation to the corresponding components of the true density.
Naturally, the validity of a clustering does therefore depend on a
consistent density estimate and beyond that on a robust definition of
connectivity between data points in \(L^\prime\).
A generic realisation of connectivity that does in principle work with arbitrary
density estimates is, to give another example, found in the following approach.
Consider a Voronoi partitioning of the \(n\) points in a data set \(\mathcal{D}\) into
\(n\) cells, i.e. a spatial tessellation in which any other point not in \(\mathcal{D}\)
would fall into the same cell as its respectively closest neighbour in \(\mathcal{D}\).
Then take the Delaunay triangulation by connecting each data point to other data\stdmarginnote{Delaunay connectivity}
points that are in adjacent cells.
Given a point-wise density estimate, for example the reciprocal
\(k\)-nearest distance (with say \(k = 1\)), one can drop those points
from the just created network that fall below a specified density
threshold and reveal the clusters as the remaining connected components.\cite{Azzalini2014}
Note that this is very similar to the previously discussed grid-based
example only that the grid served there as the density estimate as well
while it does here primarily establish connectivity.
A slightly different realisation using a \(k\)-nearest neighbours graph from
which low density nodes are removed at each iteration of \(\lambda\) is
available with DeBaCl clustering.\cite{Kent2013}
The following sections will discuss connectivity-based density-based
clustering procedures that solve the problem of establishing a density
estimate and connectivity differently.

\section{DBSCAN}
\label{sec:dbscan}

\firstsecpar{A very popular density-based clustering} method is
DBSCAN,\cite{Ester1996} which stands for \textit{density-based spatial
clustering for applications with noise}.
Occasionally, DBSCAN and density-based clustering are even used as
synonyms in the literature.
As mentioned in section~\ref{sec:density_based_grid}, the method is
conceptually very similar to the previously described density-based
clustering using grid cells.
Virtually, the only difference lies in how density is estimated based
on the samples in a data set.
Instead of the number of objects per cell, DBSCAN takes the number
of neighbouring points as a density estimate for individual points.
Typically, the neighbourhood of a point is the open- or
closed-ball neighbourhood \(\mathcal{B}_r\) as defined in
and~\ref{eq:closed_ball_neighbourhood}, using a distance metric
(usually the Euclidean distance) and a fixed radius \(r\).
The cardinality of the set \(\mathcal{B}_r\) provides the density
estimate for a single point.
Frankly speaking, the volume element corresponding to a neighbourhood
can also be viewed as a special type of cell, making the relation to
grid-based clustering quite obvious---only that the cells are centred
around individual points and partially overlapping.
Just as in grid-based clustering, the density-estimate alone does
not suffice to group data points into clusters.
We still need a connectivity concept to decide when two points
should be part of the same cluster.
In the original formulation of DBSCAN,\cite{Ester1996} this is
defined leveraging a threshold.
Points that possess a density estimate exceeding the threshold, i.e.
that have at least a number of \(n_\mathrm{c}\) neighbours with respect
to a neighbour search radius \(r\), are called \textit{core points}.
Core points that are neighbours of each other are in turn considered to be
connected.
Fundamentally, the set of all present connections constitutes a graph
for the data and clusters are the maximally connected components of this
graph.
This notion is in line with the formulation of the clustering problem in
terms of level-sets (compare section~\ref{sec:level_set_method}).
Additionally, points that are not themselves core points but neighbours
of a core point are termed \textit{border points}.
Connections of border points to core points can optionally be added to
the graph as well, but note that this introduces some ambiguity because
border points may be connected to core points in different clusters.
In this case, it has to be decided to which cluster a point should be
assigned, which can be done randomly or for example by choosing the
closest core point or by evaluating the membership of all neighbouring
core points against each other.
Figure~\ref{fig:dbscan_moons} illustrates the DBSCAN density criterion
with a threshold
and a clustering result for the previously used scikit-learn
\textit{moons} data set.

\begin{illustration}
    \includegraphics[]{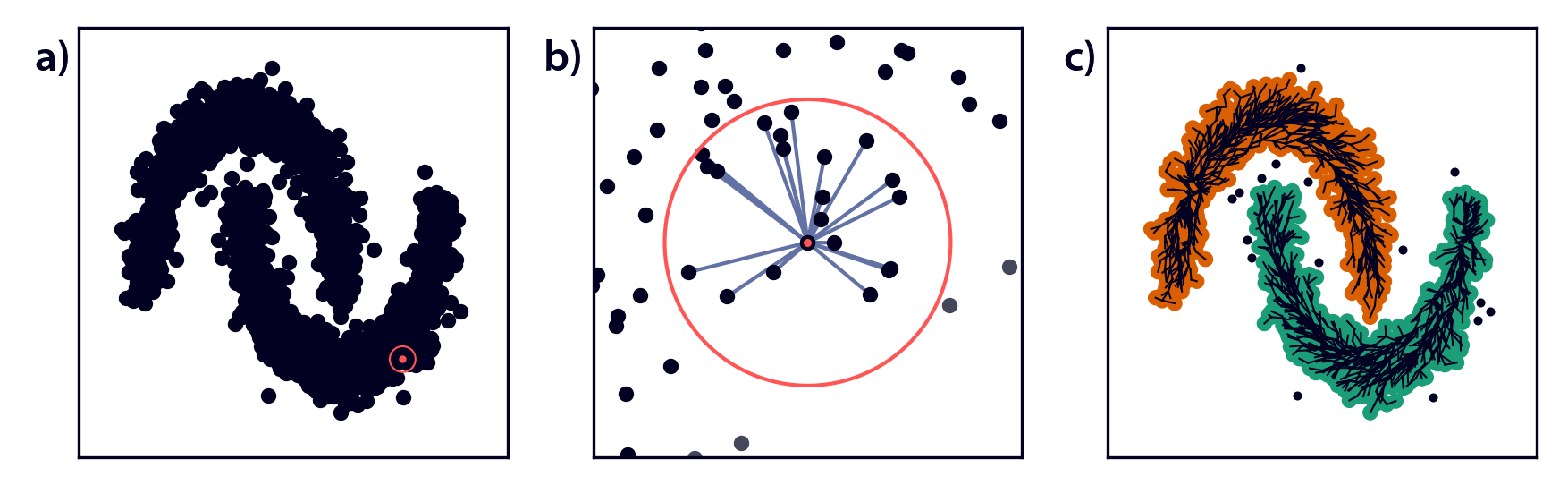}
    \captionofbottom{figure}{%
        DBSCAN for the \textit{moons} data set}{%
        DBSCAN for the \textit{moons} data set}{%
            \subl{a} Original data points with neighbour search radius
            \(r=0.15\) shown for a single point with a red circle. \subl{b} For
            a density threshold of \(n_\mathrm{c} = 10\), the point highlighted
            in \subl{a} is a core-point since it has 20 neighbours. All its
            neighbours are connected to it as indicated by blue lines. Border
            points are drawn in gray. \subl{c} The connections between the data
            points form a graph in which clusters correspond to maximally
            connected components. The data points are coloured by their cluster
            labels (noise in black). The edges of a graph build from a connected component
            search (see code snippet further below) are shown with black lines.}
    \label{fig:dbscan_moons}
\end{illustration}

The DBSCAN publication coins the terms \textit{density-reachable} and
\textit{density-connected}.
A point \(b\) is directly density-reachable from a point \(a\) if \(a\)
is a core point and \(b\) is a neighbour of \(a\).
\stdmarginnote{density-reachability and connectivity}
Furthermore, a point \(c\) is indirectly density-reachable from \(a\) if
there is a chain of directly density-reachable points from \(a\) to
\(c\), say there is a point \(b\) that is reachable from \(a\) while
\(c\) is reachable from \(b\).
Finally, two points are density-connected if there is a point from
which both are (indirectly) reachable.
Density-connectivity is a symmetric definition while density-reachability
is only symmetric for core points.
Clusters are identified as sets of density-connected points.
The paper does not identify this with the concept of connected
components of a graph in which edges correspond to the direct
reachability of two data points but the notion is exactly equivalent.
Practically, DBSCAN can be implemented using a connected component
search algorithm like \glsfirst{bfs}.
Starting with any point of the data set as the root of the first
cluster, the full cluster can be explored iteratively by adding all
points that are connected to this first point, then adding all points
connected to the newly added points and so forth.\stdmarginnote{implementation}
Once no other point can be added, the next cluster can be explored by
choosing a new root if there are still unassigned points left in the
data set.
During such a search, it is possible to either build a graph of
connections explicitly or to just directly assign cluster labels to
data points.
The short code snippet below gives an example for how to do both at the
same time.
The example assumes that core points have been previously identified
by checking the neighbour count against a threshold.
As usual, the algorithm depends on a FIFO queueing structure to collect
points from which the cluster can be grown further and an indexable
indicator structure to keep track of which points have been already
visited.
The graph in figure\extref{fig:dbscan_moons}{c} has been generated in
this way.
Note that this graph contains only a (minimum) set of required
connections that depends on how the graph is explored.
Note also that the code example will add border points to the first
possible core point if the used neighbour lists contain not only core
points.

\pagebreak
\begin{lstlisting}
  clabel = 1
  for p in core_points:
    if visited[p]:
      continue
    visited[p] = True
    labels[p] = clabel
    queue.push(p)

    while queue:
      p = queue.pop()
      for q in neighbours[p]:
        if visited[q]:
          continue
        visited[q] = True
        labels[q] = clabel
        graph.add_edge(p, q)
        queue.push(q)

    clabel += 1
\end{lstlisting}

As a side note, one of the earliest density-based clustering method,
which was proposed by Wishart in 1969,\cite{Wishart1969} does something
very close to DBSCAN with a different notion of how to
establish\stdmarginnote{Wishart variant} connectivity.
The idea is to identify core points using the same density estimate, but
than using classic single-linkage clustering on these core points to
find the clusters.
\custompar{Like for the grid-based approach}, the choice of the threshold
value tunes the outcome of the clustering in DBSCAN.
A complete screen of the threshold results in a hierarchy\stdmarginnote{hierarchy}
where higher thresholds create splits in denser data regions while
larger portions of the data fall below the threshold and are declared
noise.
Choosing a suitable threshold can, however, be unintuitive and
cumbersome, not to mention that a systematic test of many different
thresholds can be fairly expensive for larger data sets.
There exists a number of possible approaches to use the DBSCAN density
estimate without a threshold instead, though.
The equivalent to what was described for the grid-based example, would
be to define a weight for connections between two data points based
on their density estimate.
This could be the minimum density of the two points.
An evaluation of all connections ordered by their weight will then lead
to the complete hierarchy of clusterings where each hierarchy level
corresponds to a specific density threshold
(compare~\ref{fig:grid_intro_hierarchy}).
In this case, one needs to additionally define when two points should be
connected to each other in the first place.
For the grid example, connections where considered only between adjacent
cells.
Correspondingly, connections can be considered only between neighbouring points
here.
Note, however, that this in turn again depends on the neighbour search
radius \(r\), which kind of acts as a resolution parameter comparable
to the bin size in the grid case.
Another variation that turns the DBSCAN concept upside down was proposed
with HDBSCAN.\cite{Campello2013,McInnes2017}
The idea is to transform the point-wise density estimate into a new\stdmarginnote{HDBSCAN}
metric called \textit{mutual reachability distance} that can be used as
a connection weight between data points.
In conventional DBSCAN, the question is \textit{\enquote{how many neighbours
does a point have?}} or rather \textit{\enquote{does a point have at least
\(n_\mathrm{c}\) neighbours?}}.
In HDBSCAN, this is turned into \textit{\enquote{how large does \(r\) need to be
so that a point has at least \(n_\mathrm{c}\) neighbours?}}
The neighbour search radius \(r\) at which a point fulfils this density
criterion is called the point's core distance \(d_\mathrm{core}\)---the
radius for which a point becomes a core point---and is equal to the
\(k\)-nearest distance for \(k = n_\mathrm{c}\), i.e. the distance to
its \(n_\mathrm{c}\)th closest neighbour.
The mutual reachability distance between two points \(a\) and \(b\) is
then defined as
\begin{equation}
    d_\mathrm{mutual}(a, b) = \max\left(d_\mathrm{core}(a), d_\mathrm{core}(b), d(a, b)\right)\,,
    \label{eq:mutual_reachability_distance}
\end{equation}
where \(d(a, b)\) is a regular (the Euclidean) distance between the points.
Effectively, points that are in relatively sparse data regions and have
large core distances are pushed further away from other points.
Dense points with low core distances remain at their original distance to
other dense points.
DBSCAN can be reformulated as single-linkage clustering on this new
metric.
In practice, however, the set of all pairwise connections between the
data points can be reduced to a minimal set of relevant connections,
namely a \gls{mst}.
Bottom-up iteration over the edges of this \gls{mst}
starting with the two most closely connected points creates the
hierarchy of clusters.
A certain slice of the hierarchy is exactly what is achieved with
conventional DBSCAN with a certain density threshold.
Unfortunately, single-linkage hierarchies can be complex for larger data sets.
For \(n\) data points, the \gls{mst} has \(n-1\) edges and
hence the cluster hierarchy comprises \(n-1\) merges.
The number of merges one eventually might be interested in, can be much
smaller, though, because many of them typically correspond to a
situation where two small clusters merge or where a small cluster is
swallowed by a big one.
To narrow the number of merges down to those where two big clusters are
joined, one can define a minimum cluster size.
Clusters with a member count lower than this minimum requirement can be
regarded as noise.
A merge of a noise cluster can be simply seen as cluster growing and
can be ignored in the hierarchy of merges.
Figure~\ref{fig:hdbscan_iris_trees} shows an example application of HDBSCAN to
the \textit{Iris} data set including the single-linkage hierarchy obtained
directly from the \gls{mst} of mutual reachability and a condensed
hierarchy using a minimum size for relevant clusters.

The full hierarchy of clustering results is more powerful than individual
flat clusterings.
For one thing, the hierarchy can guide the choice of again a simple threshold
to eventually extract a certain slice of the hierarchical
tree.\stdmarginnote{analysing hierarchies}
On the other hand, the hierarchy can be processed to select a final clustering
result as a combination of clusters from different branches of the tree, which
means with possibly different thresholds for each cluster.
This selection of child clusters can be just done rationally by the user but
HDBSCAN does also provide a heuristic automatic approach for it.
Starting with the outer leaf clusters, a persistence, or in other words
a live time, is determined for each cluster in terms of the threshold range
in which the cluster exists.
If the live time of a parent cluster is longer than the summed live times of
its children, the parent will be kept as the more relevant cluster.
There are several alternatives to process cluster hierarchies
programmatically also in the context of \gls{commonnn} clustering.

\begin{illustration}
    \includegraphics[]{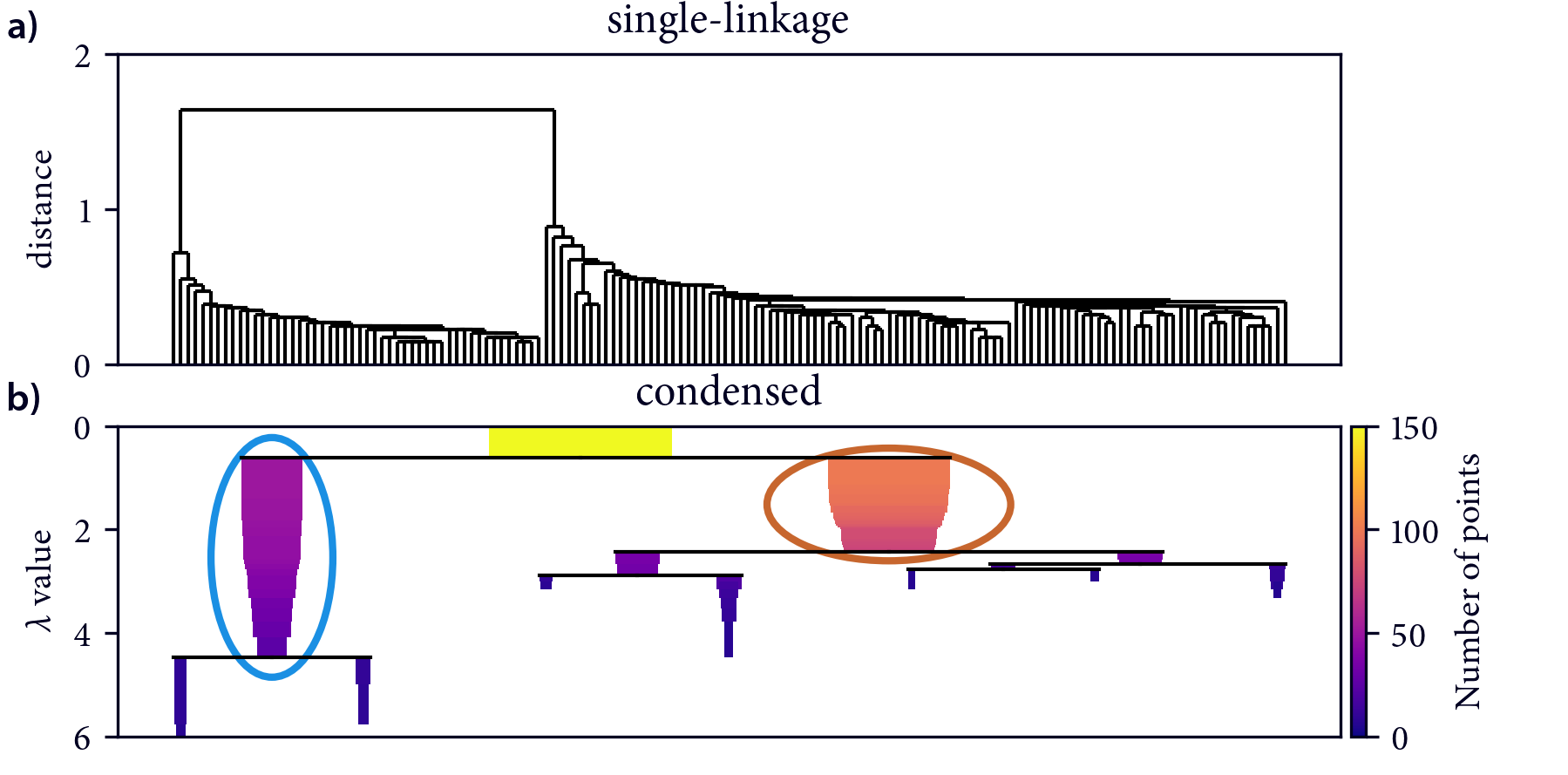}

    \includegraphics[]{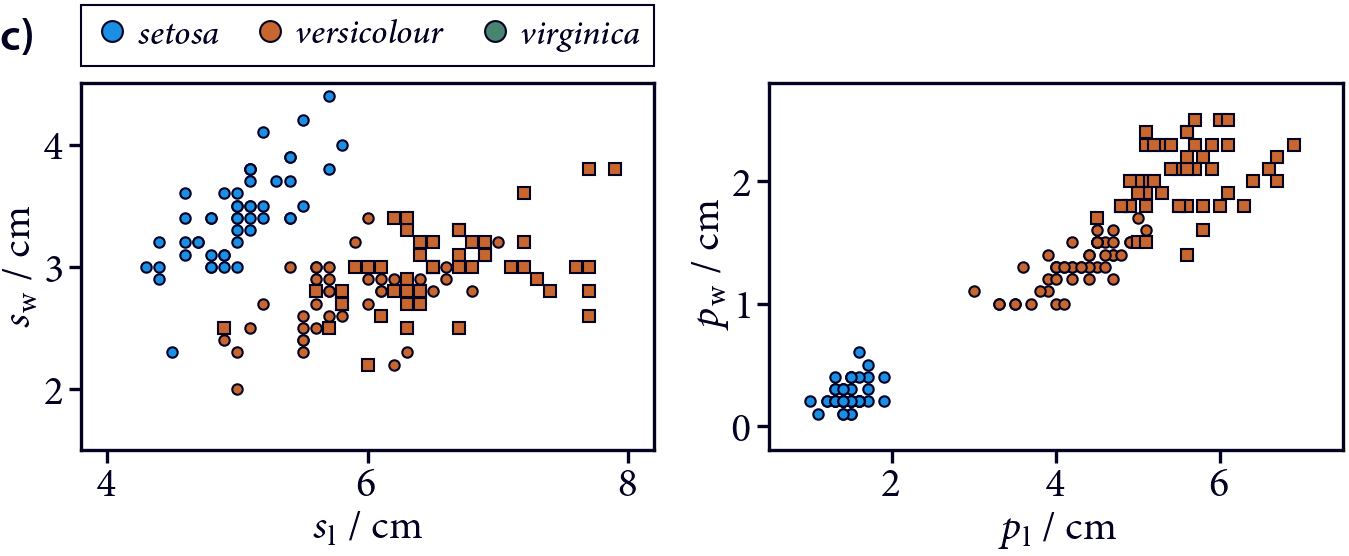}
    \captionofbottom{figure}{%
        \textit{Iris} data HDBSCAN hierarchy and clustering result}{%
        \textit{Iris} data HDBSCAN hierarchy and clustering result}{%
            \subl{a} Dendrogram for a single-linkage clustering on a
            \gls{mst} using the mutual reachability distance for
            \(n_\mathrm{c} = 2\) (see
            equation~\ref{eq:mutual_reachability_distance}) and \subl{b}
            condensed tree applying a minimum cluster size value of 5.
            Note that while the single-linkage tree is labelled on the
            \(y\)-axis with the actual distance value at which a
            respective merge can be observed, the condensed tree is
            given in values of \(\lambda = 1/d_\mathrm{mutual}\), which
            is an efficient density estimate. Based on the relative
            persistence of the clusters in the condensed tree, HDBSCAN
            (\cminline{hdbscan.HDBSCAN})
            suggests a preferred clustering result as a combination of
            clusters from different branches of the tree (highlighted
            with circles around them). \subl{c} Recommended clustering
            result based on cluster persistency. \SI{67}{\percent} of
            the cluster labels match the true classification labels.
            Non-matching assignments are marked with squares.}
    \label{fig:hdbscan_iris_trees}
\end{illustration}

Of course, it still depends on the application if such a processing of a
hierarchy leads to a desired outcome.
For the \textit{Iris} data set, the HDBSCAN live time rational prefers two
clusters, which is not well in agreement with the biological assignment (compare
figure~\ref{fig:hdbscan_iris_trees}
and~\ref{fig:iris_data_points_true_labels}).
Figure~\ref{fig:dbscan_iris_selected_threshold} shows a more agreeable
clustering for which the threshold was selected based on the hierarchy
as the smallest value where three clusters can be obtained.
Finding this threshold without the hierarchy would necessitate a manual
try-and-evaluate approach with different threshold values.

\begin{illustration}
        \includegraphics{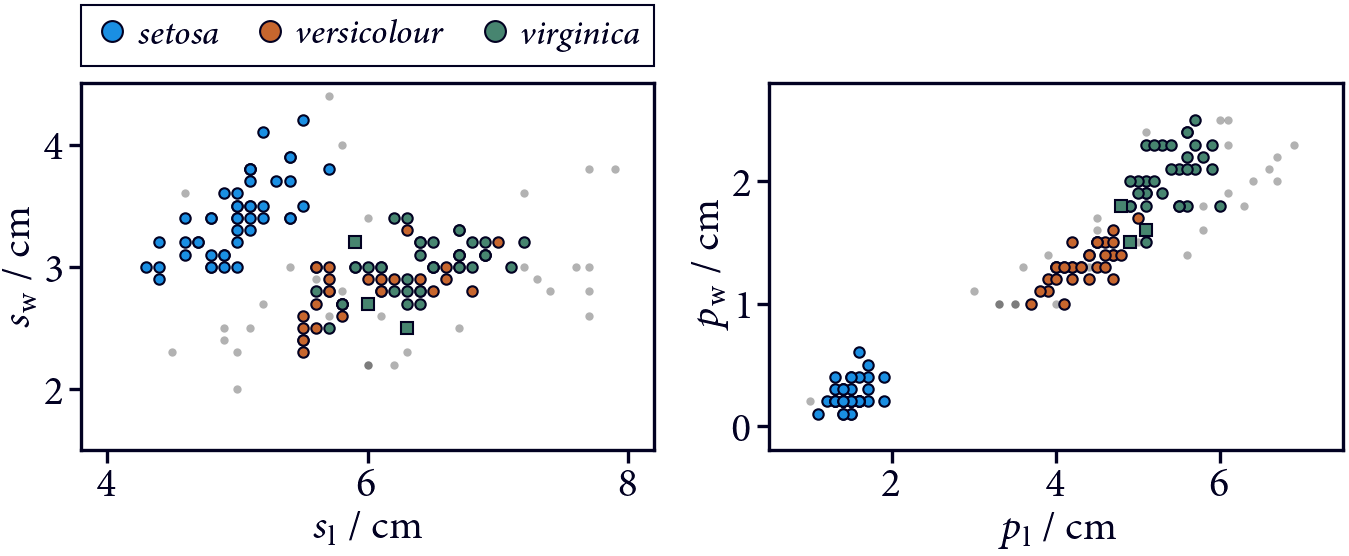}
        \captionofbottom{figure}{%
        \textit{Iris} data DBSCAN with a threshold selected from the hierarchy}{%
        \textit{Iris} data DBSCAN with a threshold selected from the hierarchy}{%
            Investigating the single-linkage hierarchy (compare
            figure~\ref{fig:hdbscan_iris_trees}), it is possible to select
            \(d_\mathrm{mutual} = 0.4\) as a threshold value at which the data
            set is partitioned into three clusters of similar size. The
            respective cluster labels can be extracted from the HDBSCAN
            hierarchy. Conventional DBSCAN
            (using \cminline{sklearn.cluster.DBSCAN})
            with \(r = d_\mathrm{mutual}\) and \(n_\mathrm{c}=2\) gives the
            same result. \SI{81}{\percent} of the cluster labels match the true
            classification labels (\SI{98}{\percent} if noise points are
            neglected). Non-matching assignments are marked with squares.}
    \label{fig:dbscan_iris_selected_threshold}
\end{illustration}

\section{Jarvis-Patrick clustering}
\label{sec:jp_clustering}

\firstsecpar{In the previous sections}, we discussed density-based clustering
protocols that defined the notion of how density is estimated using either a
certain kind of volume elements (grid cells) or the neighbourhoods of
individual data points.
For the actual identification of clusters, a separate introduction of a
connectivity concept was necessary in these cases because the entities for
which the density is estimated have no intrinsic density-related relationship
to each other.
Connectivity was derived by mixing another type of relation (the adjacency of
grid cells or the neighbourhood relation of points) with the density estimate.

\enlargethispage{2\baselineskip}
The Jarvis-Patrick clustering methodology is relatively unpopular (at least in
comparison to DBSCAN) but it does in contrast provide a density estimate that
directly serves as a connectivity definition as well.\cite{Jarvis1973}\stdmarginnote{density
estimate}
Density is taken as the number of neighbours that two points share with respect
to their \(k\)-nearest neighbourhood.
Since this density estimate involves two points, it establishes a connection
between them.
The definition is also sometimes referred to as the \textit{shared
nearest neighbour} (SNN) similarity.

\begin{illustration}
    \includegraphics[]{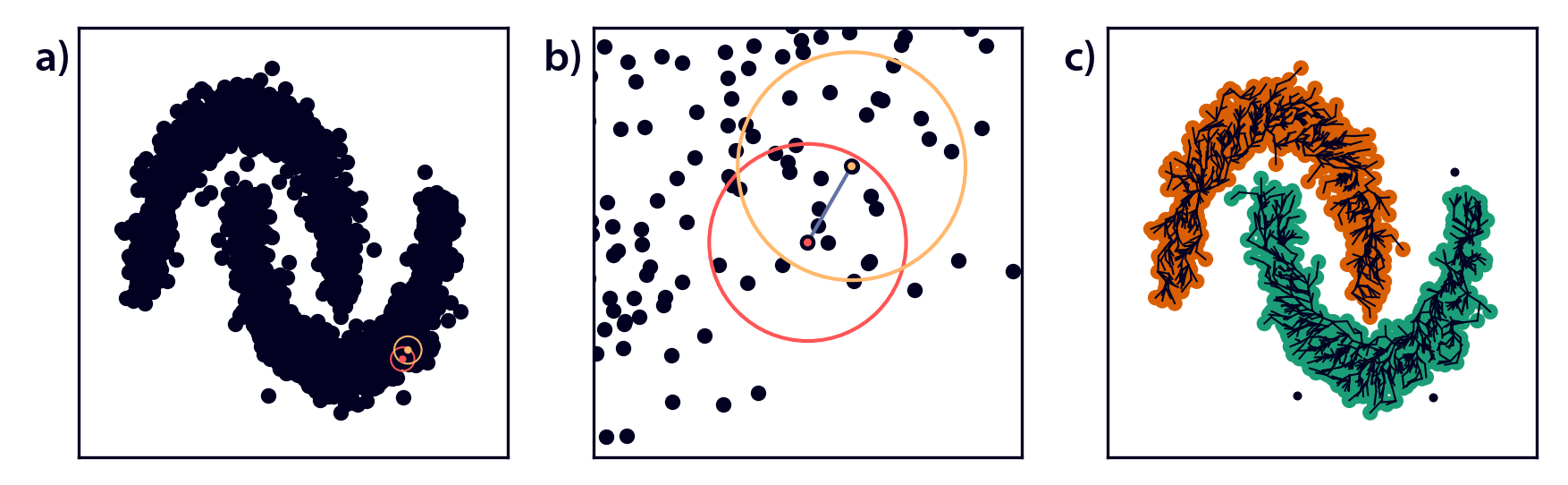}
    \captionofbottom{figure}{%
        Jarvis-Patrick for the \textit{moons} data set}{%
        Jarvis-Patrick for the \textit{moons} data set}{%
            \subl{a} Original data points with \(k\)-nearest radii
            (\(k=20\)) shown for a data point (red) and its 15th closest
            neighbour (orange). \subl{b} For a density threshold of
            \(n_\mathrm{c} \leq 11\), the points highlighted in \subl{a}
            are connected since they share 11 neighbours. \subl{c} The
            connections between the data points form a graph in which
            clusters correspond to maximally connected components. For a
            clustering with \(n_\mathrm{c} = 10\), the data points are
            coloured by their cluster labels (noise in black). The edges
            of a respective graph build from a connected component
            search (see code snippet in section~\ref{sec:dbscan}) are
            shown with black lines.}
    \label{fig:jp_moons}
\end{illustration}

\begin{illustration}
    \includegraphics[]{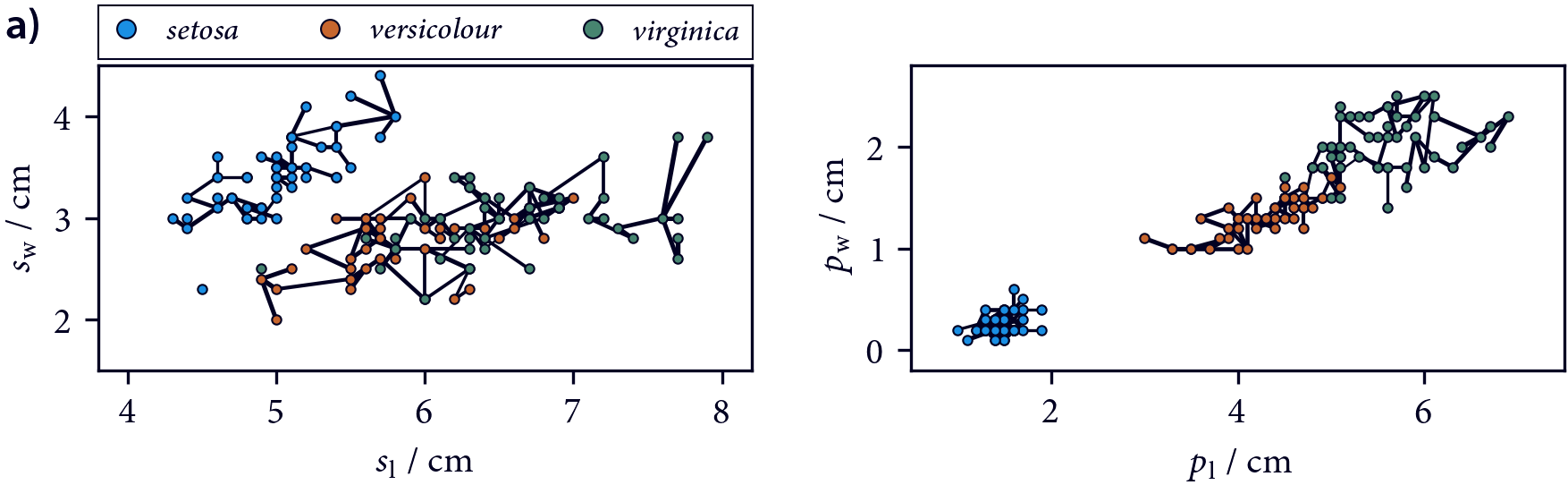}
    \includegraphics[]{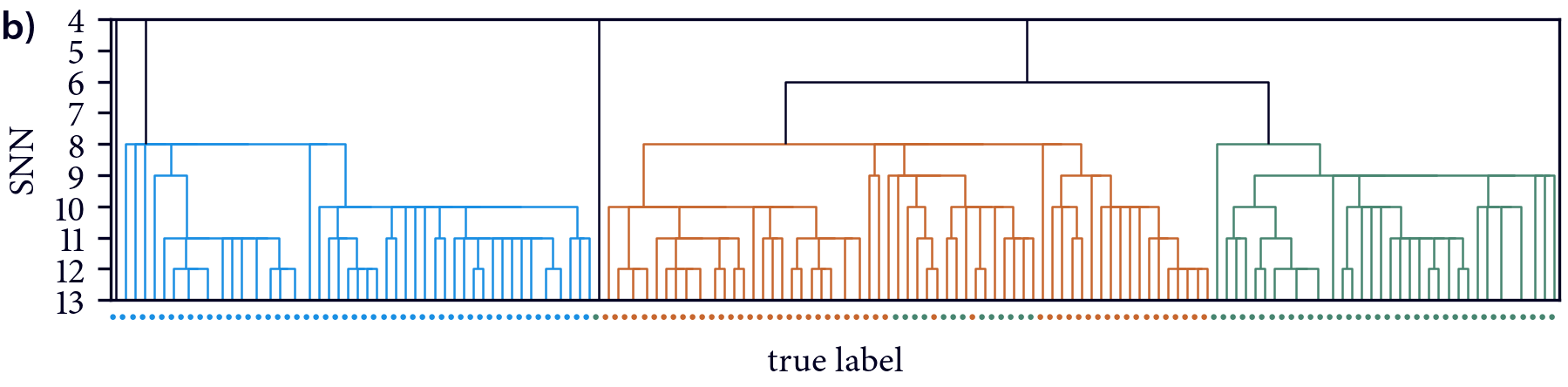}
    \includegraphics[]{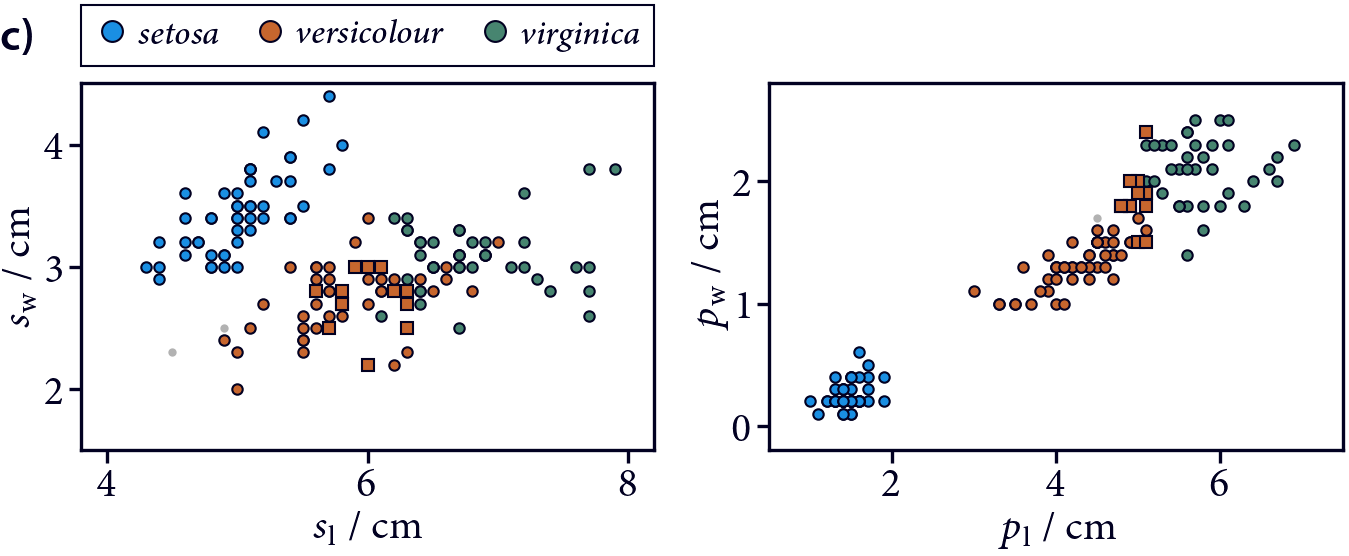}
    \captionofbottom{figure}{%
        \textit{Iris} data Jarvis-Patrick hierarchically}{%
        \textit{Iris} data Jarvis-Patrick hierarchically}{%
            \subl{a} Original data points with true classification
            labels (compare
            figure~\ref{fig:iris_data_points_true_labels}) and \gls{mst}
            using the Jarvis-Patrick connectivity criterion for \(k =
            15\). The edges of the tree are shown with black lines,
            scaled by their weight. \subl{b} Single-linkage hierarchy
            illustrated as a dendrogram. By investigating the hierarchy,
            it is possible to select three clusters, corresponding to a
            flat clustering with \(n_\mathrm{c} \in \{6, 7\}\). The legs
            of the tree are coloured by the resulting cluster labels
            while circles on the bottom denote the true classification
            of individual points. A threshold \(n_\mathrm{c} \leq 6\)
            would lead to a merge of the orange and green cluster. A
            threshold \(n_\mathrm{c} \geq 7\) creates several splits in
            the shown clusters. \subl{c} Flat clustering for
            \(n_\mathrm{c} = 7\) result as scatter plot.
            \SI{90}{\percent} of the cluster labels match the true
            classification labels (\SI{91}{\percent} if noise points are
            neglected). Non-matching assignments are marked with
            squares.}
    \label{fig:jp_iris_selected_clusters}
\end{illustration}

The size of the \(k\)-nearest neighbourhoods to evaluate is kind of a resolution
parameter comparable to the neighbour search radius \(r\) in DBSCAN.
It should be noted, though, that the \(k\)-nearest neighbourhoods can not be
associated with a fixed spatial volume.
For data points in sparse environments, the distances to their \(k\)th
neighbour are naturally much larger than for dense points.
In principle, the density estimate can be done for all possible pairs of points,
which essentially constructs a similarity matrix.
However, a crucial additional requirement is usually made: points for which the
similarity is evaluated must be a \(k\)th nearest neighbour of each other.
The similarity is set to zero for all other pairs.
Without this limitation, the outcome of the clustering is altered
substantially.
For the identification of clusters as connected components, a
minimal set of local connections between points is sufficient.
Consequently, what is practically dealt with in implementations of this
clustering is not the full similarity matrix but rather a subset of necessary
connections in terms of edges in a graph.
Similar to the already described density-based clustering schemes,
Jarvis-Patrick clustering is traditionally used with a threshold to produce a
flat clustering.
Points that share at least \(n_\mathrm{c}\) common neighbours with respect
to their \(k\)-nearest neighbourhoods are identified to belong to the same
cluster.
Figure~\ref{fig:jp_moons} illustrates the Jarvis-Patrick density-criterion
on the scikit-learn \textit{moons} data set.

What has been stated about hierarchical clustering in the previous sections,
can also be translated almost one-to-one for Jarvis-Patrick clustering as well.
The basic idea is to use the similarity between data points quantitatively
instead of converting it to a binary relation using a threshold.
Figure~\extref{fig:jp_iris_selected_clusters}{a} shows a \gls{mst} of
edges corresponding to the Jarvis-Patrick density estimate for the
\textit{Iris} data set.
By single-linkage clustering of this tree, the full hierarchy of clustering
results with increasing threshold can be built and analysed as shown in
figure~\extref{fig:jp_iris_selected_clusters}{b}.
Three clusters in close agreement with the expectation can be selected,
which are shown in figure~\extref{fig:jp_iris_selected_clusters}{c}.
%

It should be noted, however, that the \gls{mst} of the data
and consequently the hierarchy of clustering results still depends on
the cluster parameter \(k\), i.e. number of nearest neighbours to be
considered in the point neighbourhoods.
%

\section{\Glsfmtlong{commonnn} clustering}
\label{sec:commonnn_theory}

\firstsecpar{A variation of the Jarvis-Patrick} clustering approach
is found in the formulation of an independent method, referred to as
\glsfirst{commonnn} clustering.
As an alternative to the use of \(k\)-nearest neighbourhoods and the SNN
similarity in Jarvis-Patrick clustering, \gls{commonnn} clustering uses
fixed radius neighbourhoods with an accordingly modified \textit{shared
fixed radius near neighbours} similarity (for which to our knowledge no
widely accepted abbreviation exists).
In other words, the similarity between a pair of data points in
\gls{commonnn} clustering is defined as the number of points that can be
found in both the fixed radius neighbourhoods of each point, i.e. as the
number of their in this sense \textit{common} neighbours.
A differentiation between common near(est) and shared nearest neighbours
in terms of a similarity definition just by the naming is admittedly a
bit blurry, though.
If the character of the neighbour lists is neglected, Jarvis-Patrick and
\gls{commonnn} clustering are basically identical.
From an objective standpoint it would make sense to treat the two
clustering methods as different flavours of essentially the same method,
let's call it \textit{shared neighbours} clustering for that
matter.\stdmarginnote{shared neighbours clustering}
Besides, shared neighbours clustering with fixed radius near neighbours
or \(k\)-nearest neighbours, any other neighbourhood definition might be
used as well---possibly, though, with drastically different outcome.
Like for Jarvis-Patrick clustering, we will see for shared neighbours
clustering in general that it can in turn be viewed as a form of
single-linkage clustering with a density derived similarity measure.

\begin{illustration}
    \includegraphics[]{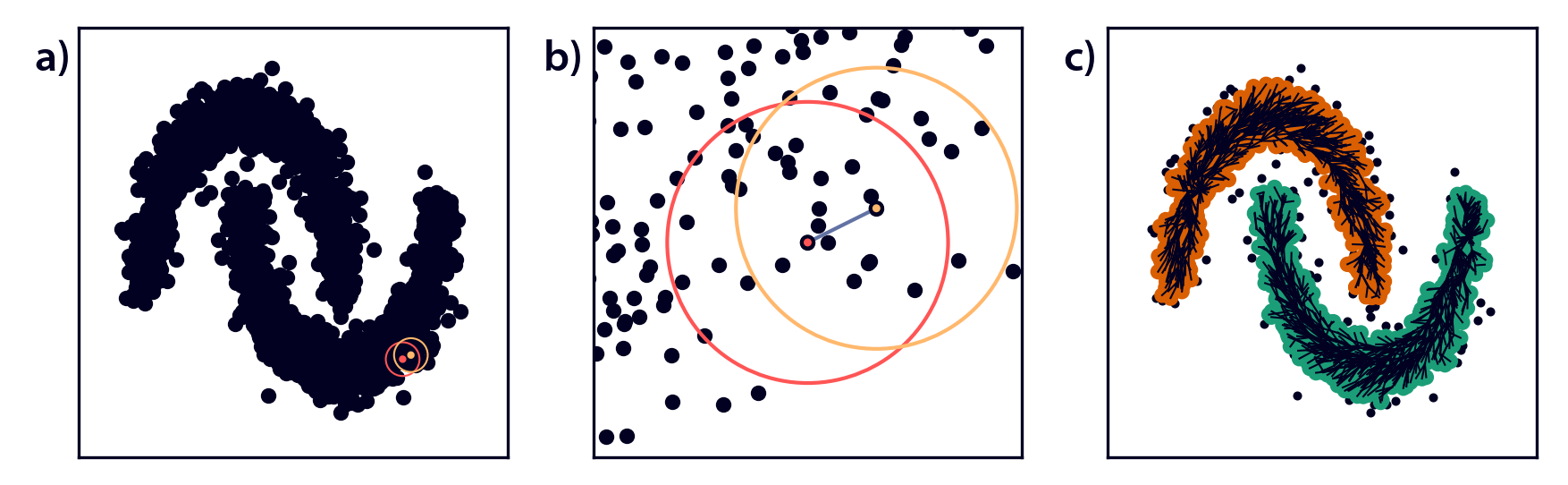}
    \captionofbottom{figure}{%
        \Gls{commonnn} for the \textit{moons} data set}{%
        \Gls{commonnn} for the \textit{moons} data set}{%
            \subl{a} Original data points with fixed near neighbourhood
            radii (\(r=0.2\)) shown for a data point (red) and one of its
            neighbours (orange). \subl{b} For a density threshold
            of \(n_\mathrm{c} \leq 18\), the points highlighted in
            \subl{a} are connected since they share 18 neighbours.
            \subl{c} The connections between the data points form a
            graph in which clusters correspond to maximally connected
            components. For a clustering with \(n_\mathrm{c} = 10\), the
            data points are coloured by their cluster labels (noise in
            black). The edges of a respective graph build from a
            connected component search (see code snippet in
            section~\ref{sec:dbscan}) are shown with black lines.}
    \label{fig:commonnn_moons}
\end{illustration}

\begin{illustration}
    \includegraphics[]{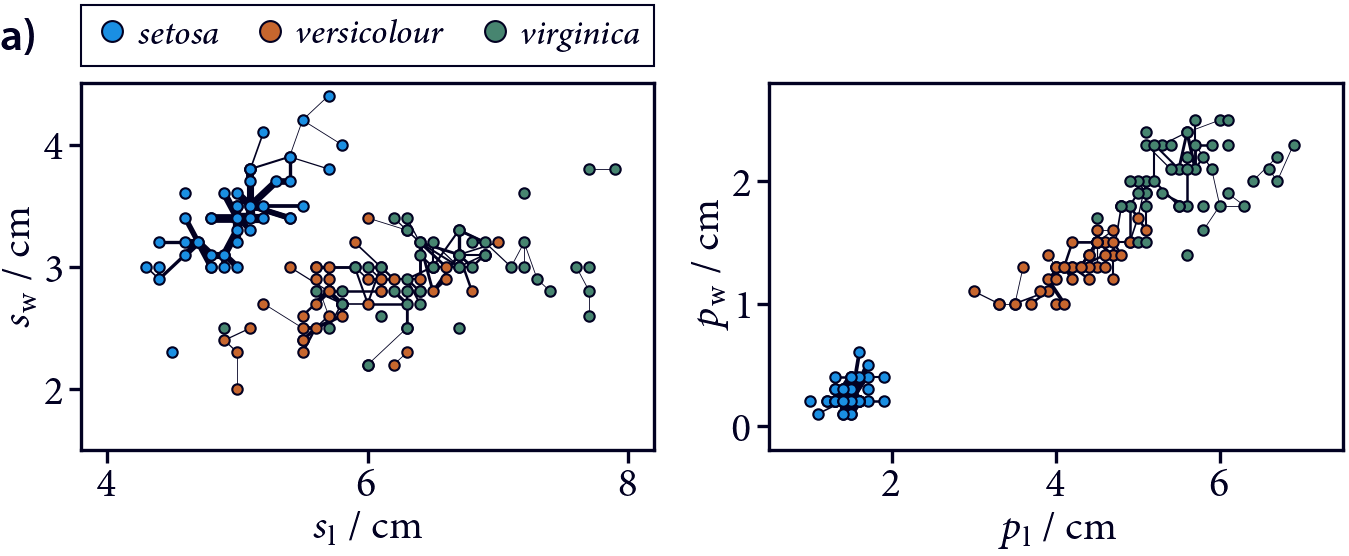}
    \includegraphics[]{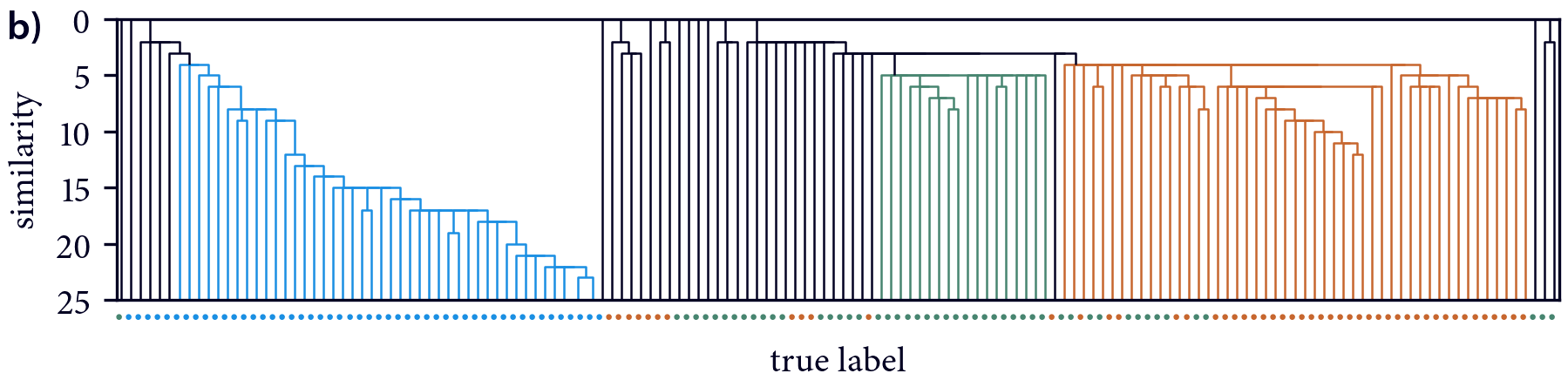}
    \includegraphics[]{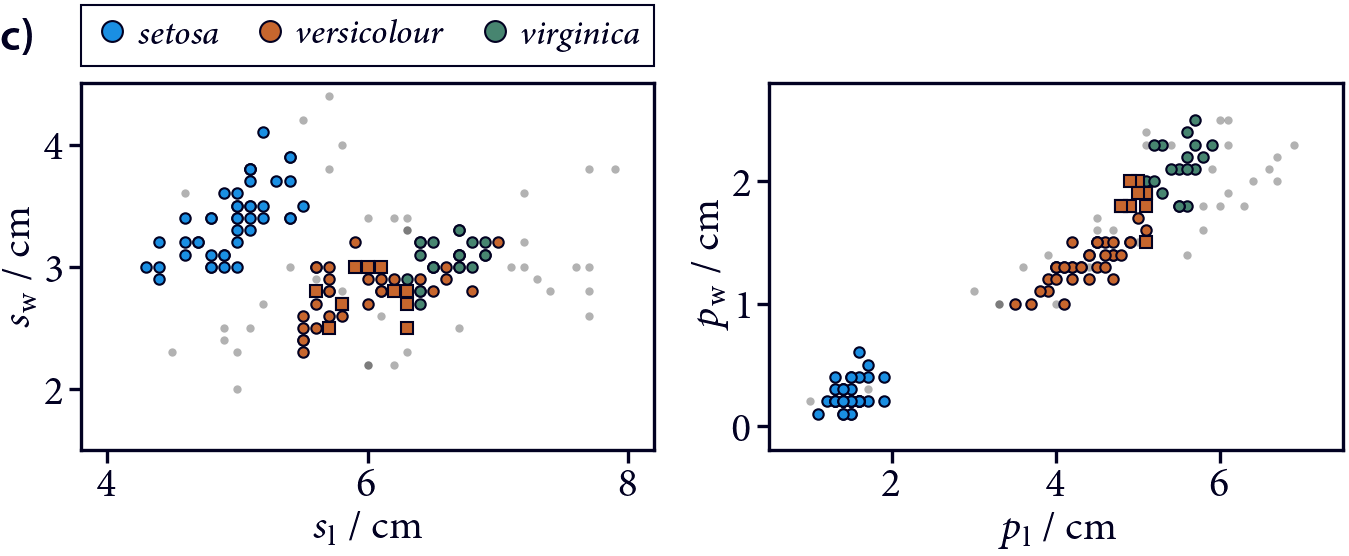}
    \captionofbottom{figure}{%
        \textit{Iris} data \gls{commonnn} hierarchically}{%
        \textit{Iris} data \gls{commonnn} hierarchically}{%
            \subl{a} Original data points with true classification
            labels (compare
            figure~\ref{fig:iris_data_points_true_labels}) and \gls{mst}
            using the \gls{commonnn} connectivity criterion for \(r =
            0.45\). The edges of the tree are shown with black lines,
            scaled by their weight. \subl{b} Single-linkage hierarchy
            illustrated as a dendrogram. By investigating the hierarchy,
            it is possible to select three clusters, corresponding to a
            flat clustering with \(n_\mathrm{c} = \{4\}\) (note that
            this includes an offset of 2 due to neighbour
            self-counting). The legs of the tree are coloured by the
            resulting cluster labels while circles on the bottom denote
            the true classification of individual points. \subl{c} Flat
            clustering result for \(n_\mathrm{c} = 4\) as scatter plot.
            \SI{67}{\percent} of the cluster labels match the true
            classification labels (\SI{90}{\percent} if noise points are
            neglected). Non-matching assignments are marked with
            squares.}
    \label{fig:commonnn_iris_selected_clusters}
\end{illustration}

In the original publication of 2010\cite{Keller2010a}, the name
\glsfmtlong{commonnn} clustering was coined in analogy to the previously developed
\textit{neighbour} algorithm, which is conceptionally, however, quite
different.\cite{Daura1999}
Later, the abbreviation CNN clustering was used.\cite{Lemke2016,Lemke2018}
This is avoided here, however, because CNN is prominently occupied by
the concept of \textit{convolutional neural networks}.
While this term falls into the same broader topic of machine learning
like clustering does in general, it is otherwise completely
unrelated and thus can be only confusing.
Currently, the method is developed under the name \glsfmtshort{commonnn}
clustering as it is refered to here.\footnote{Visit the development
repository on GitHub:
\href{https://github.com/bkellerlab/CommonNNClustering}{\captionurl{https://github.com/bkellerlab/CommonNNClustering}}}

Figure~\ref{fig:commonnn_moons} illustrates the \gls{commonnn}
density-criterion on the scikit-learn \textit{moons} data set.
Like the other connectivity based clustering methods discussed so far,
it is traditionally used with a threshold.
Points that share a number of at least \(n_\mathrm{c}\) neighbours with
respect to a neighbour search radius \(r\) will be considered connected,
rendering the final clusters connected components of the graph formed
by these connections.

\Gls{commonnn} clustering can also be done hierarchically by building a
\gls{mst} from the un-truncated density criterion.
Figure~\ref{fig:commonnn_iris_selected_clusters} exemplifies this
with the \textit{Iris} data set.
Note that the result may not be fully satisfactory in this case but
this can be attribute to the relatively low number of samples in the data
set.
Density-based clustering in general depends on a sufficiently high number
of data points for a robust density estimate.
This is even more true for \gls{commonnn} clustering  where the density
estimate relies on a sufficient sampling of neighbourhood intersections.
The fact that other clusterings like DBSCAN and Jarvis-Patrick clustering
seem to perform better for the \textit{Iris} data set should not be taken
prematurely as a general qualitative difference.


\section{Density-peaks}
\label{sec:density_peaks}

\firstsecpar{The last density-based clustering procedure} I want to put
some attention to is density-peaks clustering.\cite{Rodriguez2014}
So far we discussed connectivity-based density-based clusterings that
essentially aimed on the identification of connected components of a
super level-set on an approximate probability density.
The hierarchies (or hierarchy slices) presented by these methods can be
understood in terms of level-set trees, i.e. there has been a focus on
where a considered data set splits when a density tuning parameter exceed
a certain threshold value.
In a way we could say that this view is centred on the minima of the probability
distribution underlying a data set.
By excluding regions of low density, the high density regions reveal
themselves as disjoint components.
Density-peaks clustering is interesting because it can be in contrast
considered a prototype-based approach.
This perspective is focused on the maxima of the (approximate) probability
density of a sample set.\stdmarginnote{prototypes}
By identification of the highest density data points, the method tries
to find those that could be suitable prototypes for the modes of the distribution
that attract the lower density points around them.
This idea is similar to what is done by mean-shift
clustering,\cite{Comaniciu2002} where a set of test points is converged
to the closest density maxima by updating their position iteratively to
be the mean of their respective neighbourhoods.
The way how density-peaks finds the maxima is, however, quite different.
An important thing to point out is that density-peaks is able to find
non-spherical clusters of arbitrary shape and form---a trait commonly
only attributed to connectivity-based clustering procedures.
Density-peaks is based around two assumptions: 1) the desired cluster
prototypes (the cluster centres) are points of relatively high density,
surrounded by neighbouring points with lower density, and 2) they are relatively
far away from other points of high density in the sense of 1).
Practically, we need to assess the density of individual points and
their distance to the nearest point of higher
density.\stdmarginnote{implementation}
Local point density can be estimated around each point, like seen before
in DBSCAN (section~\ref{sec:dbscan}), as the number of neighbouring
points with respect to a neighbour search radius \(r\).
Lets denote this number by \(\rho^\prime_a\), the density estimate for
point \(a\).
The authors of density-peaks claim that the clustering is robust against
variations in \(r\) because only the relative proportion of density
differences between points is of interest, not the absolute value of the
density estimate.
So while \(r\) can still be seen as a resolution parameter that should
be set in a reasonable range---a too low value might give a noisy
estimate, a too large value might not be able to resolve density
differences---it should not have a direct effect on the clustering
result in terms of a tuning parameter.
Next, let's denote the distance of each point to the closet denser point
as
\begin{equation}
    \delta_a = \min_{b\,|\,\rho^\prime_b > \rho^\prime_{a\phantom{b}}}\vspace{-2pt}\bigl(d(a, b)\bigr)\,,
    \label{eq:delta_dpc}
\end{equation}
where \(d(a, b)\) is a distance function.
The densest point is conventionally assigned the maximum distance found
in the data set.
Proper prototypes are expected to have a much larger \(\delta_a\) value
than other points.
Points that at the same time have a very low density may be rather
considered outliers, though.
For the actual selection of clusters, density-peaks provides a decision
heuristic in terms of a two dimensional plot of \(\delta_a\) versus\stdmarginnote{cluster selection}
\(\rho\prime_b\) (see figure~\extref{fig:iris_data_points_dpc}{a}).
The user can manually select those points that are both dense and far away
from other dense points.
All remaining points will be assigned recursively to the same cluster
as their closest point with higher density.
Note that in contrast in mean-shift clustering, points are typically
assigned to their closest prototype, which tends to yield globular
clusters and ignores the fact that the closest prototype may represent
an actually different data region of high density.

\begin{illustration}
    \begin{minipage}[t]{0.25\textwidth}
        \vspace{0pt}
        \includegraphics{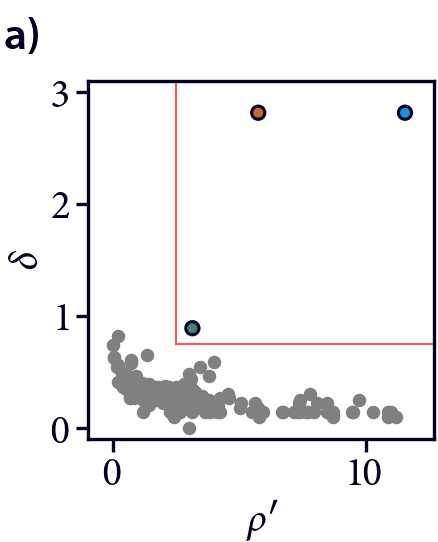}
    \end{minipage}
    \begin{minipage}[t]{0.75\textwidth}
        \vspace{0pt}
        \includegraphics{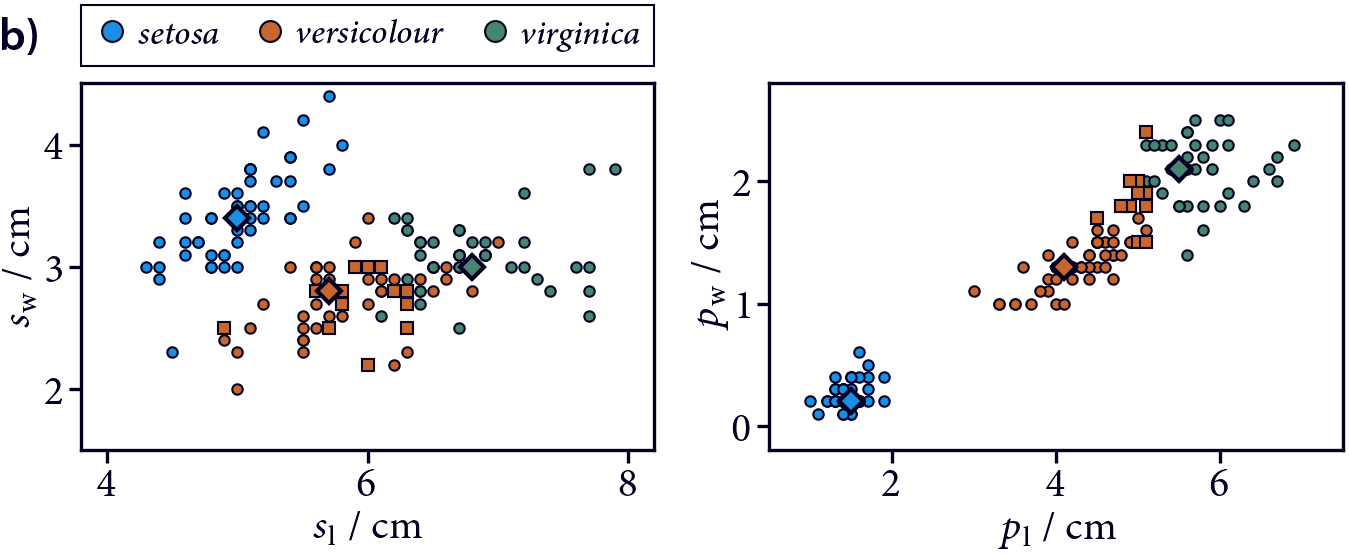}
    \end{minipage}
    \begin{minipage}[t]{1\textwidth}
        \vspace{0pt}
        \captionofbottom{figure}{%
            \textit{Iris} data set density-peaks (3 clusters)}{%
            \textit{Iris} data set density-peaks (3 clusters)}{%
                \subl{a} Density-peaks decision graph for cluster
                selection. Valid cluster prototypes are supposed to
                stand out in terms of a relatively high density and a
                relatively large distance to the next densest data
                point. \subl{b} Data points (compare
                figure~\ref{fig:iris_data_points_true_labels}) with
                cluster labels from density-peaks clustering
                (\cminline{pydpc.Cluster}\footnotemark).
                Prototypes highlighted with diamond shapes.
                \SI{91}{\percent} of the cluster labels match the true
                classification labels. Non-matching assignments are
                marked with squares.}
            \label{fig:iris_data_points_dpc}
    \end{minipage}
\end{illustration}
\footnotetext{Visit the development repository on GitHub: \href{https://github.com/cwehmeyer/pydpc}{\captionurl{https://github.com/cwehmeyer/pydpc}}}

Within the limits of the applied heuristic to select clusters,
density-peaks provides kind of a hierarchical view on the data set as well,
only that child-parent relationships are not made explicit.\stdmarginnote{hierarchy}
By systematically including an increasing number of cluster centres in
the result, data regions that where mingled with the respectively
closest denser region become succeedingly separated as their own
clusters.
The decision of which density-peaks are relevant in the end has to be made
by the user.
%

\clearpage
\AtNextBibliography{\footnotesize}
\phantomsection
\thispagestyle{plain}
\renewcommand\bibname{References\label{chap:references}}
\begin{multicols}{2}[\printbibheading]
\printbibliography[heading=none]
\end{multicols}
\addcontentsline{toc}{chapter}{References}

\end{document}